\documentclass[11pt,twoside]{article}

\usepackage{fullpage}

\usepackage{amsmath,amssymb,fullpage,graphicx,xcolor,mathtools,nicefrac,comment}
\usepackage[shortlabels]{enumitem}

\usepackage{amssymb,amsmath,comment}
\usepackage[most]{tcolorbox}

\usepackage{amsthm}
\makeatletter
\def\th@plain{%
  \thm@notefont{}
  \itshape 
}
\def\th@definition{%
  \thm@notefont{}
  \normalfont 
}
\makeatother

\usepackage[shortlabels]{enumitem}

\usepackage{thmtools,thm-restate}

\usepackage{hyperref}

\usepackage{amsmath,amssymb}
\usepackage{verbatim,float,url}
\usepackage{graphicx,psfrag,subcaption}
\usepackage[numbers]{natbib}
\usepackage{xcolor}
\usepackage{microtype}
\usepackage{xparse}
\usepackage{xspace}
\usepackage{tikz,textcomp,thmtools,nameref,cleveref}
\usetikzlibrary{positioning}
\usepackage{tikz-qtree}

\usepackage{algpseudocode,algorithmicx,algorithm}

\usepackage[font=small]{caption}

\newtheorem{theorem}{Theorem}

\newtheorem{dhruvprop}{Proposition}
\newtheorem{dhruvdef}[theorem]{Definition}
\newtheorem{dhruvthm}{Theorem}
\newtheorem{dhruvlemma}{Lemma}
\newtheorem{dhruvcorollary}{Corollary}

\DeclareMathOperator*{\argmin}{arg\,min}

\newcommand{\qeddhruv}{$\blacksquare$}

\newcommand{\bigo}{\widetilde{\mathcal{O}}}
\newcommand{\littleo}{\widetilde{o}}

\newcommand{\numact}{K}
\newcommand{\ActSet}{\mathcal{X}}

\newcommand{\cpReg}{\mathcal{R}^{\text{cp}}}

\newcommand{\E}{\mathbb{E}}
\newcommand{\bigtheta}{\widetilde{\Theta}}
\newcommand{\bigomega}{\widetilde{\Omega}}

\newcommand{\UTB}{\mathsf{UTB}}
\newcommand{\tbprob}{\mathsf{tb}}
\newcommand{\alg}{\mathcal{A}}
\newcommand{\algset}{\overline{\mathcal{A}}}

\newcommand{\R}{\mathbb{R}}

\newcommand{\mup}{M}

\begin{document}

\begin{center}
{\bf{\LARGE{Weighted Tallying Bandits: \\Overcoming Intractability via Repeated Exposure Optimality}}}


\vspace*{.2in}

{\large{
\begin{tabular}{cccc}
Dhruv Malik & Conor Igoe & Yuanzhi Li & Aarti Singh
\end{tabular}
}}
\vspace*{.2in}

\begin{tabular}{c}
Machine Learning Department \\
Carnegie Mellon University
\end{tabular}

\vspace*{.2in}

\today

\end{center}
\vspace*{.2in}

\begin{abstract}
In recommender system or crowdsourcing applications of online learning, a human's preferences or abilities are often a function of the algorithm's recent actions. Motivated by this, a significant line of work has formalized settings where an action's loss is a function of the number of times that action was recently played in the prior $m$ timesteps, where $m$ corresponds to a bound on human memory capacity. To more faithfully capture decay of human memory with time, we introduce the Weighted Tallying Bandit (WTB), which generalizes this setting by requiring that an action's loss is a function of a \emph{weighted} summation of the number of times that arm was played in the last $m$ timesteps. This WTB setting is intractable without further assumption. So we study it under Repeated Exposure Optimality (REO), a condition motivated by the literature on human physiology, which requires the existence of an action that when repetitively played will eventually yield smaller loss than any other sequence of actions. We study the minimization of the complete policy regret (CPR), which is the strongest notion of regret, in WTB under REO. Since $m$ is typically unknown, we assume we only have access to an upper bound $M$ on $m$. We show that for problems with $K$ actions and horizon $T$, a simple modification of the successive elimination algorithm has $\bigo \left( \sqrt{KT} + (m+M)K \right)$ CPR. Interestingly, upto an additive (in lieu of mutliplicative) factor in $(m+M)K$, this recovers the classical guarantee for the simpler stochastic multi-armed bandit with traditional regret. We additionally show that in our setting, any algorithm will suffer additive CPR of $\bigomega \left( mK + M \right)$, demonstrating our result is nearly optimal. Our algorithm is computationally efficient, and we experimentally demonstrate its practicality and superiority over natural baselines.
\end{abstract}

\section{Introduction}
\label{sec:intro}
When online learning algorithms are deployed in interactive applications, the algorithm's decisions impact the state of the environment. In turn, this impacts the quality of subsequent decisions made by the algorithm. This is especially true in human-centered applications such as recommender systems or crowdsourcing. For instance, consider a crowdsourcing setting where at each timestep we want to select a worker to perform a task, without prior knowledge of any worker's ability. The task may be complex or require some fine-tuning, and each worker might need a calibration period where they repeatedly perform the task, before they start exhibiting their true performance. The existence of such a calibration period has been extensively demonstrated in tasks that require visuomotor calibration~\citep{adams61}, such as throwing darts~\citep{wunderlich20} or shooting a basketball~\citep{phatak20}. Hence, an algorithm that asks workers to alternately perform the task, without intelligently allowing each worker time to calibrate themselves to the task, may bias its estimation of each worker's true ability. This interaction between an algorithm and its environment separates this scenario from classical non-interactive frameworks such as the multi-armed bandit.

To capture one aspect of this interactivity, a significant research thrust in online learning has studied settings where an action's loss is described by the number of times that action was recently played in the prior $m$ timesteps~\citep{heidari16, levine17, seznec19, seznec20, lindner21, awasthi22, malik22}. The quantity $m$ typically corresponds to a bound on human memory capacity or capability. For example, in the aforementioned scenario $m$ would be the number of timesteps required by a worker to fine-tune and calibrate themselves to the task, before revealing their true ability.

Of course, such settings are an approximation to reality. For instance, psychological research demonstrates that humans typically display a better memory for more recently occurring events~\citep{klatzky80, ricker16}. So if we play an action once in the previous $m$ timesteps, its impact on the present may greatly differ depending on whether it was played on the previous timestep or $m$ timesteps ago. In the context of the aforementioned crowdsourcing setting, a worker may need a shorter calibration period if they performed the task on the previous timestep, as opposed to many timesteps ago. However, prior formalizations are oblivious to this difference. Motivated by these considerations, we make the following contributions:
\begin{itemize}
\item We introduce the Weighted Tallying Bandit (WTB), which generalizes prior formalizations by requiring that an action's loss is described by a \emph{weighted summation} of the number of times that action was played in the prior $m$ timesteps. Since this setting is dynamic and interactive, we eschew the traditional regret, and instead study the minimization of the strongest notion of regret known as the complete policy regret (CPR).
\item We show that minimizing CPR in WTB is generally intractable. So we study it under the additional condition of Repeated Exposure Optimality (REO), which enforces the existence of an action that when repetitively played $m$ times will yield smaller loss than other action sequences. In the context of the aforementioned example, REO is interpreted as the existence of a worker that once calibrated to the task, will perform better than other (calibrated or uncalibrated) workers. We motivate this condition via literature on human physiology.
\item For WTB problems with $K$ actions and horizon $T$ that satisfy REO, and in the regime where only an upper bound $M$ on the true value of $m$ is known, we show that a slight modification of the classical successive elimination algorithm achieves a CPR guarantee (upto a logarithmic factor) of $\bigo \left( \sqrt{KT} + (m + M)K \right)$. Besides an additive factor in $(m+M)K$, this matches the lower bound on the weaker traditional regret of the stochastic multi-armed bandit (which is the special $m=1$ case of WTB with REO).
\item While one may desire an algorithm that is fully adaptive to $m$ and requires no such upper bound $M$, we show this is impossible. Concretely, we show that any algorithm with sublinear CPR must require such an upper bound $M$, and then show that a linear dependency on this input $M$ is necessary. This implies a lower bound of $\bigo \left( \sqrt{KT} + mK + M \right)$ on the achievable CPR of any algorithm in our setting, highlighting our algorithm's near optimality.
\item Via diverse numerical simulations, we demonstrate our (computationally efficient) method's practicality and superiority over various baselines.
\end{itemize}

\section{Problem Formulation}
\label{sec:prob_form}
\subsection{Weighted Tallying Bandit}
\label{sec:prob_form_wtb}
We begin by formally introducing the Weighted Tallying Bandit as an online learning game with bandit feedback over time horizon $T$, where the player has access to an action set $\mathcal{X}$ with finite cardinality $K$. A long line of prior work has studied the scenario where an action's loss at any timestep is a function of the number of timesteps it was played in the prior $m$ timesteps~\citep{heidari16, levine17, seznec19, seznec20, lindner21, awasthi22, malik22}. We refer to these settings as ``tallying'' settings. Our goal is to generalize this work, to the case where an action's loss is a function of a weighted tally of the number of times it was played in the past $m$ timesteps.

To this end, we first introduce some notation. Assume the player has played the game for $t$ timesteps, and for each timestep $1 \leq t' \leq t$ the player plays action $a_{t'}$. For a fixed action $x \in \ActSet$, we define the vector $y^{t,x,m} \in \{ 0, 1 \}^m$ in a componentwise fashion as
$$
y^{t,x,m}_i =
\begin{cases}
\mathbb{I}( a_{t - i + 1} = x ) &\text{ if $t - i + 1 \geq 1$} \\
0 &\text{ if $t - i + 1 < 1$}
\end{cases},
$$
for each component $1 \leq i \leq m$. Hence, the vector $y^{t,x,m}$ stores the timesteps where action $x$ was played in the previous $m$ timesteps upto (and including) the current timestep $t$. With this notation in hand, we are now in a position to formally define the Weighted Tallying Bandit.

\begin{dhruvdef}[Weighted Tallying Bandit (WTB)]
An online learning game is said to be an $(m, w, h)$-weighted tallying bandit with memory capacity $m$, if there exists an integer $m \geq 1$, a set of vectors $\{ w_x \}_{x \in \ActSet} \subset (0,1]^m$, and a set of functions $\{ h_x \}_{x \in \ActSet}$ each mapping from $\R$ to $[0,1]$, such that the following is true. For each $x \in \ActSet$, the expected loss incurred at timestep $t$ by playing action $a_t = x$ is given by
$$
h_{x} \left( w_x^{\top} y^{t,x,m} \right),
$$
and the player observes as feedback a random observation $\widetilde{h}_{x}(w_x^{\top} y^{t,x,m}) \in [0,1]$, that is independent of all other random observations, and satisfies
$$
\E \left[ \widetilde{h}_{x}(w_x^{\top} y^{t,x,m}) \right] = h_{x} \left( w_x^{\top} y^{t,x,m} \right).
$$
\end{dhruvdef}
\noindent In general, the quantities $m, \{ w_x \}_{x \in \ActSet}, \{ h_x \}_{x \in \ActSet}$ are all unknown, and the player only learns about them via bandit feedback over time. When $m=1$, then WTB recovers the stochastic multi-armed bandit (sMAB)~\citep{lai85, auer02stochastic}. However, WTB with $m \geq 2$ is often a better model for human-centered applications that require calibration. To understand this, let us concretize the crowdsourcing application introduced in Section~\ref{sec:intro}. Assume that the task to be performed is throwing a dart at a dartboard, and that each worker is a different darts player. Without prior knowledge of any player's true ability to hit the dartboard, our goal is to discover which of the $K$ players is the best, by picking (at each timestep) a player to throw a dart and seeing whether they hit or miss. At first glance, this seems to be a classical sMAB problem, where each player has some true ability, and each time we query a player we (stochastically) observe their true ability.

Unfortunately, this sMAB formulation is agnostic to the calibration period that darts players require before they can exhibit their true performance. The existence of such a calibration period has been demonstrated in the literature on visuomotor calibration. For instance, Wunderlich et al.~\citep{wunderlich20} show that when professional darts players toss darts in a row, the first toss is significantly less accurate than the remainder of the darts, although the performance stabilizes after the first dart toss. They attribute this phenomenon to the \emph{warm-up decrement}~\citep{adams61, anshel93, anshel95}, which describes the decline in performance due to a break in a specific motor skill, as well as its recovery once the skill is resumed. Simply put, a player performs better once they are ``in motion'' and have fine-tuned their movement parameters after their first toss.

This phenomenon affects the design of algorithms for our darts setting, since we do not observe the true performance of a dart player until after their first toss. Furthermore, this \emph{cannot} be resolved by simply having each player toss once, so that they are calibrated, and then running a standard sMAB algorithm while assuming that the players stay calibrated forever. Indeed, Wunderlich et al.~\citep{wunderlich20} demonstrate that even small interruptions in the dart tosses (such as the few seconds required for the player to retrieve their darts from the board) can cause the player to ``reset'', and subconsciously lose their fine-tuned movement parameters. Hence, the sMAB is hence a poor model for this setting. By contrast, the WTB with $m \geq 2$ is a more faithful model, since $m$ describes the number of times a player must toss a dart in a row before we (stochastically) observe their true performance. The ``reset'' phenomenon that exists in this motivating example (as well as our forthcoming examples) requires that if we model this problem with WTB, then $m$ should be non-trivially smaller than the horizon $T$. We will assume this throughout our paper.

WTB more naturally models this phenomenon than the aforementioned tallying settings~\citep{heidari16, levine17, seznec19, seznec20, lindner21, awasthi22, malik22}, which are all special cases of the WTB where $w_x$ is the all ones vector $\vec 1$ for each $x \in \ActSet$. As a stylized example, assume the task is shooting basketball free-throws, and that we need to find the best of two players $x_1, x_2$. Consider two different sequences of selecting players --- $x_1, x_2, x_1$ versus $x_2, x_1, x_1$. Phatak et al.~\citep{phatak20} show that players require a calibration period of length at least 3 while shooting free-throws, and that their shooting performance monotonically improves with each successive free-throw. This implies that picking $x_1, x_2, x_1$ (i.e., $x_1$ shoots, then $x_2$, then $x_1$ again) will cause $x_1$ to have a worse expected performance on her final shot, relative to her performance if we select $x_2, x_1, x_1$ (i.e., $x_2$ shoots, then $x_1$ shoots twice). If we model this with WTB where $m = 3$ and $w_x = \vec 1$, then we cannot distinguish these two scenarios, since in both cases $x_1$ shot twice in the past $m$ timesteps. By contrast, WTB with $w \neq \vec 1$ allows us to model different losses for these two scenarios. For instance, if we use the model $w_{x_1} = [1, 1/2, 1/4]$ and $h_{x_1}(z) = 1 - z/3$, then this model says that selecting $x_2, x_1, x_1$ will cause $x_1$ to have better expected performance on her final shot than if we selected $x_1, x_2, x_1$.

More broadly, WTB significantly generalizes prior tallying settings, by allowing us to better approximate the decay in memory strength that occurs with passage of time, that has been documented extensively by studies on short and long term human memory~\citep{klatzky80, ricker16}. This more naturally models the human-centered applications that motivate tallying settings. For instance, Malik et al.~\citep{malik22} motivate their study via recommender systems, arguing that recommended content impacts human preferences, and assume the quantity $m \ll T$ bounds the length of time that a human remembers past recommendations. But their formulation is agnostic to how recently a piece of content was recommended \emph{within} this window of length $m$. So if some content was recommended $k$ times in the past $m$ timesteps, then their framework requires that this incurs the same loss regardless of the ordering of those $k$ recommendations. This is rather limiting, since human preferences today may depend only mildly on recommendations that occurred $\bigomega(m)$ timesteps ago. Our WTB formulation is more fine grained, and allows for the possibility of different losses incurred by each of the different orderings of those $k$ recommendations.
\subsection{Complete Policy Regret}
A key property of WTB is that the loss incurred by an action depends on the past actions of the algorithm. In such dynamic scenarios, it has been well established that the popular traditional regret is inappropriate to measure the performance of an algorithm~\citep{arora12}. Instead, one typically opts for the stronger notion of policy regret~\citep{cesa-bianchi13, arora18}. In line with prior work on tallying settings~\citep{heidari16, levine17, seznec19, seznec20, lindner21, awasthi22, malik22}, we study the minimization of the complete policy regret (CPR), which is the strongest possible notion of regret. Given an $(m, w, h)$-weighted tallying bandit and an algorithm that plays action sequence $(a_1, a_2 \dots a_T) \in \mathcal{X}^T$, the CPR $\cpReg$ of the algorithm is defined as
\begin{equation}
\label{eqn:cpr_main}
\cpReg = \sum_{t=1}^T h_{a_t} \left( w_{a_t}^{\top} y^{t,a_t,m} \right) - \min_{(x_1, x_2 \dots x_T) \in \mathcal{X}^T} \sum_{t=1}^T h_{x_t} \left( w_{x_t}^{\top} y^{t,x_t,m} \right).
\end{equation}
Following prior convention, we refer to any length $T$ sequence of actions (i.e., an element of $\mathcal{X}^T$) as a policy. The CPR is hence the cumulative loss experienced by the algorithm, relative to the minimum loss achieved by the best policy in $\mathcal{X}^T$. Minimizing CPR is hence equivalent to minimizing the cumulative expected loss of the algorithm, and we note that this performance metric is identical to the one used in reinforcement learning~\citep{jin18, wang20}. If the CPR of an algorithm is sublinear in $T$ and polynomial in $m, K$ then we say it has statistically efficient CPR.

Prior work has shown that in the case of WTB with $w_x = \vec 1$ for each $x \in \ActSet$, without any further assumption, there exists an algorithm with statistically efficient CPR~\citep{malik22}. Unfortunately, the following result shows that such an algorithm does not exist in WTB with $w_x \neq \vec 1$.

\begin{dhruvprop}
\label{prop:1}
For any $m \geq 1$, there exists an $(m,w,h)$-weighted tallying bandit with $K=2$ such that the following is true. Any (possibly randomized) algorithm has expected CPR satisfying $\E \left[ \cpReg \right] = \bigomega \left( \min \{ 2^m, T \} / m \right)$.
\end{dhruvprop}
\noindent The proof of Proposition~\ref{prop:1} is deferred to Appendix~\ref{app:prop1_proof}. At a high level, the proof shows that if $w_x \neq \vec 1$ then $h_x$ can take on $\bigomega \left( 2^m \right)$ different values, and so discovering the optimal sequence of actions requires $\bigomega \left( 2^m \right)$ queries. This result demonstrates that if we desire an algorithm with statistically efficient CPR, then we must impose structure on the WTB setting that restricts the set of optimal action sequences. We motivate and formalize such structure in the sequel.

\subsection{Repeated Exposure Optimality}
\label{sec:prob_form_reo}
To motivate additional structure in the types of problems that are modeled by WTB, we recall the darts setting illustrated in Section~\ref{sec:prob_form_wtb}. Notably, if we ask a player to toss darts in a row, then on the \emph{first} toss, their uncalibrated performance is poor and not necessarily indicative of their subsequent performance. But on successive tosses after the first toss, Wunderlich et al.~\citep{wunderlich20} show that their calibrated performance stabilizes and is \emph{better} than the uncalibrated performance on the first toss. A similar observation holds for shooting free-throws~\citep{phatak20}. So if we let $x^\star \in \mathcal{X}$ denote the player with the best calibrated performance, then this implies that the calibrated performance $x^\star$ is better than not only the calibrated performances of player $x \neq x^\star$, but also the uncalibrated performances of all players. We formalize this insight in the following condition.

\begin{dhruvdef}[Repeated Exposure Optimality ($\alpha$-REO)]
An $(m, w, h)$-weighted tallying bandit satisfies the Repeated Exposure Optimality condition with parameter $\alpha$, if there exists an action $x^\star \in \ActSet$, such that for each $x \in \ActSet$ and each $y \in \{ 1 \} \times \{ 0, 1 \}^{m-1}$ we have
$$
h_{x^\star} \left( \| w_{x^\star} \|_1 \right) \leq h_{x} \left( w_x^{\top} y \right) + \alpha.
$$
\end{dhruvdef}
\noindent The $\alpha$-REO condition thus requires that there is some action $x^\star \in \mathcal{X}$, which when played repetitively for at least $m$ times in a row, will have smaller loss (upto the suboptimality $\alpha$) than other action sequences. Two remarks are in order, to understand this condition in the context of prior work. First, observe that even when we additionally impose the $\alpha$-REO condition on WTB, the sMAB remains a special case of this setting via a choice of $\alpha = 0, m=1$. Second, significant prior work on tallying settings has focused on when the loss functions $\{ h_x \}_{x \in \ActSet}$ are monotonic. For instance, the improving bandit~\citep{heidari16} is a special case of WTB under significant additional restrictions, including (but not limited to) the facts that $\{ w_x \}_{x \in \ActSet} = \{ \vec 1 \}$ and $\{ h_x \}_{x \in \ActSet}$ are decreasing. We note that this property of decreasing $\{ h_x \}_{x \in \ActSet}$ functions is a special case of the $0$-REO condition.

We have motivated REO via the warm-up decrement phenomenon that has been extensively demonstrated in the psycho-physiological literature. And we believe REO may also be relevant in other interactive settings such as recommender systems, as we discuss in Section~\ref{sec:discussion}. Nevertheless, we acknowledge that our setting is ultimately a mathematical abstraction that falls short of ground truth reality, and fails to model many subtleties that make human-centered applications challenging. A complete formulation and study of all these subtleties is beyond the scope of our paper, and we relegate discussion of important avenues for future work to Section~\ref{sec:discussion}.

With the REO condition thus motivated and formalized, the following question is natural:
\begin{center}
\emph{Consider any $(m, w, h)$-weighted tallying bandit satisfying $\alpha$-REO. Is there a computationally efficient and practical algorithm that solves this problem with a statistically efficient CPR guarantee?}
\end{center}
The remainder of our paper's analysis is devoted to answering this question.

\section{Main Results}
\label{sec:main_results}
We present two categories of results. In Section~\ref{sec:upper_bound} we present a statistically and computationally efficient algorithm that can solve WTB problems satisfying REO. This method requires only an upper bound $M$ on the true memory capacity $m$, whose exact value is often unknown. In Section~\ref{sec:adaptivity_memory}, we show the impossibility of an algorithm that is fully adaptive to an unknown $m$ (i.e., does not require knowledge of an upper bound $M < T$ on $m$). We also show that if such an upper bound $M < T$ on $m$ is known, then the dependency of our method on $M$ is optimal.

\subsection{A Statistically \& Computationally Efficient Algorithm}
\label{sec:upper_bound}
Before we present our algorithm, let us consider some natural approaches. Since the WTB is a subclass of reinforcement learning (RL) problems, one may attempt to use RL algorithms to solve it. But even when $\{ w_x \}_{x \in \ActSet} = \{ \vec 1 \}$, such algorithms will suffer $\bigomega \left( K^m \right)$ CPR~\citep{awasthi22, malik22}. One may also attempt to extend the classical UCB algorithm from sMAB to WTB as follows. Solve the problem in epochs of length $m$, where at the beginning of each epoch, we select the action that minimizes the usual UCB estimate, and then play it $m$ times in a row instead of just once. Then we record the loss observed in the most recent play, since this is an unbiased estimate of the action's eventual loss, and use it to update the action's UCB estimate. While this seems like a reasonable heuristic, each epoching has an $m$-length overhead which can substantially increase regret.

\begin{algorithm}[hbt!]
\caption{Successive Elimination for WTB}
\label{alg:main}
\begin{algorithmic}[1]
\Require upper bound $\mup$ on memory capacity $m$, time horizon $T$, failure probability tolerance $\delta \in (0,1)$, number of actions $\numact$
\State Define $S = \log_2 \left( \frac{T}{4 \numact \mup} + 1 \right)$.
\State Define $A_1 = \ActSet$.
\State Define $n_s = \numact \mup 2^s / \vert A_s \vert$ and $C_s = \sqrt{ \frac{32}{n_s} \log \left( \frac{2 \numact S}{\delta} \right) }$.
\For{$s \in \{ 1, 2 \dots S \}$}
\For{$x \in A_s$}
	\State Execute action $x$ for $n_s \geq m$ times and store nothing.
	\State Execute action $x$ for $n_s$ times and store $\{ \widetilde{h}_x( \| w_x \|_1 )_{s,k} \}_{k=1}^{n_s}$.
	\State Define $\widehat{\mu}_s(x) = \frac{1}{n_s} \sum_{k=1}^{n_s} \widetilde{h}_x(\| w_x \|_1)_{s,k}$.
\EndFor
\State Select $\widehat{x}_s \in \argmin_{x \in A_s} \widehat{\mu}_s(x)$.
\State Construct $A_{s+1} = \left \{ x \in A_{s} \text{ s.t. } \widehat{\mu}_s(x) \leq \widehat{\mu}_s(\widehat{x}_s) + 2 C_s \right \}$.
\EndFor
\end{algorithmic}
\end{algorithm}

A different idea is to adapt algorithms from prior tallying settings for our problem. But prior tallying settings that are most comparable to ours all have CPR bounds that scale multiplicatively with $m$ (see Section~\ref{sec:related_work} for details). Our key theoretical contribution is to demonstrate that due to the additional presence of REO, we can solve not just these tallying settings but also WTB with a CPR guarantee that is only \emph{additive} (in lieu of multiplicative) in $m$. The algorithm that achieves this bound is a slightly modified version of successive elimination (SE), and is presented in Algorithm~\ref{alg:main}. Our inspiration for this is due to Malik et al.~\citep{malik22}, who adapt SE for their tallying bandit setting, although their modification is more involved. By contrast, our modification is very simple, since REO permits us to only estimate the eventual losses of each action. We now present our main result, which bounds the CPR of this algorithm.

\begin{dhruvthm}
\label{thm:upper}
Fix any $(m,w,h)$-weighted tallying bandit problem satisfying Repeated Exposure Optimality with parameter $\alpha$. When Algorithm~\ref{alg:main} is run with inputs $\mup \geq m$ and $\delta \in (0,1)$, then with probability at least $1 - \delta$ it has complete policy regret upper bounded as
\begin{equation}
\label{eqn:thm_upper_main_eqn}
\cpReg \leq 4 \numact \mup + \numact m \log(T) + 800 \sqrt{\numact T \log \left( \frac{2 \numact \log(T)}{\delta} \right)} + \alpha T.
\end{equation}
\end{dhruvthm}
\noindent The proof of Theorem~\ref{thm:upper} is deferred to Appendix~\ref{app:upper}. Let us highlight some key aspects of this result. \\

\noindent \textbf{Comparison to sMAB \& Tallying Settings.} Recall that in the classical sMAB, which is a special case of WTB with $0$-REO via $m=1$, any algorithm suffers $\bigomega \left( \sqrt{KT} \right)$ traditional regret. Theorem~\ref{thm:upper} thus shows that the much larger class of WTB with $\bigo \left( \sqrt{K / T} \right)$-REO problems can be solved with essentially this guarantee on CPR, upto a logarithmic factor and an additive dependence on $mK$. Our guarantee scales more favorably than those obtained for prior comparable tallying settings (see Section~\ref{sec:related_work} for details). \\

\noindent \textbf{Efficiency \& Practicality}. Algorithm~\ref{alg:main} is computationally efficient and scalable. Its total runtime over $T$ iterations is $\bigo \left( T + K \log(T) \right)$ and the space complexity required at any timestep is $\bigo(K)$. This is in contrast to results on prior comparable tallying settings (see Section~\ref{sec:related_work} for details). Moreover, implementing Algorithm~\ref{alg:main} does not require exact knowledge of unknown quantities such as $\{ h_x \}_{x \in \ActSet}$, $\{ w_x \}_{x \in \ActSet}$, $\alpha$ or $m$; an upper bound $M$ on $m$ suffices. While Algorithm~\ref{alg:main} appears to require the time horizon $T$ as an input, we note that the method is already performing a doubling trick. This means that for any $T$ representable on a computer (say $T \leq 2^{64} = 2^{2^6}$), and since $C_s = \bigo \left( \log \log T \right)$, a short numerical computation reveals that redefining $C_s$ as $\sqrt{\frac{64}{n_s} \log(2K/\delta)}$ and picking $\delta < 0.009$ ensures that we can get the same CPR bound (upto constants) as Eq.~\eqref{eqn:thm_upper_main_eqn}, even without providing $T$ as an input. \\

\noindent \textbf{Statistical Optimality In Various Regimes.} In the regime where $m$ is known (so $M = m$) and REO is satisfied with $\alpha = 0$, the guarantee of Theorem~\ref{alg:main} is optimal within a single logarithmic factor. To see this, note that in the RHS of Eq.~\eqref{eqn:thm_upper_main_eqn}, the $\numact m$ term cannot be improved due to Proposition 1 of Malik et al.~\citep{malik22}, and the $\sqrt{\numact T}$ term is of course tight due to the classical sMAB lower bound. Moreover, when $m$ is known and REO is satisfied with $\alpha = \bigtheta \left( \sqrt{mK / T} \right)$, then the proof of Theorem 2 of Malik et al.~\citep{malik22} shows that there is a regime of non-trivial $0 < \alpha \ll 1$ where the dependence on $\alpha T$ in Theorem~\ref{alg:main} cannot be improved, and so Theorem~\ref{alg:main} is optimal (within a logarithmic factor). We note that it is unclear whether the $\alpha T$ term in Eq.~\eqref{eqn:thm_upper_main_eqn} is optimal for \emph{all} $\alpha > 0$, and investigating this is an interesting future direction. We defer our investigation into the optimal dependency on $M$, in the regime where $m$ is unknown, to Section~\ref{sec:adaptivity_memory}. \\

\noindent \textbf{Best Arm Identification.} Algorithm~\ref{alg:main} can also be used to identify actions whose eventual loss is near that of $x^\star$. In particular, after $T$ rounds (or $S$ epochs), with probability at least $1 - \delta$ any action $x \in A_{S+1}$ satisfies $h_x(\| w_x \|_1) \leq h_{x^\star}(\| w_{x^\star} \|_1) + 4 C_s = h_{x^\star}(\| w_{x^\star} \|_1) + \bigo \left( \sqrt{K / T} \right)$. \\

\noindent The proof of Theorem~\ref{alg:main} requires some care to ensure optimal dependencies, but the technique is standard, and our contribution is not a novel analysis route. Rather, our contribution is to demonstrate that a classical algorithm for the canonical sMAB can be easily adopted to solve a much more general, and ostensibly more complex, class of problems that are very well motivated in practice. The prior tallying settings that are comparable to our WTB have inherent computational and statistical difficulties (see Section~\ref{sec:related_work} for details). We believe that our formalization of REO and Theorem~\ref{alg:main} is an important identification of well motivated structure that permits statistically and computationally efficient solutions to problems arising in interactive human-centered applications.

\subsection{Adaptivity To Memory Capacity}
\label{sec:adaptivity_memory}
While Algorithm~\ref{alg:main} does not require knowledge of the true memory capacity $m$, it does require an upper bound $M$ on $m$. Theorem~\ref{thm:upper} suggests that the CPR of Algorithm~\ref{alg:main} scales linearly in this input $M$, which is disadvantageous in scenarios where it is difficult to non-trivially upper bound $m$. In general, we desire an algorithm which scales more favorably (or not at all) with the input $M$. For instance, this could be achieved via an algorithm that maintains a confidence interval of the true value $m$, and adaptively queries to refine its estimate of $m$, in order to improve or remove its dependency on $M$. We now show that such an algorithm cannot exist, even in the simpler ``tallying setting'' that is a special case of WTB, and in the case when REO is satisfied with parameter $\alpha = 0$.

To this end, we introduce some notation. For any positive integers $T, M, K$ with $M \leq T$, let $\UTB_{T, M, K}$ denote the set of unweighted tallying bandit problems (i.e., WTB problems where $w_x$ is the all ones vector for each action $x$), that each have horizon length $T$, number of actions $K$, and memory capacity $m \in \{ 1, 2 \dots M \}$, and that satisfy $0$-REO. For any possibly randomized algorithm $\alg$ and any unweighted tallying bandit problem $\tbprob$, let $m_{\tbprob}$ denote the memory capacity of $\tbprob$, and let $\cpReg(\alg, \tbprob)$ denote the expected CPR of algorithm $\alg$ when it is used to solve $\tbprob$. And for a choice of $\epsilon = (\epsilon_1, \epsilon_2, \epsilon_3)$ satisfying $\epsilon_1, \epsilon_2 \in (0, 1)$ and $\epsilon_3 \in [0, \epsilon_2)$, and a choice of function $f: \R^2 \to \R$, let $\algset_{\epsilon, f}$ be the set of algorithms $\alg$ which, when given as input any positive integers $T, M, K$ with $M \leq T$ (and no other information), satisfy for each problem instance $\tbprob \in \UTB_{T, M, K}$ that
\begin{equation}
\label{eqn:main_M_lower}
\E[\cpReg(\alg, \tbprob)] \leq \min \left \{ T/4, f(m_\tbprob, K) \left( T^{1 - \epsilon_1} + T^{\epsilon_3} M^{1 - \epsilon_2} \right) \right \}.
\end{equation}
An algorithm $\alg$ in the set $\algset_{\epsilon, f}$ thus has a benign dependence on $M$ in the following sense. When given any positive integers $T, M, K$ with $M \leq T$, and any problem instance $\tbprob \in \UTB_{T, M, K}$, the algorithm $\alg$ does not a priori know the memory capacity $m_{\tbprob}$ of $\tbprob$, and only knows the upper bound $M$. Nevertheless, the CPR of $\alg$ when solving $\tbprob$ scales \emph{sublinearly} in the bound $M$. Unfortunately, the following result demonstrates that such an algorithm does not exist.
\begin{dhruvthm}
\label{thm:lower_bound_adaptive}
For each $\epsilon$ satisfying $\epsilon_1, \epsilon_2 \in (0, 1)$ and $\epsilon_3 \in [0, \epsilon_2)$, and each function $f$, the corresponding set $\algset_{\epsilon, f}$ is the empty set.
\end{dhruvthm}
\noindent The proof is deferred to Appendix~\ref{app:lower_bound_proof}. The result demonstrates a ``price for adaptivity'' (see, for example,~\citep{locatelli18a} for similar results in a different context), showing that if we only have an upper bound $M$ on the unknown true memory capacity, then any algorithm's CPR cannot be sublinear in both $M$ and $T$. We concretize this via two salient corollaries. The following corollary is stated for the case when we have no non-trivial bound on the memory capacity, or equivalently that $M = T$.
\begin{dhruvcorollary}
\label{corr:one}
Fix any function $f: \R^2 \to \R$. There is no (possibly randomized) algorithm which has expected CPR bounded by $\littleo(T) f(m_{\tbprob}, K)$ for each $\tbprob \in \UTB_{T, T, K}$.
\end{dhruvcorollary}
\noindent The result of Corollary~\ref{corr:one} shows that it is impossible to have an algorithm whose CPR is sublinear in $T$ for all unweighted tallying bandit instances $\tbprob$ with horizon $T$ that satisfy $0$-REO, even at the expense of arbitrarily poor dependence on $m_{\tbprob}, K$. Thus, to obtain a sublinear CPR guarantee of the sort afforded by Theorem~\ref{thm:upper}, it is necessary to have knowledge of some bound $M < T$ on the true memory capacity. The next corollary is stated for when we have a non-trivial bound $M < T$ on the memory capacity.

\begin{dhruvcorollary}
\label{corr:two}
Fix any function $f: \R^2 \to \R$. There is no (possibly randomized) algorithm, which given an input $M$, has expected CPR bounded by $f(m_{\tbprob}, K) \left( \littleo(T) + \littleo(M) \right)$ for each $\tbprob \in \UTB_{T, M, K}$.
\end{dhruvcorollary}

\noindent The result thus shows that we cannot hope to have an algorithm with sublinear dependency on both $M$ and $T$ for all unweighted tallying bandit instances $\tbprob$ with horizon $T$ that satisfy $0$-REO, even at the expense of arbitrarily bad dependence on $K, m_{\tbprob}$. Note that ignoring logarithmic factors, Theorem~\ref{thm:upper} shows that Algorithm~\ref{alg:main} has CPR bounded by $\bigo \left( \sqrt{K T} + K (M +m) \right)$ for each $\tbprob \in \UTB_{T, M, K}$. Combined with our earlier discussion of Theorem~\ref{thm:upper}, Corollary~\ref{corr:two} thus shows that any algorithm must suffer $\bigomega \left( \sqrt{KT} + mK + M \right)$ CPR, highlighting the near optimality of Algorithm~\ref{alg:main} for WTB problems satisfying $0$-REO, even in the regime when we only have an upper bound $M$ on the unknown true memory capacity.

\section{Numerical Results}
\begin{figure*}[t!]
    \centering
    \begin{subfigure}[b]{0.5\textwidth}
        \centering
        \includegraphics[height=1.75in]{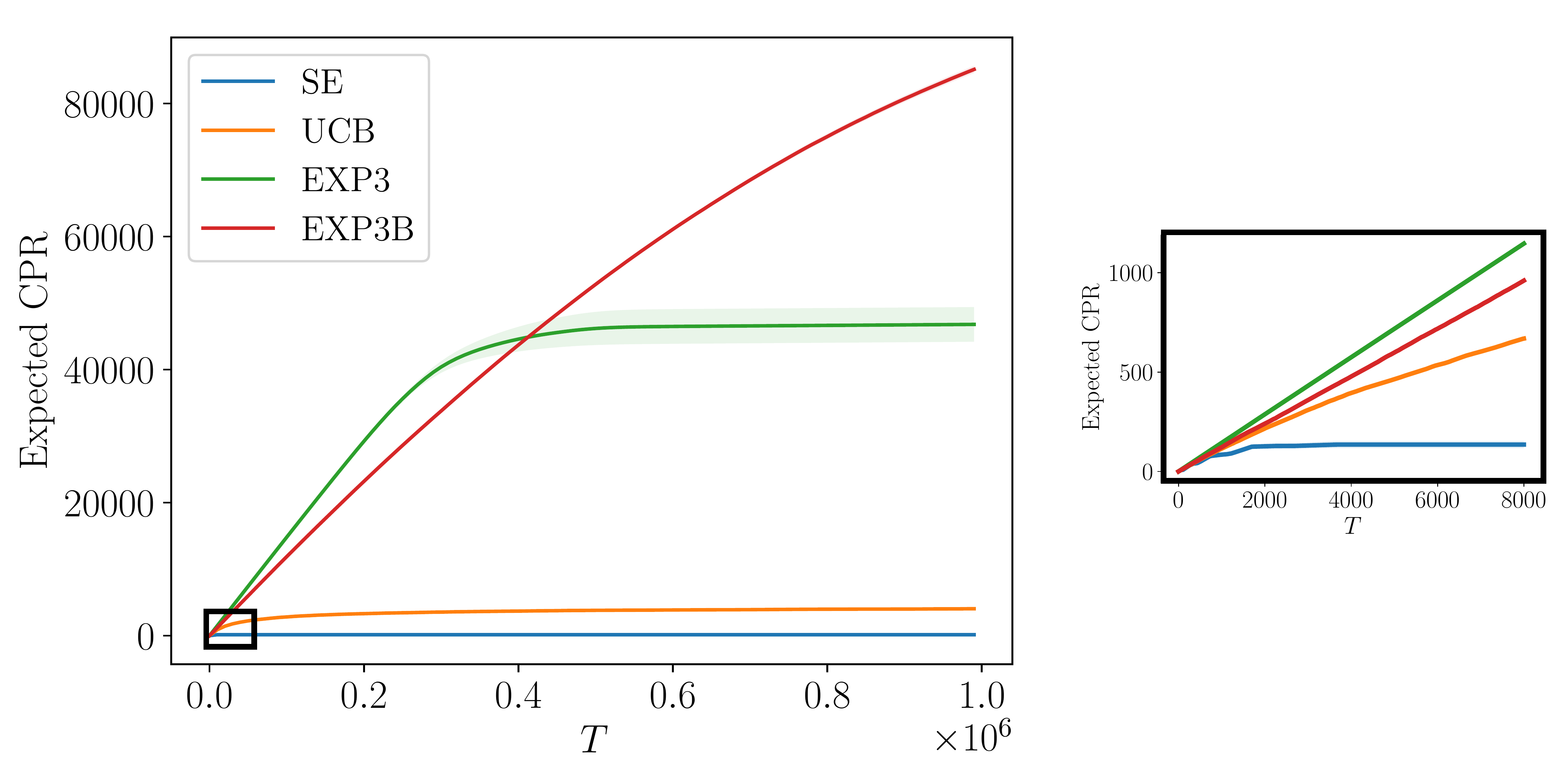}
        \caption{CPR over time horizon $T = 10^6$.}
        \label{fig:synthetic_main_cpr_over_time}
    \end{subfigure}%
    ~ \hfill
    \begin{subfigure}[b]{0.5\textwidth}
        \centering
        \includegraphics[height=1.75in]{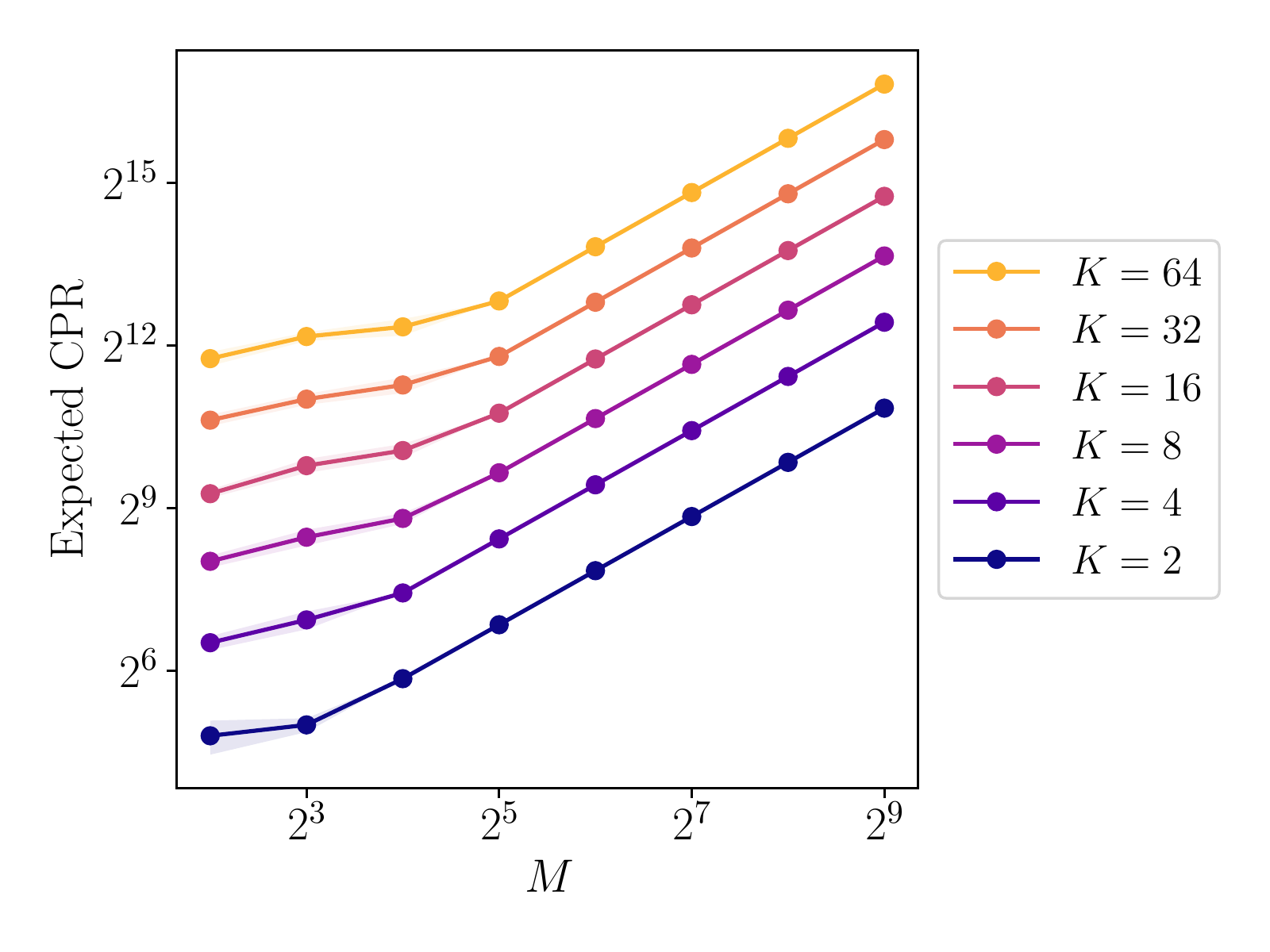}
        \caption{CPR as a function of $M$.}
        \label{fig:synthetic_main_M_scaling}
    \end{subfigure}
    \caption{We plot the expected CPR of each algorithm. In both plots, each datapoint is obtained by averaging over 20 problem instances, and the shaded region depicts $\pm 1$ standard error around the mean. In (a) we fix $K = 5$, $m=3$ and $M=3$. In (b) we fix $m=4$ and $T = 10^6$.}
\end{figure*}
In this section, we evaluate the performance of Algorithm~\ref{alg:main}, which we denote SE, in different domains which are modeled as WTB problems satisfying REO. In each domain, we compare this performance to the following three baselines --- (A) The EXP3 algorithm~\citep{auer02}, which has sublinear traditional regret in our setting (B) The batched version of EXP3 described by Arora et al.~\citep{arora12}, denoted as EXP3B, which has a statistically efficient CPR guarantee in our setting (C) The modified UCB algorithm described in Section~\ref{sec:main_results}.

\subsection{Synthetic Loss Functions on Unweighted Tallying Bandit}
\label{sec:numerical_results_unweighted}
We consider $\{ w_x \}_{x \in \ActSet} = \{ \vec 1 \}$, and fix some $x^\star \in \ActSet$. We define $h_x = 0.5$ for each $x \in \ActSet$, with the modification that $h_{x^\star}(\| w_{x^\star} \|_1) = 0.35$. Hence, the losses are identical, except until we play $x^\star$ at least $m$ times, implying that this instance satisfies $0$-REO. To define the feedback model, we require the random variable $\widetilde{h}_{x}(w_x^{\top} y^{t,x,m})$ has distribution $\text{Bernoulli}(h_{x}(w_x^{\top} y^{t,x,m}))$. When $m=1$, this is a hard instance for sMAB~\citep{slivkins19}, and UCB is of course optimal. For $m > 1$, we note that the UCB variant will perform best in regimes where $h_x \approx h_x( \| w_{x^\star} \|_1)$, since the loss incurred during steps of the $m$-length overhead are nearly equivalent to the eventual losses of repetitively playing an action. Hence, we consider our experimental design to be as favorable to the UCB variant as possible. In Figure~\ref{fig:synthetic_main_cpr_over_time}, we plot the expected CPR of each method over time. As expected, SE outperforms each baseline. In Appendix~\ref{app:numerical_results_unweighted}, we present similar results for other choices of $m, K, M$, and also present results for a problem where $\alpha$-REO is satisfied with $\alpha > 0$. Separately, we study the deterioration of the performance of SE as a function of its input $M$, for the same fixed $m, T, K$. In Figure~\ref{fig:synthetic_main_M_scaling}, we observe that the CPR of SE is at most a linear function of $M$, as one would expect from Theorem~\ref{thm:upper}. However, we also see that in several cases, the scaling is sublinear and hence better than the worst case linear scaling predicted by Theorem~\ref{thm:upper}.

\begin{figure*}[t!]
    \centering
    \begin{subfigure}[b]{0.5\textwidth}
        \centering
        \includegraphics[height=1.75in]{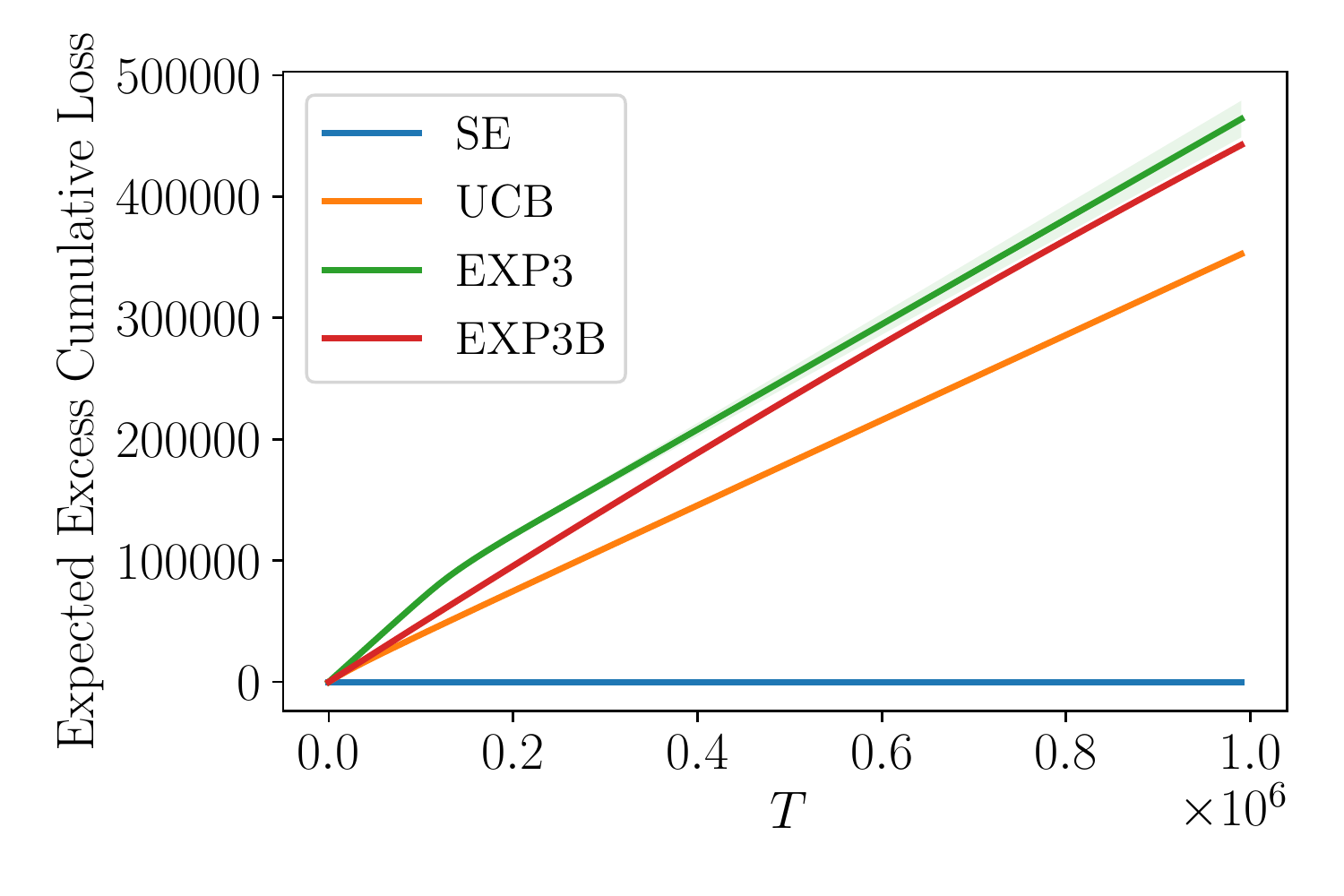}
        \caption{Excess loss in WTB with $\{ w_x \}_{x \in \ActSet} \neq \{ \vec 1 \}$.}
        \label{fig:weighted_tb_correct}
    \end{subfigure}%
    ~ \hfill
    \begin{subfigure}[b]{0.5\textwidth}
        \centering
        \includegraphics[height=1.75in]{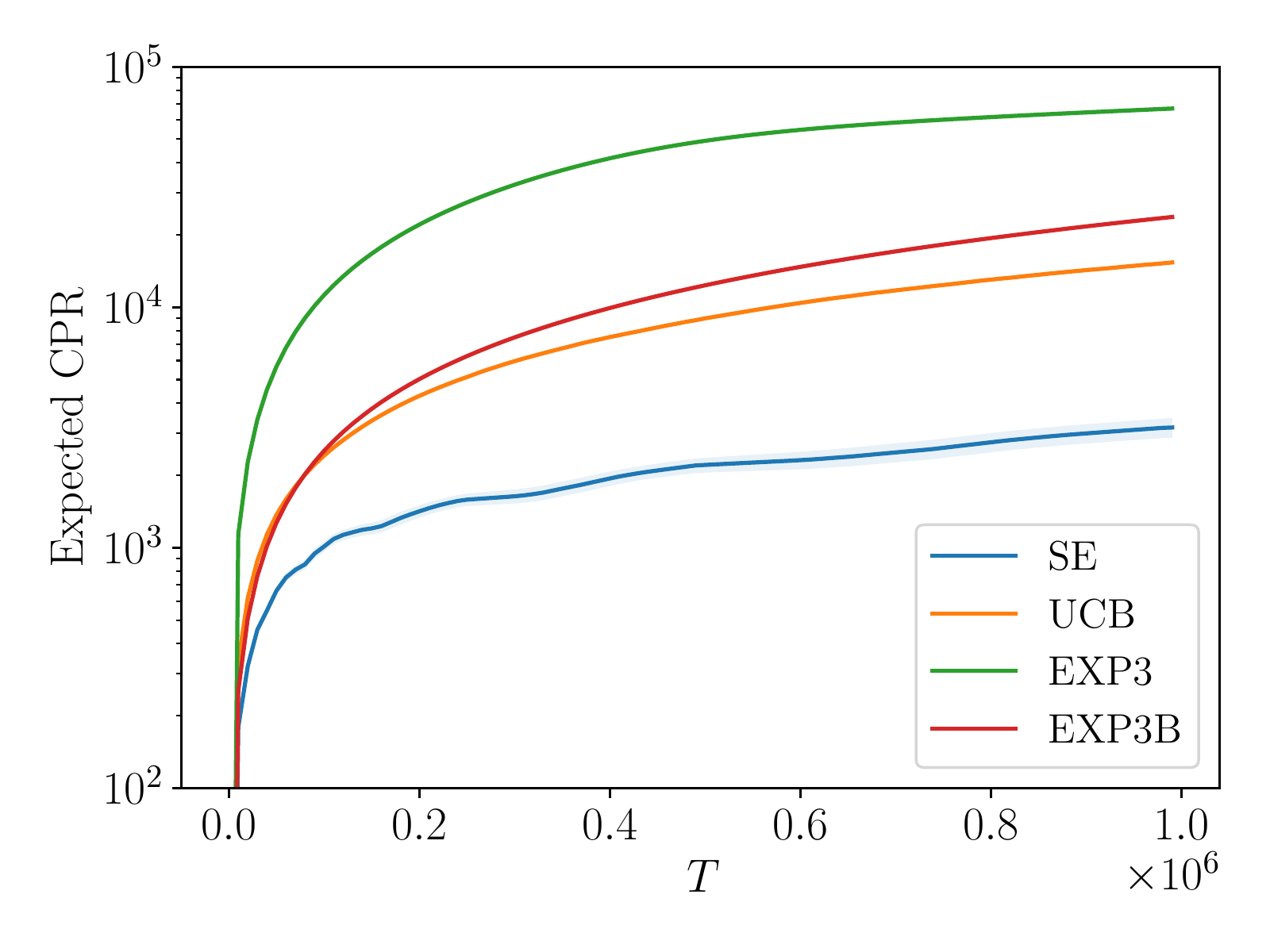}
        \caption{CPR in darts tournament.}
        \label{fig:darts_correct}
    \end{subfigure}
    \caption{In (a), we plot as a function of time the expected cumulative loss of each algorithm in excess of that of SE, in the WTB instance where $\{ w_x \}_{x \in \ActSet} \neq \{ \vec 1 \}$ described in Section~\ref{sec:numerical_results_weighted_tb}, with $K = 5$, $m = 4$ and $M = 4$. In (b), we plot as a function of time the expected CPR of each algorithm in the simulated darts tournament described in Section~\ref{sec:numerical_results_darts}, and truncate the $y$-axis below $10^2$ for illustrative purpose. In both (a) \& (b), data is obtained by averaging over 20 problem instances, and the shaded region depicts $\pm 1$ standard error around the mean.}
\label{fig:weighted_tb_and_darts}
\end{figure*}

\subsection{Synthetic Loss Functions on Weighted Tallying Bandit}
\label{sec:numerical_results_weighted_tb}
We now consider a WTB problem satisfying REO where $\{ w_x \}_{x \in \ActSet} \neq \{ \vec 1 \}$. We relegate the discussion of the precise loss functions used to Appendix~\ref{app:numerical_results_weighted_tb}. Since the optimal policy is difficult to compute for this problem, the CPR is also difficult to compute. So in lieu of the CPR, we plot the expected cumulative loss of each algorithm in excess of SE's loss (hence the CPR at any time is obtained by applying a constant shift to each algorithm's excess loss). The results are shown in Figure~\ref{fig:weighted_tb_correct}, and demonstrate the superiority of our method over the baselines.

\subsection{Simulated Dart Throwing Tournament}
\label{sec:numerical_results_darts}
Motivated by prior work showing the existence of a calibration period in motor tasks~\citep{adams61, phatak20, wunderlich20}, we simulate a simplified dart throwing tournament with $K = 20$ players. As discussed in Section~\ref{sec:prob_form_reo}, Wunderlich et al.~\citep{wunderlich20} show that while a player's first toss is uncalibrated and not necessarily indicative of their subsequent performance, in immediately subsequent tosses the performance calibrates, stabilizes and is better than that of the first toss. We model each (random) instance of the tournament as a WTB with $m=2$ and arbitrary $w$, where each player $x \in \ActSet$ has expected loss function sampled from $h_x(w_{x, 1}) \sim \text{Unif}[0.68, 0.72]$ and $h_x(\| w_x \|_1) \sim \text{Unif}[0.58, 0.62]$. We obtained the bounds for these distributions from Wunderlich et al.~\citep{wunderlich20}, who showed that most players' average performance was concentrated in these intervals. To define the feedback model, we require the random variable $\widetilde{h}_{x}(w_x^{\top} y^{t,x,m})$ has distribution $\text{Bernoulii}(h_{x}(w_x^{\top} y^{t,x,m}))$. While our experimental design eschews some real world subtleties that may occur while throwing darts (for instance, missing a throw might affect the player's confidence on the next throw), we believe that it is a reasonable preliminary model for the calibration period required to throw darts optimally. In Figure~\ref{fig:darts_correct}, we plot the CPR of each method over time. As expected, SE performs significantly better than each baselines.

\subsection{Simulated F1 Tournament}
\label{sec:numerical_results_f1}
In a Formula One (F1) tournament, the goal is to discover the fastest driver out of a set of $K$ drivers. Each driver in the tournament must complete a number of laps. We simulate a modified version of an F1 tournament, where $K = 2$, and at each timestep we pick one of the two drivers to complete a lap (i.e., only a single driver can be on the track at any given timestep). After a lap is completed, we observe (a stochastic realization of) the driver's lap time.

Notably, a driver's lap time depends on the number of laps they have previously completed. To demonstrate this, we utilize F1 lap time data from 1950 to 2022~\citep{f1dataset} to fit a probabilistic lap time model for each F1 driver (details of our probabilistic model and data processing are provided in Appendix~\ref{app:f1_tournament}). In Figure~\ref{fig:racing_reo_holds}, we illustrate our probabilistic model of lap times for a typical driver pair, and show that their lap times tend to decrease as the lap index increases. There are several plausible reasons for this; for instance, the mass of the driver's vehicle decreases with fuel consumption, and the driver's calibration to the race track improves. We thus argue that the sMAB is a poor model for our F1 tournament. Instead it is better modeled as a WTB problem satisfying REO, and our tournament's goal is to discover the driver with the fastest calibrated lap time, which we only observe after repeated exposure.

We simulate multiple instances of our modified F1 tournament, each with $K=2$. The two drivers for each instance are chosen such that their calibrated performance is difficult to distinguish (details in Appendix~\ref{app:f1_tournament}) Here, we present results for a single instance. The results for other instances are presented in Appendix~\ref{app:f1_tournament}. For this instance, we use the probabilistic model depicted in Figure~\ref{fig:racing_reo_holds} to a create a WTB problem with $K = 2$ and $m = 10$. We maintain a tally of the number of times each driver was chosen in the prior $m$ timesteps. The loss associated with picking a driver is governed by the distribution parameterized by our fitted probabilistic model. In particular, if we pick driver $x$ and we have picked them $y$ times in the last $m$ timesteps, then the instantaneous loss is sampled from the distribution parameterized by our fitted probabilistic model for driver $x$ at lap index $y$. Note that in this setting, one has $o(T)$ CPR if and only if one plays the worse driver $o(T)$ many times. In Figure~\ref{fig:racing_cpr}, we plot each method's CPR over time for this tournament instance, showing that SE outperforms the baselines. Similar results are observed for the other tournament instances shown in Appendix~\ref{app:f1_tournament}.

\begin{figure*}[t!]
    \centering
    \begin{subfigure}[b]{0.5\textwidth}
        \centering
        \includegraphics[height=1.75in]{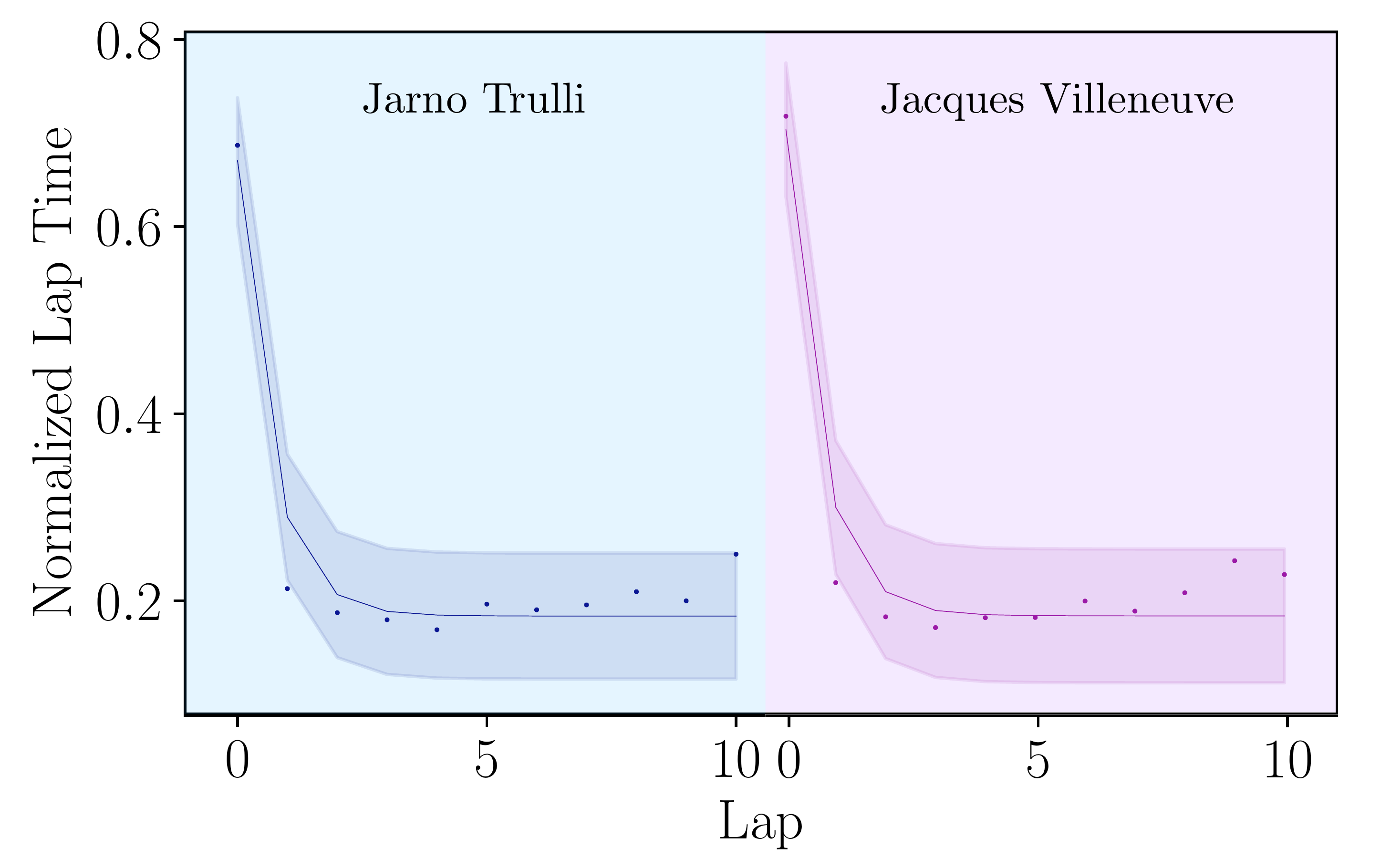}
        \caption{Illustration of probabilistic lap time model.}
        \label{fig:racing_reo_holds}
    \end{subfigure}%
    ~ \hfill
    \begin{subfigure}[b]{0.5\textwidth}
        \centering
        \includegraphics[height=1.75in]{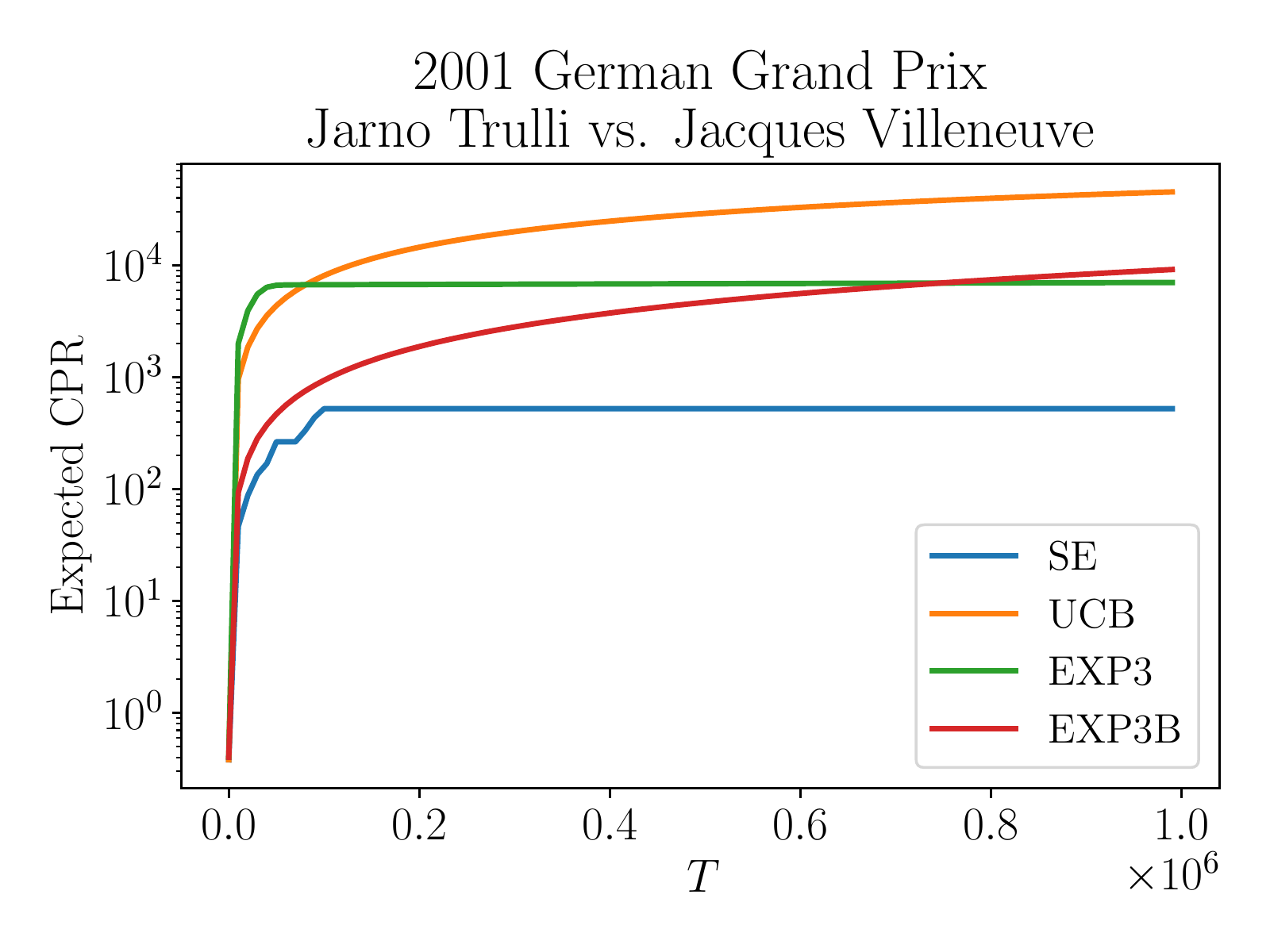}
        \caption{CPR over time horizon $T = 10^6$.}
        \label{fig:racing_cpr}
    \end{subfigure}
    \caption{In (a), we depict our fitted probabilistic lap time model for two drivers in the 2001 German Grand Prix. Our probabilistic model parameterizes a distribution over lap times for each lap index from 1 to 10. The solid line depicts the mean of this distribution, for each lap index. The shaded region contains $\pm 2$ standard deviations of this distribution, centered around the distribution's mean. The dotted points are the actual lap times. Note that all the lap times are normalized, so that each lap time lies in the interval $[0,1]$ (see Appendix~\ref{app:f1_tournament} for details). In (b), we plot as a function of time the expected CPR of each algorithm. Data is obtained by averaging over 20 problem instances, each with $K = 2$, $m = 10$, $M = 10$ and $T = 10^6$, and the shaded region depicts $\pm 1$ standard error around the mean.}
\end{figure*}

\section{Related Work}
\label{sec:related_work}
\textbf{Tallying Settings with $m=T$.} A significant thrust of prior work studies tallying settings that are special cases of WTB with $\{ w_x \}_{x \in \ActSet} = \{ \vec 1 \}$, and require that $m=T$~\citep{heidari16, levine17, seznec19, seznec20, lindner21, metelli22}. Of course, to ensure tractability they enforce various additional types of assumptions, typically in the form of monotonicity on the $\{ h_x \}_{x \in \ActSet}$ functions. Results here do not apply to the case when $m < T$, because $m < T$ causes complications in the design of algorithms since an action's loss ``resets'' if it is not played. Since our paper is primarily motivated by applications where $m < T$, we do not view these works as directly comparable to ours. Nevertheless, we note that upto an additive factor in $mK$ and a logarithmic factor, the CPR guarantees in all these works generally scale less favorably than the rates provided by our Theorem~\ref{thm:upper}. \\

\noindent \textbf{Tallying Settings with $m < T$.} A different body of prior work studies tallying settings that are special cases of WTB with $\{ w_x \}_{x \in \ActSet} = \{ \vec 1 \}$, and like us, they are motivated by applications where $m < T$~\citep{awasthi22, malik22}. These settings are more comparable to ours, since they do not enforce that $m = T$. The tallying bandit~\citep{malik22} makes no assumptions beyond $\{ w_x \}_{x \in \ActSet} = \{ \vec 1 \}$, and here any algorithm must suffer $\bigomega \left( \sqrt{mKT} \right)$ CPR. They adapt successive elimination, and our algorithm is heavily inspired by theirs. The congested bandit~\citep{awasthi22} specializes the tallying bandit by requiring that the $\{ h_x \}_{x \in \ActSet}$ functions are increasing, and even here the best known upper bounds scale as $\bigo \left( \sqrt{mKT} \right)$. The best CPR bounds of these settings thus seem to scale multiplicatively with $m$. Moreover, the computational complexities of the best known algorithms in these settings scale exponentially in $T, m$ respectively. By contrast, our REO condition allows for a computationally efficient algorithm achieving a statistically optimal CPR guarantee that is only additive in $m$. \\

\noindent \textbf{Related Non-Tallying Settings.} A massive body of work studies various settings where the loss of each action evolves over time in some structured fashion, for instance, according to some stochastic process or according to the number of timesteps since the action was last played~\citep{whittle81, moulines11, tekin12, besbes14, bouneffouf16, immorlica18, basu19, pike-burke19, cella20, cortes20, laforgue21}. The models for the evolution of loss in all these works are different than our (weighted) tallying setting. \\

\noindent \textbf{Policy Regret.} Many works study policy regret against generic $m$-memory bounded adversaries, and their algorithms apply to our setting~\citep{arora12, cesa-bianchi13, dekel14, arora18, mohri18}. However, these results ignore the special weighted tallying structure that we consider, and a direct application would result in a suboptimal CPR bound that is worse than our Theorem~\ref{thm:upper}.

\section{Discussion}
\label{sec:discussion}
In this paper, we formulated the Weighted Tallying Bandit, which generalizes prior tallying settings so that the loss at a timestep is a function of a weighted summation of the number of times it was recently played. To ensure tractability in this challenging setting, we introduced the Repeated Exposure Optimality condition, which we motivated via human-centered applications where one's best performance requires a calibration period before it stabilizes. We showed that a simple modification of the classical successive elimination algorithm achieves an optimal complete policy regret guarantee (upto a single logarithmic factor), and our numerical results demonstrate its practicality, scalability and superiority over alternative baselines. Finally, we showed that while our algorithm required as input a non-trivial upper bound $M < T$ on $m$, any algorithm that has sublinear CPR requires such an input, and that our method's dependency on this input $M$ is optimal. Collectively, this implies our algorithm's CPR is optimal for WTB problems satisfying $0$-REO.

We acknowledge our work has certain limitations. From a theoretical perspective, while there is a regime of non-trivial $\alpha = \bigtheta \left( \sqrt{mK / T} \right)$ (as discussed in Section~\ref{sec:upper_bound}) where Theorem~\ref{thm:upper} is optimal and its dependence on $\alpha$ cannot be improved, it is unclear whether this dependence is optimal for \emph{all} values of $\alpha$. Investigating this is an interesting direction for future work.

More practically, a limitation of our WTB and REO setting is that while it is a reasonable step to model the calibration period that arises before we see true best performance, it fails to model many other subtleties that arise in the human-centered domains that motivate our work to begin with. For instance, it is plausible that in more strenuous tasks, after repeatedly performing a task for a long $m < T$, a calibrated individual may begin to experience fatigue. In this case, the best model for losses associated with repeatedly playing an action would be an initial period where the individual calibrates and their performance improves to ``sweet-spot'', but then their performance eventually deteriorates as fatigue accumulates (this model is analogous to the $m=T$ setting considered by~\citep{lindner21}). Our setting can handle the initial period of calibration and performance improvement, but cannot handle the latter phase of deterioration. We believe that exploring algorithms that can handle this is a key direction for future work.

A different way to improve our model, is by studying more general settings where each action $x$ is associated with its own memory capacity $m_x$ (instead of a single $m$ for all actions), or where the memory capacity of the problem changes with time. Can we design algorithms that intelligently adapt to such complexities?

Our concrete motivating examples for the REO condition are primarily derived from the psycho-physiological literature on the warm-up decrement phenomenon. Nevertheless, we generally expect that it may additionally apply to other interactive settings such as recommender systems. For instance, it is plausible that a user needs to see a type of content multiple times before she decides her preferences for it, but if she is not shown the content for a while, then she forgets its details and requires another exploratory calibration period to re-affirm her preference for that content relative to more recent recommendations. In this case, the goal is to explore the \emph{eventual} preferences of the user, and then repetitively select the item that the user \emph{eventually} prefers the most. Such a setting would be a reasonable application for WTB with REO. Quantitatively analyzing recommender system data, and showing that such settings exist and can be modeled well by WTB with REO, is a very interesting direction for future work with potentially broad practical impact.

\subsection*{Acknowledgements}
This material is based upon work supported by the National Science Foundation Graduate Research Fellowship Program under Grant No. DGE1745016. This work is also supported in part by the National Science Foundation AIIRA Institute Grant 20216702135329. Any opinions, findings, and conclusions or recommendations expressed in this material are those of the authors and do not necessarily reflect the views of the National Science Foundation.

\appendix

\section{Analysis of Algorithm~\ref{alg:main}}
\label{app:upper}

\noindent In this section, we analyze the complete policy regret of Algorithm~\ref{alg:main}, and prove Theorem~\ref{thm:upper}. As discussed in Section~\ref{sec:main_results}, our analysis is overall rather standard, although we require a careful choice of parameters to ensure optimal dependencies in the final result. Thus, many of the computations closely follow those of Malik et al.~\citep{malik22}, but we nevertheless provide the entire argument for the sake of completeness. Before we formally prove Theorem~\ref{thm:upper}, we first introduce the function $\mu: \ActSet \to [0,1]$ which will be useful for our proofs. For any action $x \in \ActSet$ let us define $\mu(x)$ as
$$
\mu(x) = h_x( \| w_x \|_1 ).
$$
And for each epoch $s \in \{ 1, 2 \dots S \}$ in the outer loop of Algorithm~\ref{alg:main}, define $T_s = 2 \vert A_s \vert n_s$, where $A_s, n_s$ are defined in Algorithm~\ref{alg:main}. With this definition in hand, we are now in a position to formally prove Theorem~\ref{thm:upper}.

\subsection{Proof of Theorem~\ref{thm:upper}}

\noindent For any policy $\pi$, which is a length $T$ deterministic sequence of actions, let $\ell_t(\pi)$ denote the expected loss suffered at timestep $t$ while playing $\pi$. Define the policy $\pi^\star$ as
$$
\pi^\star \in \argmin_{\pi \in \ActSet^T} \sum_{t=1}^T \ell_t(\pi),
$$
so that $\pi^\star$ is an optimal policy (i.e., a policy that suffers the minimum cumulative expected loss). Note that the definition of an $(m,w,h)$-weighted tallying bandit ensures that for any timestep $t$, there exists an action $x \in \ActSet$ and $y \in \{ 1 \} \times \{ 0,1 \}^{m-1}$ such that $\ell_t(\pi^\star) = h_x(w_x^{\top} y)$. Thus, the $\alpha$-REO condition ensures that
$$
\mu(x^\star) = h_{x^\star}(\| w_{x^\star} \|_1) \leq h_x(w_x^{\top} y) + \alpha = \ell_t(\pi^\star) + \alpha.
$$
In particular, this implies that
\begin{equation}
\label{eq:approx_main_loss_by_best_arm}
\sum_{t=1}^T \ell_t(\pi^\star) \geq T \mu(x^\star) - \alpha T.
\end{equation}
\noindent Now let $\ell^s$ denote the loss experienced in epoch $s \in \{ 1, 2 \dots S \}$ of Algorithm~\ref{alg:main}. The following lemma bounds the cumulative loss of Algorithm~\ref{alg:main} relative to $T \mu(x^\star)$.

\begin{dhruvlemma}
\label{lem:main_regret_helper}
With probability at least $1 - \delta$, the total loss of Algorithm~\ref{alg:main} relative to $T \mu(x^\star)$ can be upper bounded as
$$
\sum_{s=1}^S \ell^s - T \mu(x^\star) \leq 4 \numact \mup + \numact m \log(T) + 800 \sqrt{\numact T \log \left( \frac{2 \numact \log(T)}{\delta} \right)}.
$$
\end{dhruvlemma}
\noindent The proof of this Lemma~\ref{lem:main_regret_helper} is provided in Appendix~\ref{app_proof:lem:main_regret_helper}. With the result of Lemma~\ref{lem:main_regret_helper} in hand, we now utilize it to prove Theorem~\ref{thm:upper} as follows. Note via Eq.~\eqref{eq:approx_main_loss_by_best_arm} and Lemma~\ref{lem:main_regret_helper} that the complete policy regret $\cpReg$ of Algorithm~\ref{alg:main} satisfies
\begin{align*}
\cpReg &= \sum_{s=1}^S \ell^s - \sum_{t=1}^T \ell_t(\pi^{\star}) \\
&\leq \sum_{s=1}^S \ell^s - T \mu(x^\star) + \alpha T \\
&\leq 4 \numact \mup + \numact m \log(T) + 800 \sqrt{\numact T \log \left( \frac{2 \numact \log(T)}{\delta} \right)} + \alpha T.
\end{align*}
This completes the proof of Theorem~\ref{thm:upper}. \hfill \qeddhruv

\subsection{Proof of Lemma~\ref{lem:main_regret_helper}}
\label{app_proof:lem:main_regret_helper}
\noindent To facilitate the proof, we require the following critical lemma, which bounds the loss incurred by Algorithm~\ref{alg:main} in each epoch $s \in \{ 1, 2 \dots S\}$. For the statement of the following lemma, note that completing any epoch $s \in \{ 1, 2 \dots S \}$ takes a total of $T_s = 2 \vert A_s \vert n_s$ timesteps.
\begin{dhruvlemma}
\label{lem:loss_each_s}
With probability at least $1 - \delta$, we have simultaneously for each epoch $s \in \{ 2, 3 \dots S \}$ that the total loss relative to $T_s \mu(x^\star)$ is bounded as
$$
\ell^s - T_s \mu(x^\star) \leq \vert A_s \vert \left( m + 4 (2 n_s - m) C_{s-1} \right).
$$
\end{dhruvlemma}

\noindent The proof of this Lemma~\ref{lem:loss_each_s} is provided in Appendix~\ref{app_proof:lem:loss_each_s}. Observe that by the result of Lemma~\ref{lem:loss_each_s}, we are guaranteed with probability at least $1 - \delta$ that
\begin{equation}
\label{eqn:help4}
\begin{aligned}
\sum_{s=1}^S \ell^s - T \mu(x^\star) &= \sum_{s=1}^S \left( \ell^s - T_s \mu(x^\star) \right) \\
&\leq 4 \numact \mup + \sum_{s=2}^S \left( \ell^s - T_s \mu(x^\star) \right) \\
&\leq 4 \numact \mup + \sum_{s=2}^S \vert A_s \vert \left( m + 4 (2 n_s - m) C_{s-1} \right) \\
&\leq 4 \numact \mup + S \numact m + 8 \sum_{s=2}^S \vert A_s \vert n_s C_{s-1}.
\end{aligned}
\end{equation}
Recall the definitions $n_s = \numact \mup 2^s / \vert A_s \vert$ and $T_s = 2 \vert A_s \vert n_s$ provided in Algorithm~\ref{alg:main}. Also note that
$$
C_{s-1} = \sqrt{ \frac{32}{n_{s-1}} \log \left( \frac{2 \numact S}{\delta} \right) } = \sqrt{ \frac{32}{n_s \vert A_s \vert / (2 \vert A_{s-1} \vert)} \log \left( \frac{2 \numact S}{\delta} \right) } = \sqrt{ \frac{64 \vert A_{s-1} \vert}{n_s \vert A_s \vert} \log \left( \frac{2 \numact S}{\delta} \right) }.
$$
Substituting the above relations into the final term on the RHS of Eq.~\eqref{eqn:help4}, we get that
\begin{align*}
8 \sum_{s=2}^S \vert A_s \vert n_s C_{s-1} &= 8 \sum_{s=2}^S \vert A_s \vert n_s \sqrt{ \frac{64 \vert A_{s-1} \vert}{n_s \vert A_s \vert} \log \left( \frac{2 \numact S}{\delta} \right) }\\
&= 8 \sqrt{ \log \left( \frac{2 \numact S}{\delta} \right) } \sum_{s=2}^S \vert A_s \vert n_s \sqrt{ \frac{64 \vert A_{s-1} \vert}{n_s \vert A_s \vert} } \\
&= 8 \sqrt{ \log \left( \frac{2 \numact S}{\delta} \right) } \sum_{s=2}^S \vert A_s \vert \sqrt{n_s} \sqrt{ \frac{64 \vert A_{s-1} \vert}{\vert A_s \vert} } \\
&= 8 \sqrt{ \log \left( \frac{2 \numact S}{\delta} \right) } \sum_{s=2}^S \vert A_s \vert \sqrt{\frac{\numact \mup 2^s}{\vert A_s \vert}} \sqrt{ \frac{64 \vert A_{s-1} \vert}{\vert A_s \vert} } \\
&= 8 \sqrt{ \log \left( \frac{2 \numact S}{\delta} \right) } \sum_{s=2}^S \sqrt{\numact \mup 2^s} \sqrt{ 64 \vert A_{s-1} \vert } \\
&\leq 8 \sqrt{ \log \left( \frac{2 \numact S}{\delta} \right) } \sum_{s=2}^S K \sqrt{ \mup 2^s} \sqrt{ 64 } \\
&\leq 800 \sqrt{ \log \left( \frac{2 \numact S}{\delta} \right) } \numact \sqrt{ \mup } 2^{S/2}.
\end{align*}

\noindent Now recall from Algorithm~\ref{alg:main} the definition of $S = \log_2 \left( \frac{T}{4 \numact \mup} + 1 \right)$. Substituting this into the equation above, we get that
\begin{equation}
\label{eqn:help5}
\begin{aligned}
8 \sum_{s=2}^S \vert A_s \vert n_s C_{s-1} &\leq 800 \sqrt{ \log \left( \frac{2 \numact S}{\delta} \right) } \numact \sqrt{ \mup } 2^{S/2} \\
&= 800 \sqrt{ \log \left( \frac{2 \numact S}{\delta} \right) } \numact \sqrt{ \mup } \sqrt{\frac{T}{4 \numact \mup} + 1} \\
&\leq 800 \sqrt{ \log \left( \frac{2 \numact S}{\delta} \right) } \numact \sqrt{ \mup } \sqrt{\frac{T}{\numact \mup}} \\
&= 800 \sqrt{ \log \left( \frac{2 \numact S}{\delta} \right) } \sqrt{\numact} \sqrt{T}.
\end{aligned}
\end{equation}
Combining Eq.~\eqref{eqn:help4} with Eq.~\eqref{eqn:help5} and using the upper bound $S \leq \log(T)$ yields the result. \hfill \qeddhruv

\subsection{Proof of Lemma~\ref{lem:loss_each_s}}
\label{app_proof:lem:loss_each_s}
\noindent To facilitate the proof, we leverage the following critical lemma, which bounds the gap of the value $\mu(x)$ of each action $x \in A_s$ versus $\mu(x^\star)$.

\begin{dhruvlemma}
\label{lem:opt_gap}
The event
$$
\cap_{s=2}^S \cap_{x \in A_s} \left \{ \mu(x) - \mu(x^\star) \leq 4 C_{s-1} \right \},
$$
occurs with probability at least $1 - \delta$.
\end{dhruvlemma}

\noindent The proof of this Lemma~\ref{lem:opt_gap} is provided in Appendix~\ref{app_proof:lem_opt_gap}. Let us now return to the main proof. For any epoch $s > 1$ and any action $x \in A_s$, let $\ell^{s}$ denote the total loss experienced in epoch $s$ of Algorithm~\ref{alg:main} while executing the action $x$ for $2 n_s$ times. Hence we have $\ell^s = \sum_{x \in A_s} \ell^{sx}$. \\

\noindent Note that within a single epoch $s > 1$, for each $x \in A_s$ we execute $x$ for $2 n_s$ times. For the latter $2 n_s - m$ times that $x$ is executed, action $x$ has been played $m$ times in the previous $m$ timesteps. Hence, for the latter $2 n_s - m$ times that $x$ is executed, the expected loss of playing the action $x$ is $h_x(m) = \mu(x)$. Thus, we have that
$$
\ell^{sx} \leq m + (2 n_s - m) \mu(x).
$$
Hence, for each $s > 1$ and $x \in A_s$, we can use Lemma~\ref{lem:opt_gap} to upper bound 
\begin{align*}
\ell^{sx} - 2 n_s \mu(x^\star) &\leq m + (2 n_s - m) \mu(x) - 2 n_s \mu(x^\star) + m \mu(x^\star) \\
&\leq m + (2 n_s - m) \left( \mu(x) - \mu(x^\star) \right) \\
&\leq m + 4 (2 n_s - m) C_{s-1}.
\end{align*}
This bound holds uniformly for each $x \in A_s$. Recalling that $T_s = 2 \vert A_s \vert n_s$, we hence have that
\begin{align*}
\ell^s - T_s \mu(x^\star) &= \sum_{x \in A_s} \ell^{sx} - 2 \vert A_s \vert n_s \mu(x^\star) \\
&= \sum_{x \in A_s} \left( \ell^{sx} - 2 n_s \mu(x^\star) \right) \\
&\leq \sum_{x \in A_s} \left( m + 4 (2 n_s - m) C_{s-1} \right) \\
&= \vert A_s \vert \left( m + 4 (2 n_s - m) C_{s-1} \right).
\end{align*}
This completes the proof. \hfill \qeddhruv

\subsection{Proof of Lemma~\ref{lem:opt_gap}}
\label{app_proof:lem_opt_gap}
To facilitate the proof, we require the following two critical helper results. The first result bounds the error incurred when estimating $\mu(x)$ via the stochastic realizations $\{ \widetilde{h}_x(m)_{s,k} \}$. The second result shows that while running Algorithm~\ref{alg:main}, which is based on successive elimination of inferior actions over epochs $s \in \{ 1, 2 \dots S \}$, at any epoch $s$ we never eliminate $x^\star$ from our set $A_s$ of feasible actions.

\begin{dhruvlemma}
\label{lem:mu_conc}
Fix any $s \in \{ 1, 2 \dots S \}$, and let $B_s$ denote the event that for all actions $x \in A_s$ we simultaneously have that
$$
\left \vert \widehat{\mu}_s(x) - \mu(x) \right \vert \leq C_s.
$$
Then $B_s$ occurs with probability at least $1 - \delta/S$.
\end{dhruvlemma}

\begin{dhruvlemma}
\label{lem:pi_star_in_As}
The event $\cap_{s=1}^S B_s$, where the event $B_s$ is defined in Lemma~\ref{lem:mu_conc}, implies the event that
$$
x^\star \in \cap_{s=1}^S A_s \text{ and } \cap_{s=1}^S \left \{ 0 \leq \widehat{\mu}_s(x^\star) - \widehat{\mu}_s(\widehat{x}_s) \leq 2 C_s \right \}.
$$
\end{dhruvlemma}

\noindent The proofs of Lemma~\ref{lem:mu_conc} and Lemma~\ref{lem:pi_star_in_As} are provided in Appendix~\ref{app_proof:lem:mu_conc} and Appendix~\ref{app_proof:lem:pi_star_in_As} respectively. \\

\noindent Let us now return to the proof. By the result of Lemma~\ref{lem:mu_conc} and a union bound, the event $\cap_{s=1}^S B_s$ occurs with probability at least $1- \delta$. Furthermore, the result of Lemma~\ref{lem:pi_star_in_As} shows that the event $\cap_{s=1}^S B_s$ implies the event
\begin{equation}
\label{eqn:help3}
x^\star \in \cap_{s=1}^S A_s.
\end{equation}
So on the event $\cap_{s=1}^S B_s$, note that for any $s > 1$ and any action $x \in A_s$ we have
\begin{align*}
\mu(x) - \mu(x^\star) &\overset{(i)}{\leq} \widehat{\mu}_{s-1}(x) - \mu(x^\star) + C_{s-1} \\
&\overset{(ii)}{\leq} \widehat{\mu}_{s-1}(\widehat{x}_{s-1}) - \mu(x^\star) + 3C_{s-1} \\
&\overset{(iii)}{\leq} \widehat{\mu}_{s-1}(x^\star) - \mu(x^\star) + 3C_{s-1} \\
&\overset{(iv)}{\leq} \mu(x^\star) - \mu(x^\star) + 4C_{s-1} \\
&= 4 C_{s-1},
\end{align*}
where step $(i)$ follows from Lemma~\ref{lem:mu_conc}, step $(ii)$ follows from the definition of $A_s$ and the fact that $x \in A_s$, step $(iii)$ follows from the definition of $\widehat{x}_{s-1}$ and Eq.~\eqref{eqn:help3}, and step $(iv)$ follows again from Lemma~\ref{lem:mu_conc} and Eq.~\eqref{eqn:help3}. This completes the proof. \hfill \qeddhruv

\subsection{Proof of Lemma~\ref{lem:mu_conc}}
\label{app_proof:lem:mu_conc}
Fix any $x \in \ActSet$. Recalling the definition of $C_s$ provided in Algorithm~\ref{alg:main}, Hoeffding's bound~\citep{hoeffding63} ensures that the event
\begin{equation}
\label{eqn:helper1}
\left \vert \widehat{\mu}_s(x) - \mu(x) \right \vert = \left \vert \mu(x) - \frac{1}{n_s} \sum_{k=1}^{n_s} \widetilde{h}_x(m)_{s,k} \right \vert \leq \sqrt{ \frac{32}{n_s} \log \left( \frac{2 \numact S}{\delta} \right) } = C_s,
\end{equation}
occurs with probability at least $1 - \delta/(\numact S)$. Since $\vert A_s \vert \leq \numact$, a union bound then ensures that the above event occurs simultaneously for all $x \in A_s$ with probability at least $1 - \delta/S$. \hfill \qeddhruv

\subsection{Proof of Lemma~\ref{lem:pi_star_in_As}}
\label{app_proof:lem:pi_star_in_As}
\noindent Assume that the event $\cap_{s'=1}^S B_{s'}$ is true. On this event, we prove the lemma by induction on $s$. First we demonstrate the base case of $s=1$, which is that $x^\star \in A_1$ and $0 \leq \widehat{\mu}_1(x^\star) - \widehat{\mu}_1(\widehat{x}_1) \leq 2 C_1$. Then for the inductive step we show that if the event $x^\star \in A_{s-1}$ and $0 \leq \widehat{\mu}_{s-1}(x^\star) - \widehat{\mu}_{s-1}(\widehat{x}_{s-1}) \leq 2 C_{s-1}$ occurs, then we also have that the event 
$$
x^\star \in A_s \text{ and } 0 \leq \widehat{\mu}_{s}(x^\star) - \widehat{\mu}_{s}(\widehat{x}_{s}) \leq 2 C_{s},
$$
is also true. \\

\noindent For the base case, note that by definition we are guaranteed $x^\star \in A_1$. And by the definition of $\widehat{x}_1$, we know that $0 \leq \widehat{\mu}_1(x^\star) - \widehat{\mu}_1(\widehat{x}_1)$. Furthermore, recalling the definition of the event $B_1$ in Lemma~\ref{lem:mu_conc}, on the event $B_1$ we have that
$$
\widehat{\mu}_1(x^\star) - \mu(x^\star) \leq C_1 \text{ and } \mu(\widehat{x}_1) - \widehat{\mu}_1(\widehat{x}_1) \leq C_1.
$$
Putting these equations together and using the fact that $\mu(x^\star) \leq \mu(\widehat{x}_1)$ ensures that
$$
\widehat{\mu}_1(x^\star) - \widehat{\mu}_1(\widehat{x}_1) \leq 2C_1.
$$
This verifies the base case. \\

\noindent For the inductive step, assume that $x^\star \in A_{s-1}$ and $0 \leq \widehat{\mu}_{s-1}(x^\star) - \widehat{\mu}_{s-1}(\widehat{x}_{s-1}) \leq 2 C_{s-1}$ occurs. Then the definition of $A_s$ and the inductive hypothesis directly imply that $x^\star \in A_s$. Hence, it is true by definition of $\widehat{x}_s$ that $0 \leq \widehat{\mu}_s(x^\star) - \widehat{\mu}_s(\widehat{x}_s)$. Then recalling the definition of the event $B_s$ in Lemma~\ref{lem:mu_conc}, on the event $B_s$ we have that
$$
\widehat{\mu}_s(x^\star) - \mu(x^\star) \leq C_s \text{ and } \mu(\widehat{x}_s) - \widehat{\mu}_s(\widehat{x}_s) \leq C_s.
$$
Putting these equations together and using the fact that $\mu(x^\star) \leq \mu(\widehat{x}_s)$ ensures that
$$
\widehat{\mu}_s(x^\star) - \widehat{\mu}_s(\widehat{x}_s) \leq 2C_s.
$$
This verifies the inductive step. As argued earlier, this is sufficient to complete the proof. \hfill \qeddhruv

\section{Proof of Theorem~\ref{thm:lower_bound_adaptive}}
\label{app:lower_bound_proof}
\noindent Assume for the sake of contradiction that the statement is false. Then there exists some $\epsilon$ satisfying the given conditions and some function $f$, such that $\algset_{\epsilon, f}$ is not empty. This implies the existence of an algorithm $\alg$, such that when it is given as input any positive integers $T, K, M$ with $M \leq T$, the algorithm $\alg$ satisfies that
\begin{equation}
\label{eqn:lower_primary_bound}
\E[\cpReg(\alg, \tbprob)] \leq \min \left \{ T/4, f(m_\tbprob, K) \left( T^{1 - \epsilon_1} + T^{\epsilon_3} M^{1 - \epsilon_2} \right) \right \} \text{ for all } \tbprob \in \UTB_{T, M, K}.
\end{equation}
If $\alg$ was a randomized algorithm, then this implies the existence of a deterministic algorithm with the same property. So we can assume without loss of generality that $\alg$ is deterministic. \\

\noindent Fix some integer $K \geq 2$. Pick some sufficiently large $T, M$ such that the following conditions hold simultaneously
\begin{equation}
\label{eqn:lower_bound_conditions}
M < T/4 \text{ and } f(1, K) \left( T^{1 - \epsilon_1} + T^{\epsilon_3} M^{1 - \epsilon_2} \right) < M/2.
\end{equation}
To see these conditions are simultaneously feasible, recall that $\epsilon_1 \in (0, 1)$ and $0 \leq \epsilon_3 < \epsilon_2 < 1$. Let $\gamma = \min \{ \epsilon_1, \epsilon_2 - \epsilon_3 \} > 0$. So if we choose $M = T^{1 - \gamma/2}$, then since this $M$ satisfies $M = T^{1 - \gamma/2} < T$, we have that
\begin{align*}
f(1, K) \left( T^{1 - \epsilon_1} + T^{\epsilon_3} M^{1 - \epsilon_2} \right) &< f(1, K) \left( T^{1 - \epsilon_1} + T^{\epsilon_3} T^{1 - \epsilon_2} \right) \\
&= f(1, K) \left( T^{1 - \epsilon_1} + T^{1 - (\epsilon_2 - \epsilon_3)} \right) \\
&\leq 2 f(1, K) T^{1 - \gamma}.
\end{align*}
So for sufficiently large $T$, we have for this choice of $M = T^{1 - \gamma/2}$ that $M < T/4$ and also that $f(1, K) \left( T^{1 - \epsilon_1} + T^{\epsilon_3} M^{1 - \epsilon_2} \right) < M/2$. This shows that Eq.~\eqref{eqn:lower_bound_conditions} is feasible. \\

\noindent We will now define two unweighted tallying bandit problems, each of which have $K$ actions. Recall that in an unweighted tallying bandit problem with memory capacity $m$, the loss associated with playing an action at a given timestep is fully defined by the number of times that action was played in the last $m$ timesteps. Concretely, assume that in some unweighted tallying bandit problem $\tbprob$, we play action $x$ on the current timestep, and the total number of times it has been played in the last $m$ timesteps (including the current timestep) is $1 \leq y \leq m$. Then there exists a function $h_{\tbprob, x}: \{ 1, 2 \dots m \} \to [0,1]$, such that denote the loss associated with playing this action is given by $h_{\tbprob, x}(y)$. We will use this notation to instantiate the forthcoming unweighted tallying bandit problems. \\

\noindent With this notation in hand, let us instantiate the unweighted tallying bandit problem $\tbprob_A$ with memory length $m_{\tbprob_A} = 1$ as follows. For action $x_1$, we have that $h_{\tbprob_A, x_1} = 1/2$. For action $x_2$, we have that $h_{\tbprob_A, x_2} = 1$. And for every other action $x$, let $h_{\tbprob_A, x} = 1$. We say that whenever the player plays action $x$, the player almost surely observes $h_{\tbprob_A, x}(1)$. Notice that since $m_{\tbprob_A} = 1$, and there is no stochasticity in the observation of losses when we play any action, $\tbprob_A$ is indeed a deterministic multi-armed bandit problem. \\

\noindent Now, we instantiate the tallying bandit problem $\tbprob_B$ with memory length $m_{\tbprob_B} = M$ as follows. For action $x_1$ we define $h_{\tbprob_B, x_1} = 1/2$. For action $x_2$ we define
$$
h_{\tbprob_B, x_2}(y) =
\begin{cases}
1 \text{ if } 1 \leq y < M \\
0 \text{ if } y = M
\end{cases}.
$$
And for every other action $x$, we define $h_{\tbprob_B, x} = 1$. Once again, we enforce that there is no stochasticity in the player's observation of losses when the player plays an action. So the feedback model in $\tbprob_B$ is deterministic. \\

\noindent Let the horizon length for problems $\tbprob_A$ and $\tbprob_B$ be the $T$ chosen as per Eq.~\eqref{eqn:lower_bound_conditions}. Note also that both problem instances satisfy REO with parameter $\alpha = 0$, and so $\tbprob_A, \tbprob_B \in \UTB_{T, M, K}$. So via our assumption of the determinism of $\alg$, via Eq.~\eqref{eqn:lower_primary_bound} and via the fact that $m_{\tbprob_A} = 1$, we have that
\begin{equation}
\label{eqn:lower_bound_true_bound_1}
\cpReg(\alg, \tbprob_A) \leq f(m_{\tbprob_A}, K) \left( T^{1 - \epsilon_1} + T^{\epsilon_3} M^{1 - \epsilon_2} \right) = f(1, K) \left( T^{1 - \epsilon_1} + T^{\epsilon_3} M^{1 - \epsilon_2} \right) < M/2,
\end{equation}
where the final inequality follows due to Eq.~\eqref{eqn:lower_bound_conditions}. And again by our assumption of the determinism of $\alg$ and via Eq.~\eqref{eqn:lower_primary_bound}, we have that
\begin{equation}
\label{eqn:lower_bound_true_bound_2}
\cpReg(\alg, \tbprob_B) \leq T/4.
\end{equation}
When $\alg$ is run on problem $\tbprob_A$, there are 2 cases. Either $\alg$ plays $x_2$ for $M$ times in a row (at some point in its execution for $T$ timesteps while solving $\tbprob_A$) or it does not. \\

\noindent Consider the first case, where $\alg$ plays $x_2$ for $M$ times in a row on problem $\tbprob_A$. Then we have that $\cpReg(\alg, \tbprob_A) \geq M/2$. This is because the optimal strategy for $\tbprob_A$ always plays $x_1$ on each timestep, and so any timestep where $\alg$ plays action $x \neq x_1$ will add $h_{\tbprob_A, x} - h_{\tbprob_A, x_1} = 1 - 1/2 = 1/2$ to the CPR of $\alg$. This is a contradiction to Eq.~\eqref{eqn:lower_bound_true_bound_1}. \\

\noindent Consider the second case, where $\alg$ never plays $x_2$ for $M$ times in a row on problem $\tbprob_A$. Note that the deterministic algorithm $\alg$ can be viewed as a length $T$ sequence of functions, where the $t$th function maps the past $t - 1$ action choices and loss observations to the action played at timestep $t$. Also note that the observed loss of a playing an action in $\tbprob_B$ is different from playing the same action in $\tbprob_A$, if and only if that action was $x_2$ and it was played $M$ times in the prior $M$ timesteps (including the current timestep). \\

\noindent Thus, since $\alg$ never plays $x_2$ for $M$ times in a row on problem $\tbprob_A$, it sees the \emph{identical} sequence of loss outputs when it is deployed on $\tbprob_B$, and hence makes the identical sequence of actions as it would have if deployed in $\tbprob_A$, which in turn implies that it never plays $x_2$ for $M$ times in a row on problem $\tbprob_B$. Since $M < T/4$ via Eq.~\eqref{eqn:lower_bound_conditions}, and playing any action $x \neq x_2$ will always yield loss at least $1/2$, the strategy that always plays $x_2$ is optimal and has cumulative loss of $M-1$. So, since Eq.~\eqref{eqn:lower_bound_conditions} implies that
$$
T/2 - (M-1) > T/2 - M > T/2 - T/4 = T/4,
$$
we have that
$$
\cpReg(\alg, \tbprob_B) \geq T/2 - (M-1) > T/4.
$$
This is a contradiction to Eq.~\eqref{eqn:lower_bound_true_bound_2}. \\

\noindent In either case, we have arrived at a contradiction. Hence, we have shown that for each $\epsilon$ satisfying the given conditions and each function $f$, the corresponding set $\algset_{\epsilon, f}$ is the empty set. This completes the proof. \hfill \qeddhruv

\section{Proof of Proposition~\ref{prop:1}}
\label{app:prop1_proof}
In this section, we provide a formal proof of Proposition~\ref{prop:1}. Let $\ActSet = \{ x_1, x_2 \}$. Let $w$ be defined componentwise as $w_i = \frac{1}{2^i}$ for each $1 \leq i \leq m$. Set $w_x = w$ for each $x \in \ActSet$. Now sample a bit string $y^\star$ uniformly at random from $\{ 0, 1 \}^{m-1}$, whose identity is kept hidden from the player. Define $h_{x_1} = 1$ and define
$$
h_{x_2}(w_{x_2}^{\top} y^{t,x_2,m}) = h_{x_2}(w^{\top} y^{t,x_2,m}) =
\begin{cases}
1 \text{ if } w^{\top} y^{t,x_2,m} \neq w^{\top}(1, y^\star) \\
0 \text{ if } w^{\top} y^{t,x_2,m} = w^{\top}(1, y^\star)
\end{cases}.
$$
We assume that there is no stochasticity in the loss feedback experienced by the player. This defines an $(m,w,h)$-weighted tallying bandit game. For ease in notation in the sequel, we also define $v \in \mathcal{X}^{m-1}$ as
$$
v_i = \begin{cases}
x_2 \text{ if } y^\star_i = 1 \\
x_1 \text{ if } y^\star_i = 0
\end{cases}.
$$
We claim that with our choice of $w$, if $y \neq y' \in \{ 1 \} \times \{ 0,1 \}^{m-1}$ then $w^\top y \neq w^\top y'$. We defer the formal proof of this claim for now, and use this claim to complete the proof of Proposition~\ref{prop:1}. Critically, the claim implies that we incur non-unit loss at timestep $t$ if and only if we play action $x_2$ at timestep $t$ and have $y^{t,x_2,m} = (1, y^\star)$. Equivalently, we incur non-unit loss at timestep $t$ if and only if our action sequence for the timesteps $t, t-1 \dots t-m$ is $(x_2, v)$. Thus, the policy that cyclically plays $v_{m-1}, v_{m-2} \dots v_1, x_2$ incurs a loss of zero at least once every $m$ timesteps. Meanwhile, identifying the optimal policy is at least as hard as playing the action sequence $(x_2, v)$, which in turn is at least as hard as identifying $y^\star$.

A standard ``needle in the haystack'' argument~\citep{du20lowerbound} then shows that identifying $y^\star$ requires $\bigomega \left( 2^m \right)$ timesteps. In turn, since the cyclic policy $v_{m-1}, v_{m-2} \dots v_1, x_2$ incurs a loss of zero at least once every $m$ timesteps, this implies that the expected CPR $\E \left[ \cpReg \right]$ of any (possibly randomized) algorithm is lower bounded by $\bigomega \left( \min \{ 2^m, T \} / m \right)$, where the expectation is over the (possible) randomization of the algorithm as well as the sampling of $y^\star$.

Let us now return to prove our claim that if $y \neq y' \in \{ 1 \} \times \{ 0,1 \}^{m-1}$ then $w^\top y \neq w^\top y'$. Assume for the sake of contradiction that $y \neq y'$ but $w^\top y = w^\top y'$. Let $J \subseteq \{ 2, 3 \dots m \}$ be the set of coordinates that $y, y'$ differ, and let $j^\star = \min J$. Note that $J$ is non-empty by assumption, and so $j^\star$ is well defined. Assume without loss of generality that $y_{j^\star} - y_{j^\star}' = 1$ (the case when $y_{j^\star} - y_{j^\star}' = -1$ is completely symmetric). Then observe that
\begin{equation}
\label{eqn:prop_proof_helper_1}
\begin{aligned}
0 &= w^\top (y - y') \\
&= \sum_{j = 1}^m w_j (y_j - y_j') \\
&= \sum_{j \in J} w_j (y_j - y_j') \\
&= w_{j^\star} + \sum_{j \neq j^\star \in J} w_j (y_j - y_j').
\end{aligned}
\end{equation}
We can now lower bound Eq.~\eqref{eqn:prop_proof_helper_1} as
\begin{equation}
\label{eqn:prop_proof_helper_2}
\begin{aligned}
0 &= w_{j^\star} + \sum_{j \neq j^\star \in J} w_j (y_j - y_j') \\
&\geq w_{j^\star} - \sum_{j \neq j^\star \in J} w_j \vert y_j - y_j' \vert \\
&= w_{j^\star} - \sum_{j \neq j^\star \in J} w_j \\
&\geq w_{j^\star} - \sum_{j = j^\star + 1}^{m} w_j.
\end{aligned}
\end{equation}
Now substituting in our choice of $w$ into Eq.~\eqref{eqn:prop_proof_helper_2}, we find that
$$
0 \geq w_{j^\star} - \sum_{j = j^\star + 1}^{m} w_j = \frac{1}{2^{j^\star}} - \sum_{j = j^\star + 1}^{m} \frac{1}{2^j} = \frac{1}{2^{j^\star}} \left( 1 - \sum_{j=1}^{m - j^\star} \frac{1}{2^j} \right) > 0,
$$
which of course is a contradiction. This proves our claim that if $y \neq y' \in \{ 1 \} \times \{ 0,1 \}^{m-1}$ then $w^\top y \neq w^\top y'$. \hfill \qeddhruv

\section{Extended Numerical Results \& Details}
\subsection{Unweighted Tallying Bandit}
\label{app:numerical_results_unweighted}
\begin{figure}
\centering
\begin{tabular}{ccc}
\includegraphics[width=.33\linewidth]{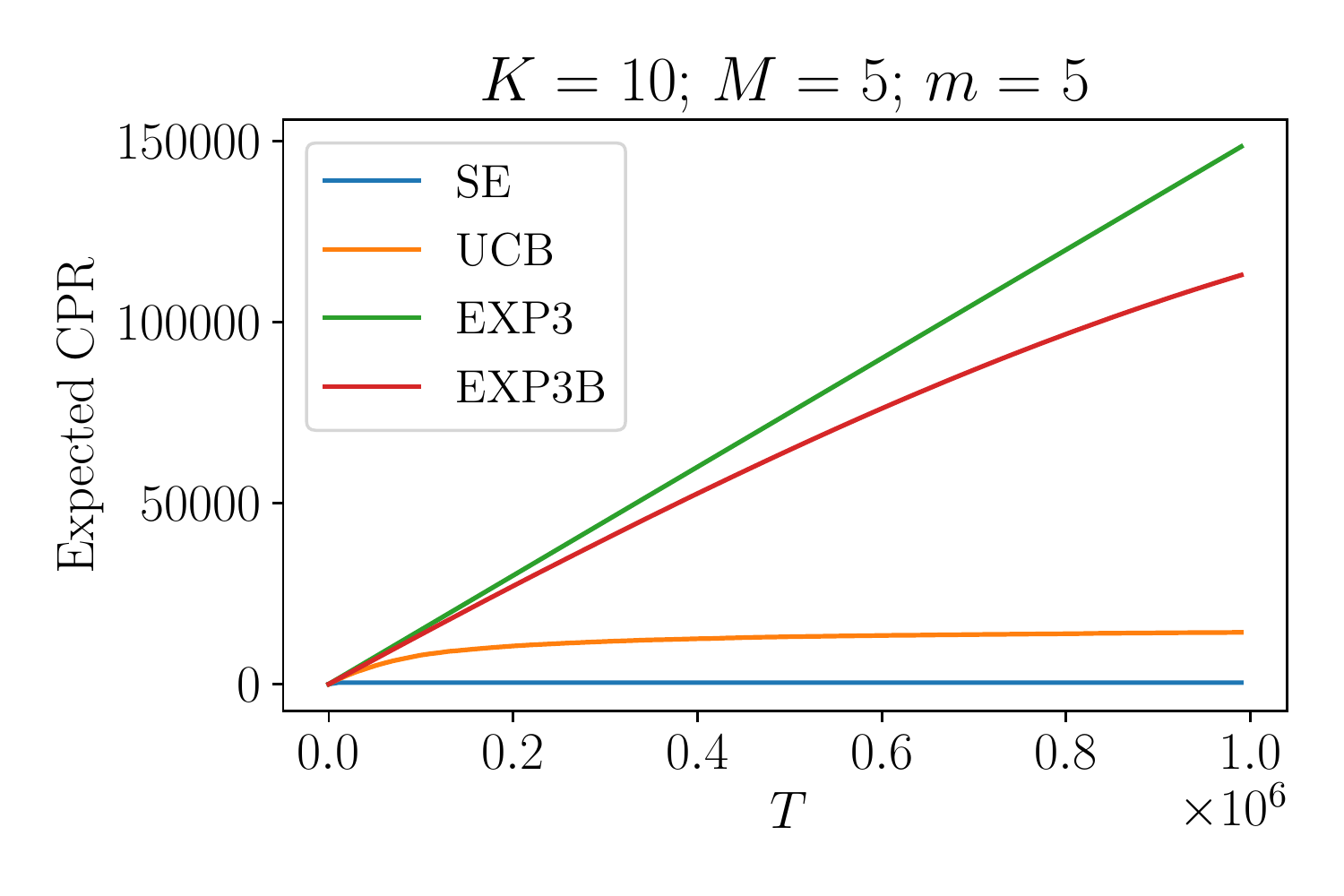} &
\includegraphics[width=.33\linewidth]{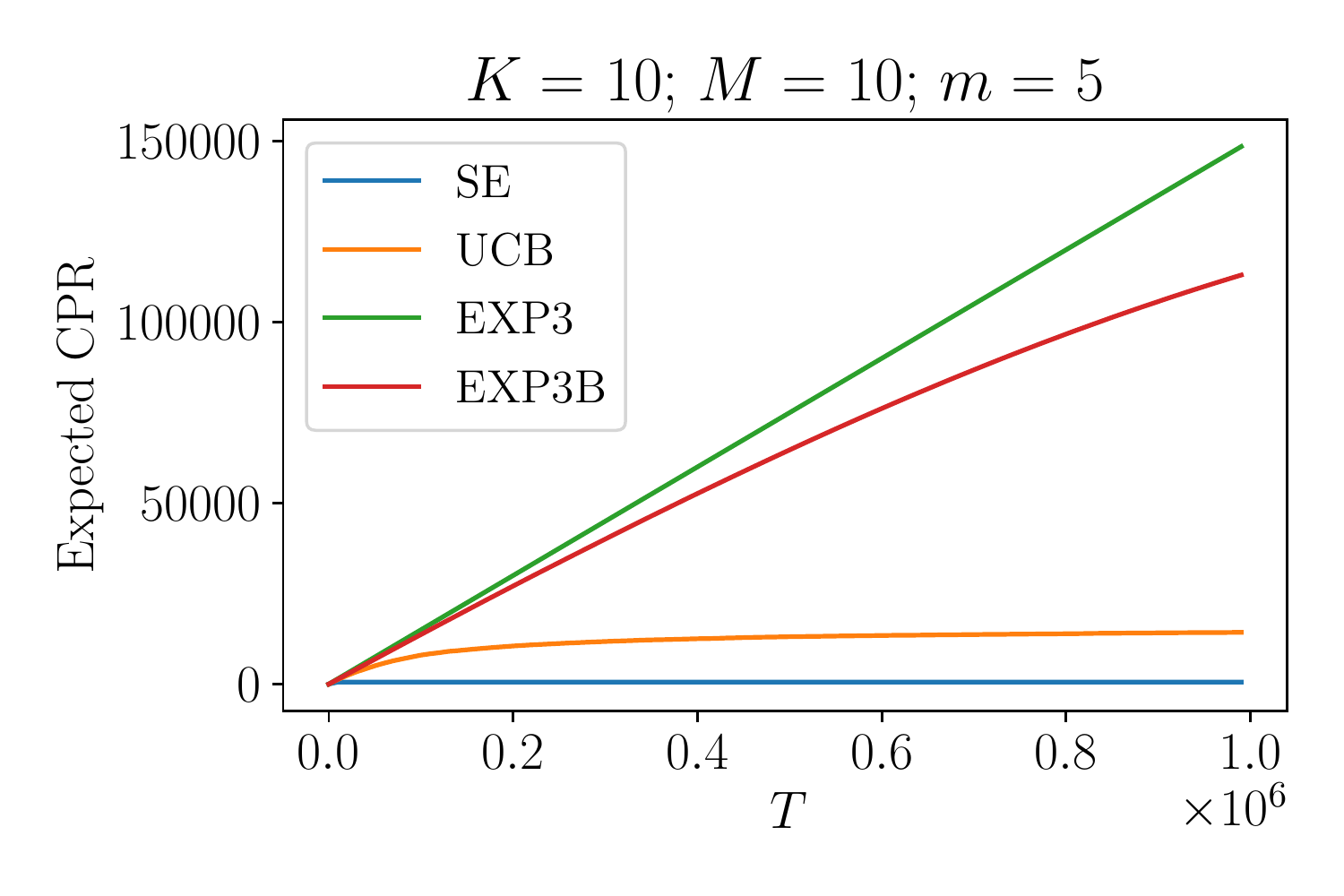} &
\includegraphics[width=.33\linewidth]{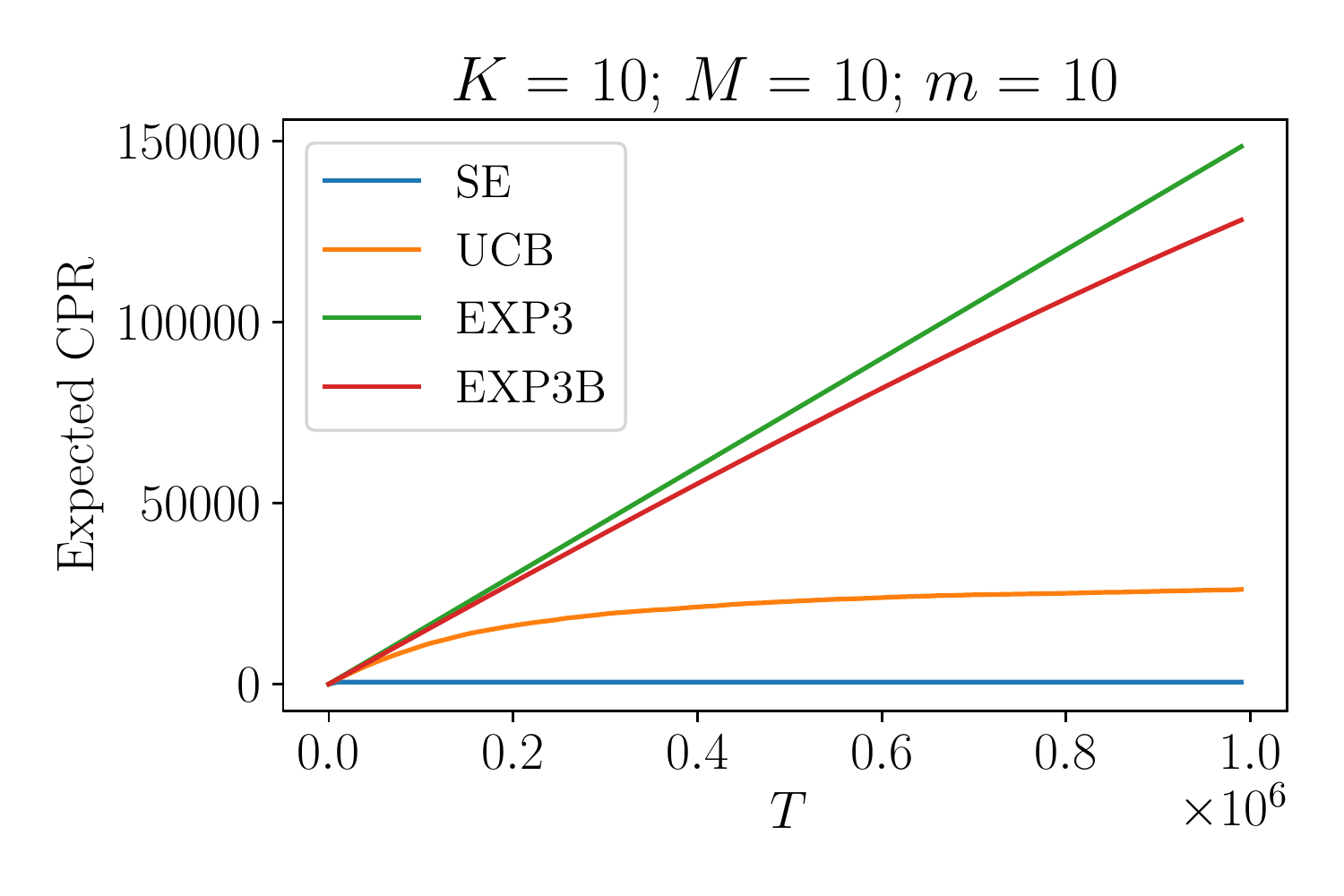} \\
\includegraphics[width=.33\linewidth]{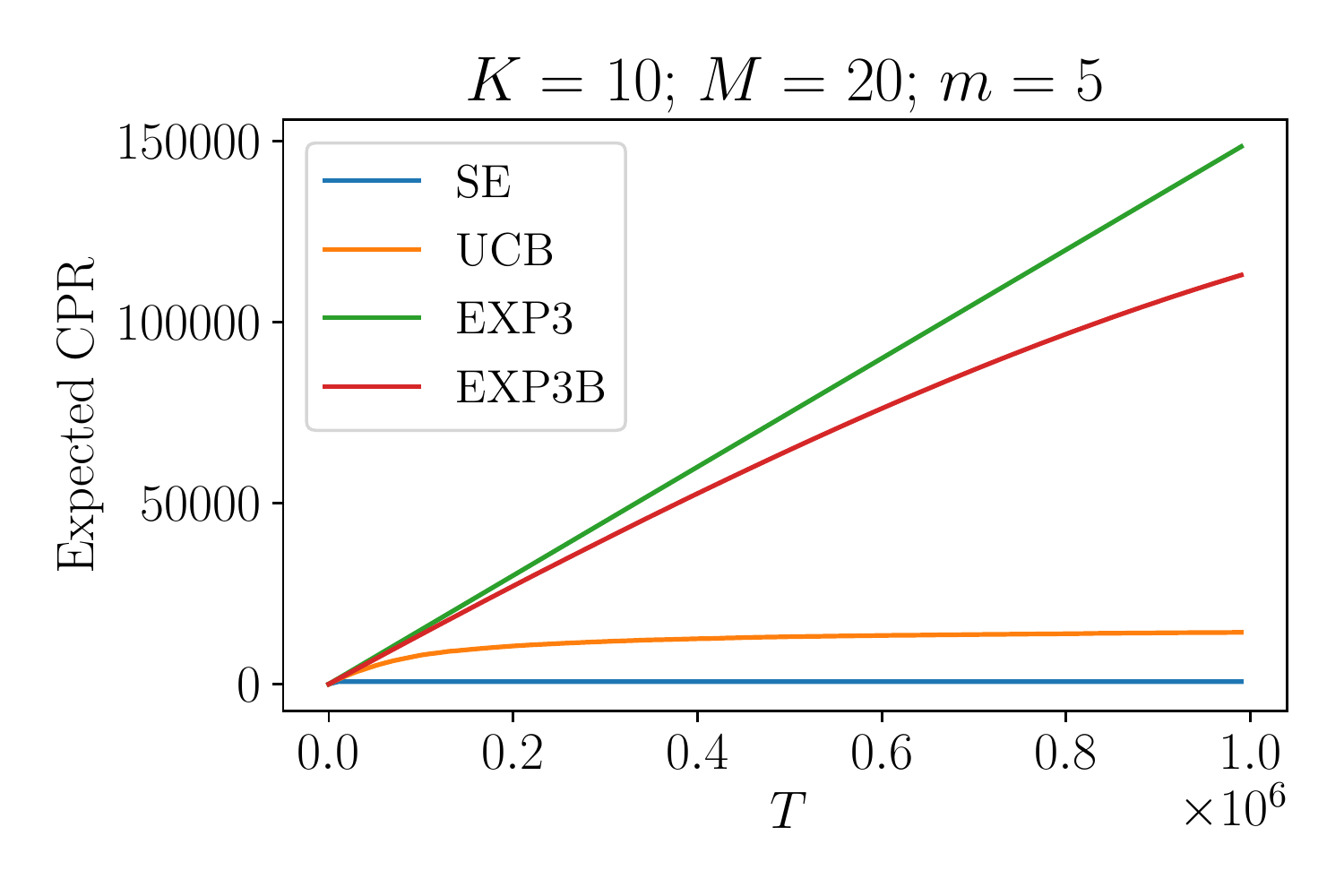} &
\includegraphics[width=.33\linewidth]{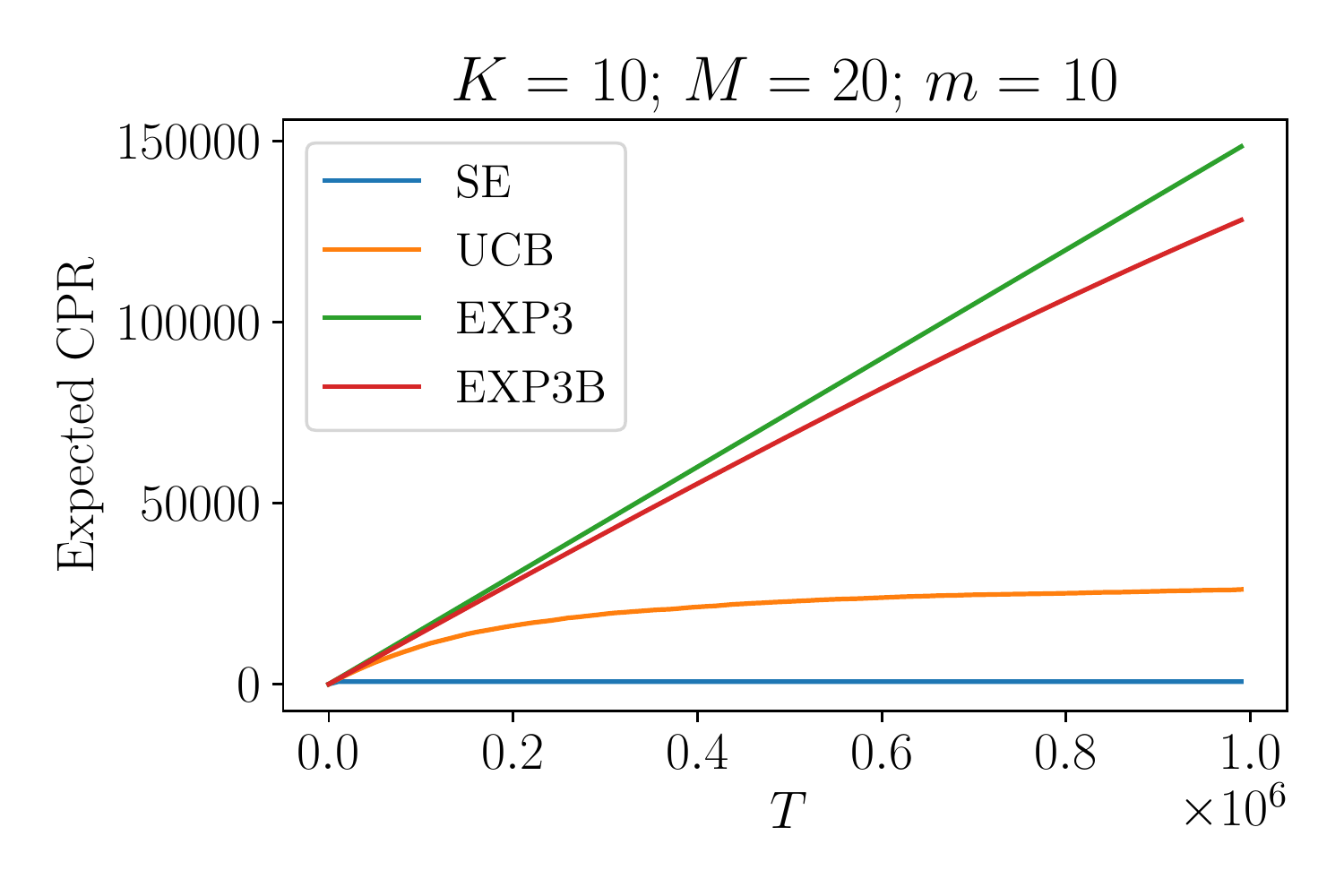} &
\includegraphics[width=.33\linewidth]{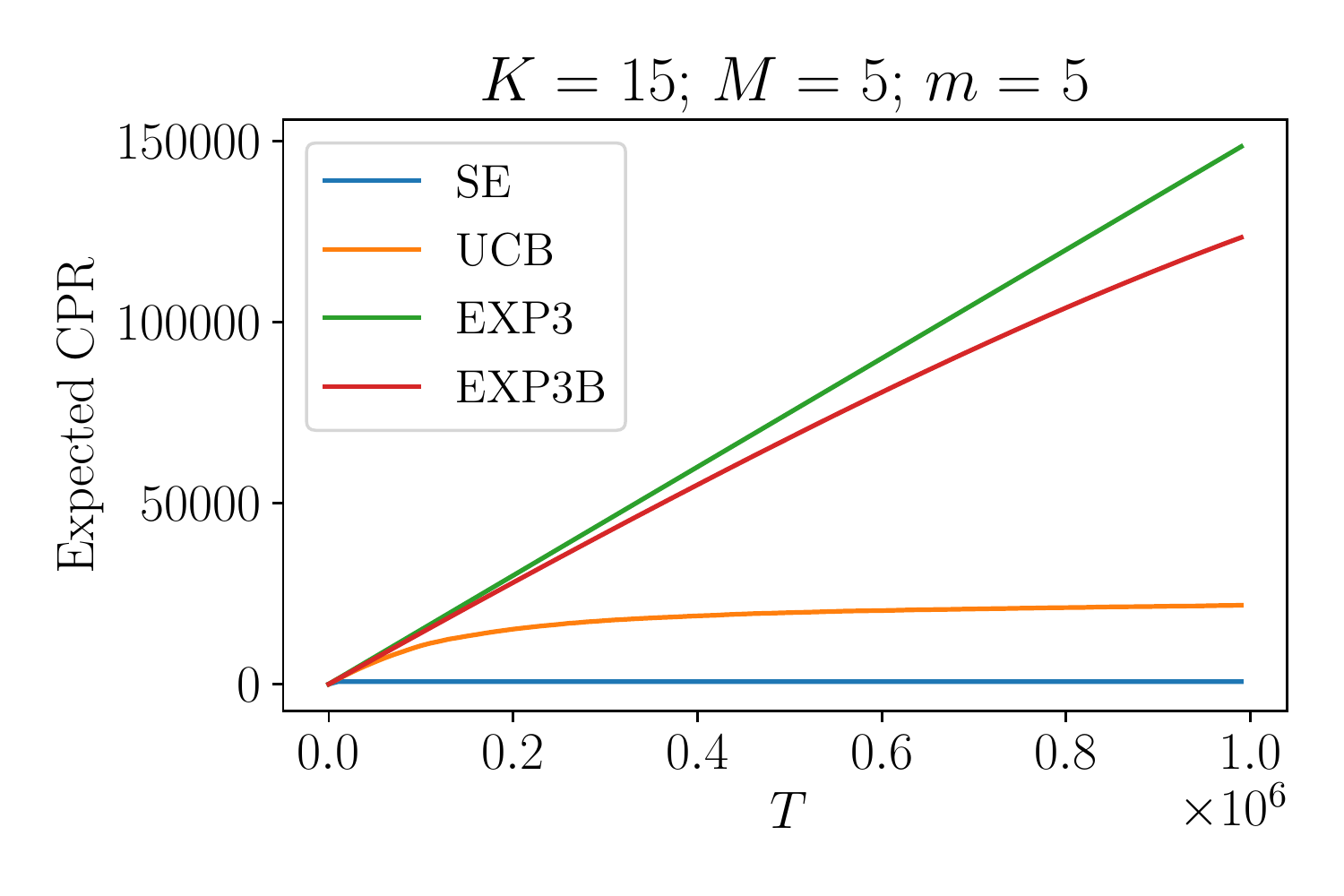} \\
\includegraphics[width=.33\linewidth]{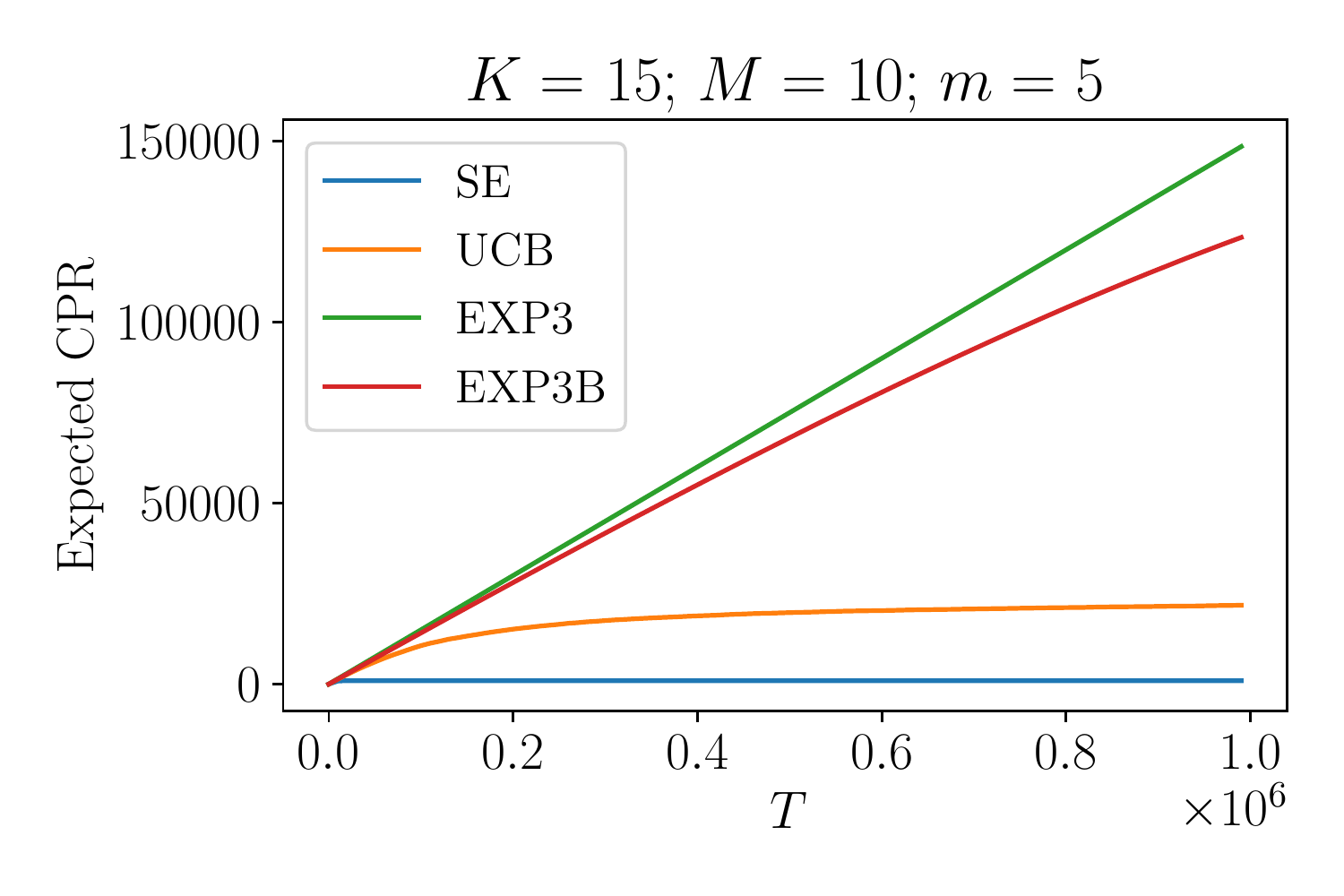} &
\includegraphics[width=.33\linewidth]{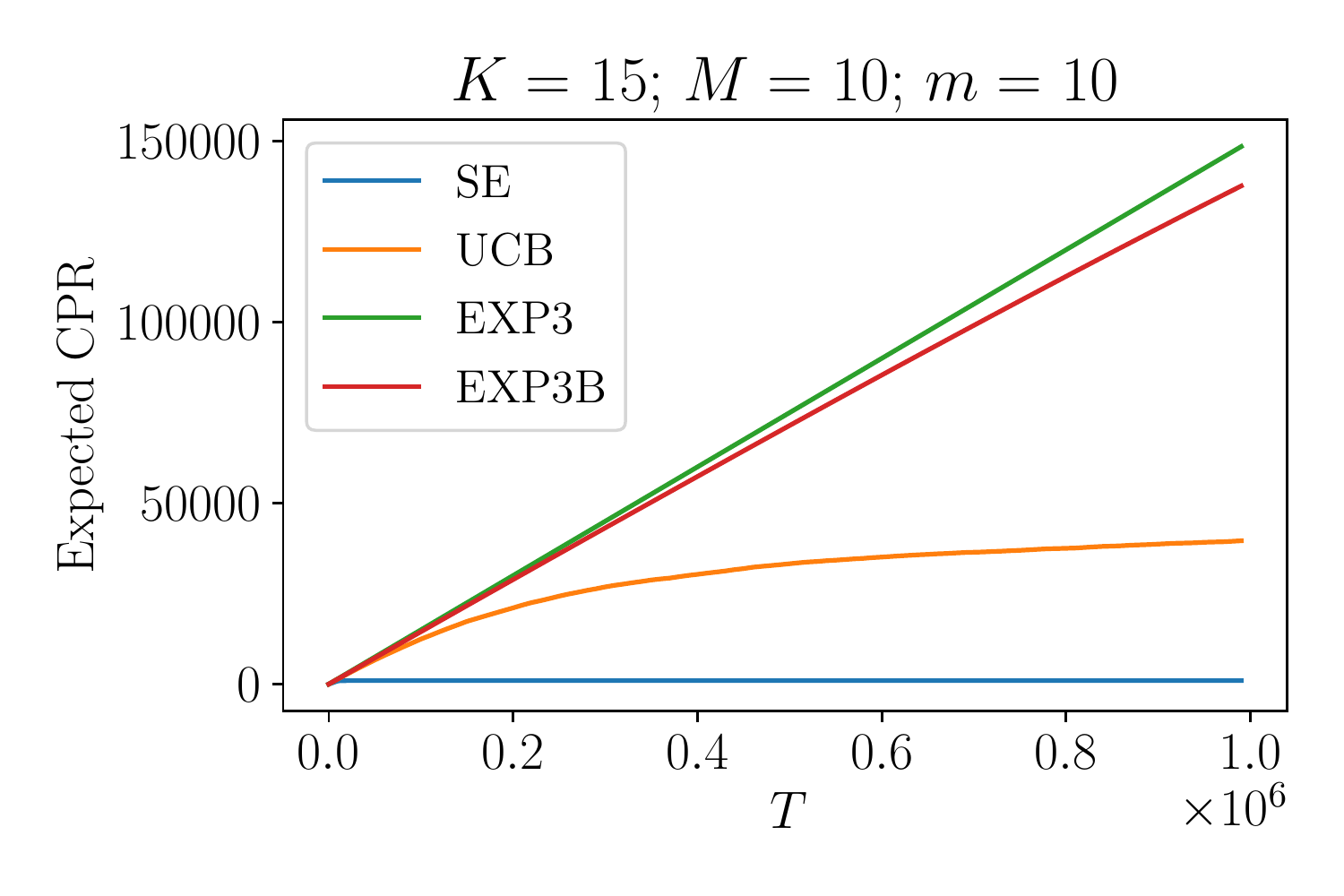} &
\includegraphics[width=.33\linewidth]{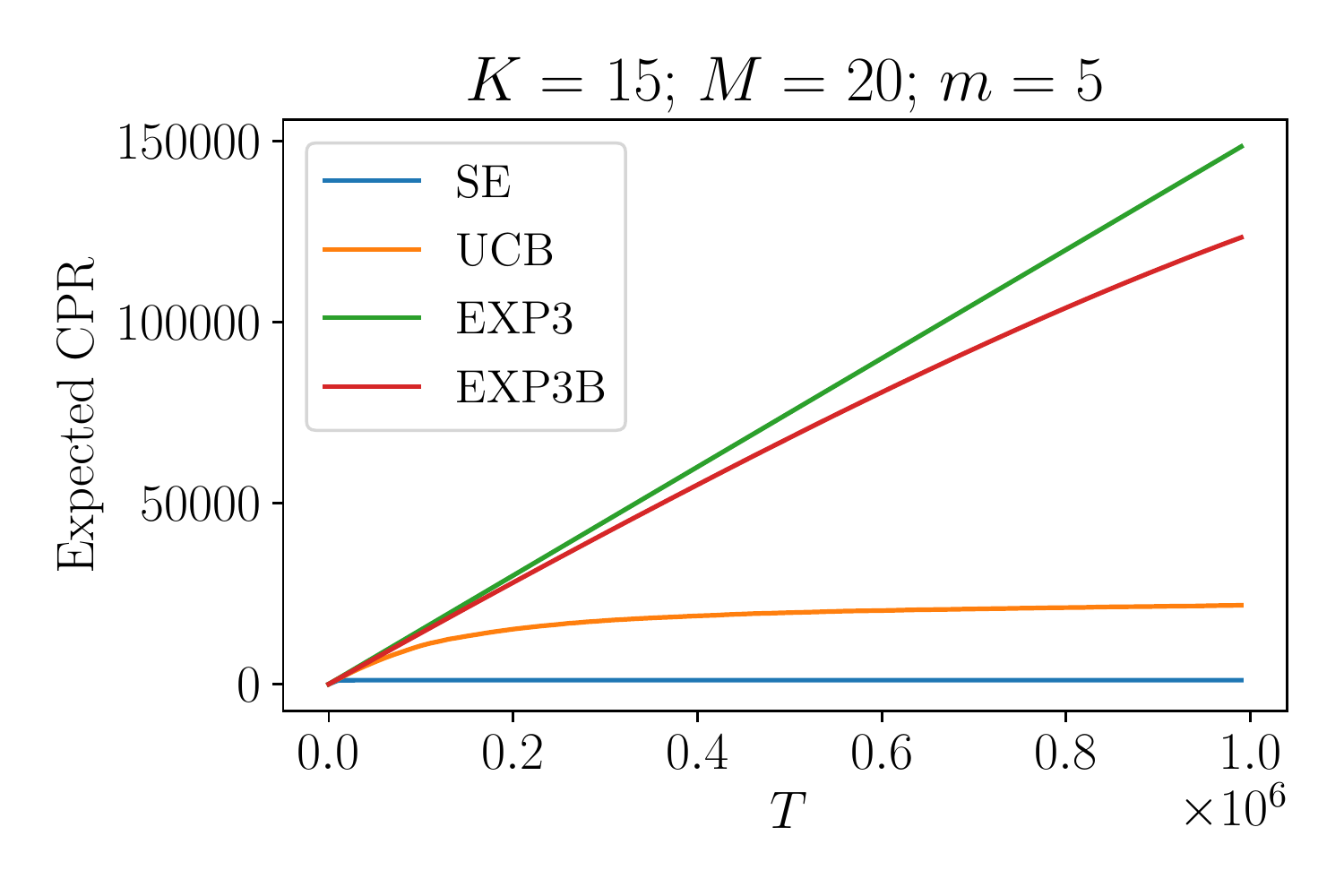} \\
& \includegraphics[width=.33\linewidth]{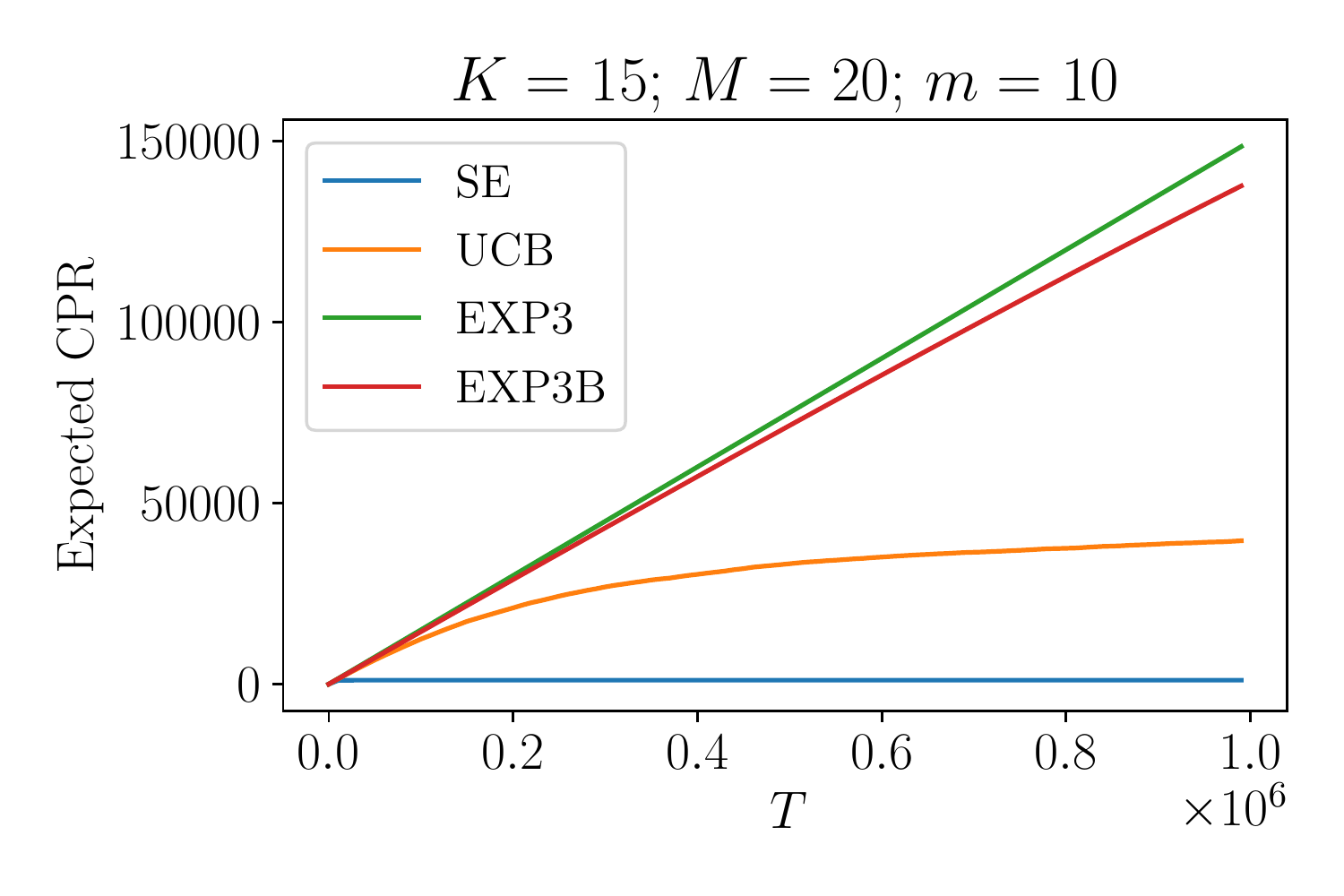}
\end{tabular}
\caption{We plot the expected CPR of each algorithm, when deployed on the unweighted tallying bandit problem described in Section~\ref{sec:numerical_results_unweighted}, with varying values of $m, K, M$. In all plots, each datapoint is obtained by averaging over 20 problem instances, and the shaded region depicts $\pm 1$ standard error around the mean.}
\label{fig:app_unweighted}
\end{figure}

In this section, we first present additional experimental results for the unweighted tallying bandit problem, where the loss functions are identical to those described in Section~\ref{sec:numerical_results_unweighted}. Hence, this problem satisfies $0$-REO. Here, we vary the values of $m, K, M$, and plot the CPR of each method over time. The results are shown in Figure~\ref{fig:app_unweighted}. In each case, we observe that SE outperforms the baselines. These results also show that the performance of Algorithm~\ref{alg:main} is robust to using a conservative upper bound $M$ on $m$.

\begin{figure}
\centering
\begin{tabular}{c}
\includegraphics[width=0.75\linewidth]{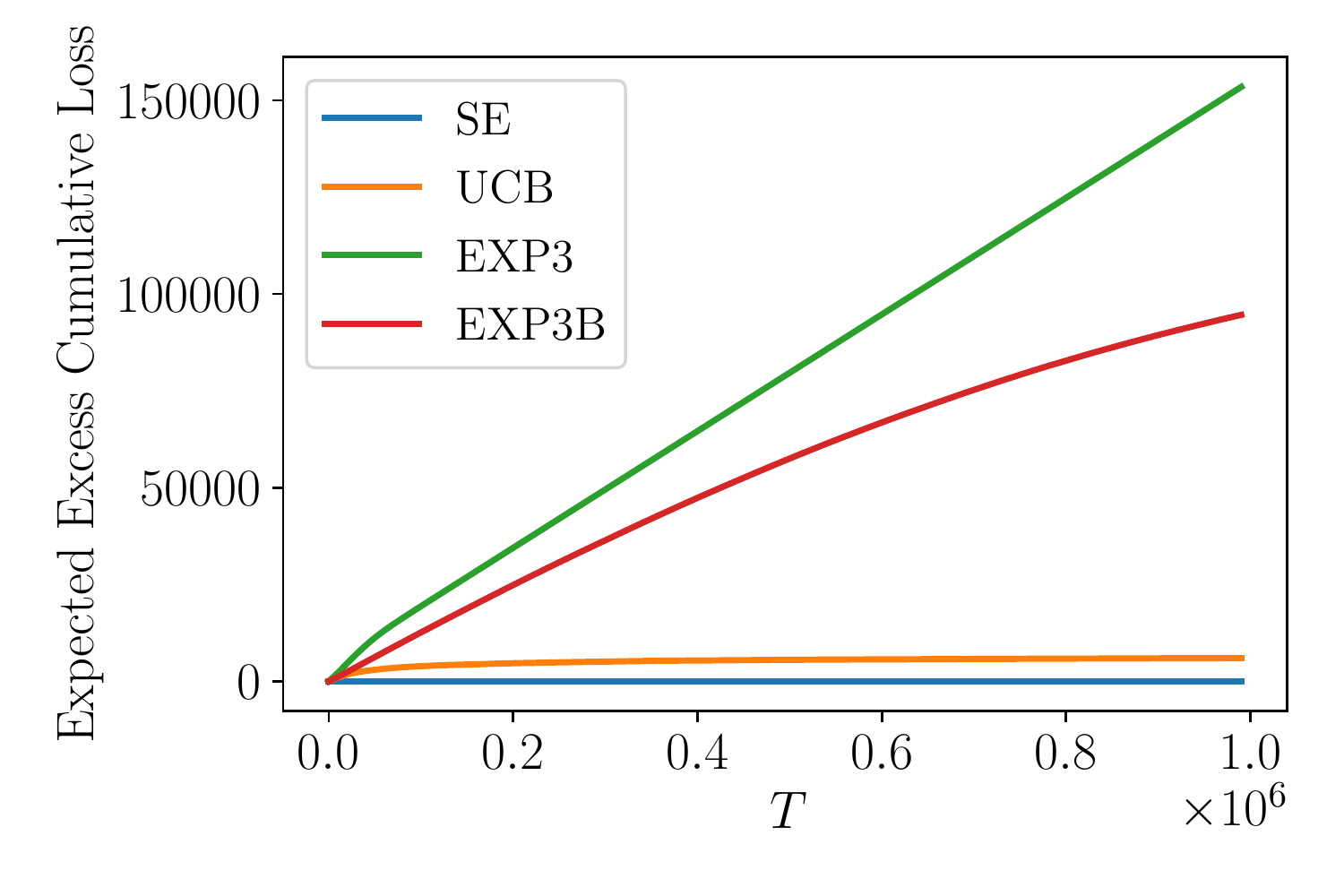}
\end{tabular}
\caption{We plot as a function of time the expected cumulative loss of each algorithm in excess of that of SE, on the unweighted tallying bandit instance described in Appendix~\ref{app:numerical_results_unweighted}, with $m = 4, K = 5, M = 4$. Note this instance satisfies $\alpha$-REO with $\alpha = 0.1125$. Each datapoint is obtained by averaging over 20 problem instances, and the shaded region depicts $\pm 1$ standard error around the mean.}
\label{fig:alpha_nonzero_main}
\end{figure}

We now present results for a different unweighted tallying bandit problem, where $\alpha$-REO is satisfied with $\alpha > 0$. For each action $x \in \ActSet$, we define $w_x = \vec{1} / (4m)$. We fix an action $x^\star \in \ActSet$ and a different action $x^{\star \star} \in \ActSet$, and then define the loss functions $\{ h_x \}_{x \in \ActSet}$ as
$$
h_{x}(y^{t, x, m}) =
\begin{cases}
1 - w_{x}^\top y^{t, x, m} - 0.15 \text{ if } x = x^\star, y^{t, x, m} = \vec{1} \\
1 - w_{x}^\top y^{t, x, m} - \frac{m - 1}{2m} - 0.2 \text{ if } x = x^{\star \star}, y^{t, x, m} = (1, 0, 0 \dots 0) \\
1 - w_{x}^\top y^{t, x, m} \text{ otherwise}
\end{cases}.
$$
A numerical computation reveals that this unweighted tallying bandit problem instance satisfies $\alpha$-REO with $\alpha = \max \{0, -0.2 + \frac{2m - 3}{4m} \}$. We empirically study the performance of Algorithm~\ref{alg:main} relative to the baselines on this problem instance, with a choice of $m = 4, K = 5, M = 4$. Since the optimal policy in this problem is not obvious, the CPR is difficult to compute. So in lieu of the CPR, we plot the expected cumulative loss of each algorithm in excess of SE's loss (hence the CPR at any time is obtained by applying a constant shift to each algorithm's excess loss). The results are shown in Figure~\ref{fig:alpha_nonzero_main}, where we observe that our method outperforms all others.

\subsection{Weighted Tallying Bandit}
\label{app:numerical_results_weighted_tb}
Here, we describe the loss functions that were used to define the WTB problem instance described in Section~\ref{sec:numerical_results_weighted_tb}. For each action $x \in \ActSet$, we define $w_x$ in the following fashion. First we define the vector $v \in \R^m$ coordinate wise by setting its $i$th coordinate as $v_i = 1/2^i$. Then we set $w_x = v / (2\| v \|_1)$ for each $x \in \ActSet$. We fix an action $x^\star \in \ActSet$, and then define the loss functions $\{ h_x \}_{x \in \ActSet}$ as
$$
h_{x}(y^{t, x, m}) =
\begin{cases}
1 - w_{x}^\top y^{t, x, m} - 0.15 \text{ if } x = x^\star, y^{t, x, m} = \vec{1} \\
1 - w_{x}^\top y^{t, x, m} \text{ otherwise}
\end{cases}.
$$
Hence, this weighted tallying bandit problem satisfies $0$-REO.

\subsection{Simulated F1 Tournament}
\label{app:f1_tournament}

In this section, we provide details of our F1 lap time dataset, data processing and probabilistic model fitting, as well as criteria used to define meaningful WTB problem instances. We conclude this section with extended results for the simulated F1 tournament.

\subsubsection{Dataset \& Data Processing}

As discussed in \ref{sec:numerical_results_f1}, we make use of pre-first-pit-stop F1 lap time data from 1950-2022 \citep{f1dataset} to learn a probabilistic lap time model for various drivers and races, which we then use to define WTB instances. This dataset consists of race data from 1120 races and 858 drivers. Note that in each race, a different subset of drivers participated in the race. For the purposes of this paper, each entry in the dataset is a tuple of the form $(\texttt{driver ID},~\texttt{race ID},~\texttt{Lap Times})$, where $\texttt{driver ID} \in \{1, \dots, 858\}$, $\texttt{race ID} \in \{1, \dots, 1120\}$, and \texttt{Lap Times} is a list of tuples, where each tuple of the form $(\texttt{Lap Time},~\texttt{Pit Stop})$. \texttt{Lap Time} denotes the official recorded time in seconds for the driver to complete a lap during the F1 tournament, and \texttt{Pit Stop} is a binary-valued variable indicating whether the lap included a pit stop. 

In order to make lap times comparable across races, all lap times in this dataset have been normalized such that they lie in the interval [0,1]. Specifically, if the raw lap time for driver $i$'s $k^{\text{th}}$ lap in race $j$ is $\ell_{i,j,k}$, then the normalized lap time is given by $ \tau_{i,j,k} := \frac{\ell_{i,j,k} - \min_{w,x} \ell_{w,j,x}}{\max_{y,z} \ell_{y,j,z} - \min_{w,x} \ell_{w,j,x}}$.

Additionally, for each race $j$, we filter lap time data to include eligible drivers that have sufficient data to justify fitting a model. Within a fixed race $j$, a driver is considered eligible if the dataset includes at least 8 consecutive lap time datapoints until their first pit stop. After the set of eligible drivers has been determined for race $j$, we shorten the sequence of all drivers' lap time data in race $j$ to match the length of the shortest sequence in race $j$. For example, if there are 3 eligible drivers in race $j$, where driver $a$ takes 8 laps until taking a pit stop, driver $b$ takes 9 laps, and driver $c$ takes 10 laps, then we only make use of the first 8 lap times of drivers $b$ and $c$ when learning their lap time model. 

\subsubsection{Probabilistic Model Fitting}
We use our normalized lap time data to fit a probabilistic model of lap times for each eligible driver-race pair with maximum likelihood. In particular, we assume that for driver $i$ in race $j$, the $k^{\text{th}}$ normalized lap time, which is denoted $\tau_{i,j,k}$, has distribution
$$
\mathcal{N} \left( \beta_{i,j} \exp( - k \alpha_{i,j}) - |\gamma_{i,j}|k,~\sigma^2_{i,j} \right).
$$
Thus, for each driver-race $(i,j)$ pair, our formulation involves fitting 4 parameters $(\alpha_{i,j}, \beta_{i,j}, \gamma_{i,j}, \sigma_{i,j})$. We use all normalized pre-first-pit-stop data (at least 8 datapoints) to fit these parameters using maximum likelihood with $\texttt{scipy.optimize.curve\_fit}$, which leverages the Levenberg-Marquardt algorithm as implemented in MINPACK~\citep{scipy20}.

\begin{figure}
\begin{tabular}{ccc}
\includegraphics[width=.33\linewidth]{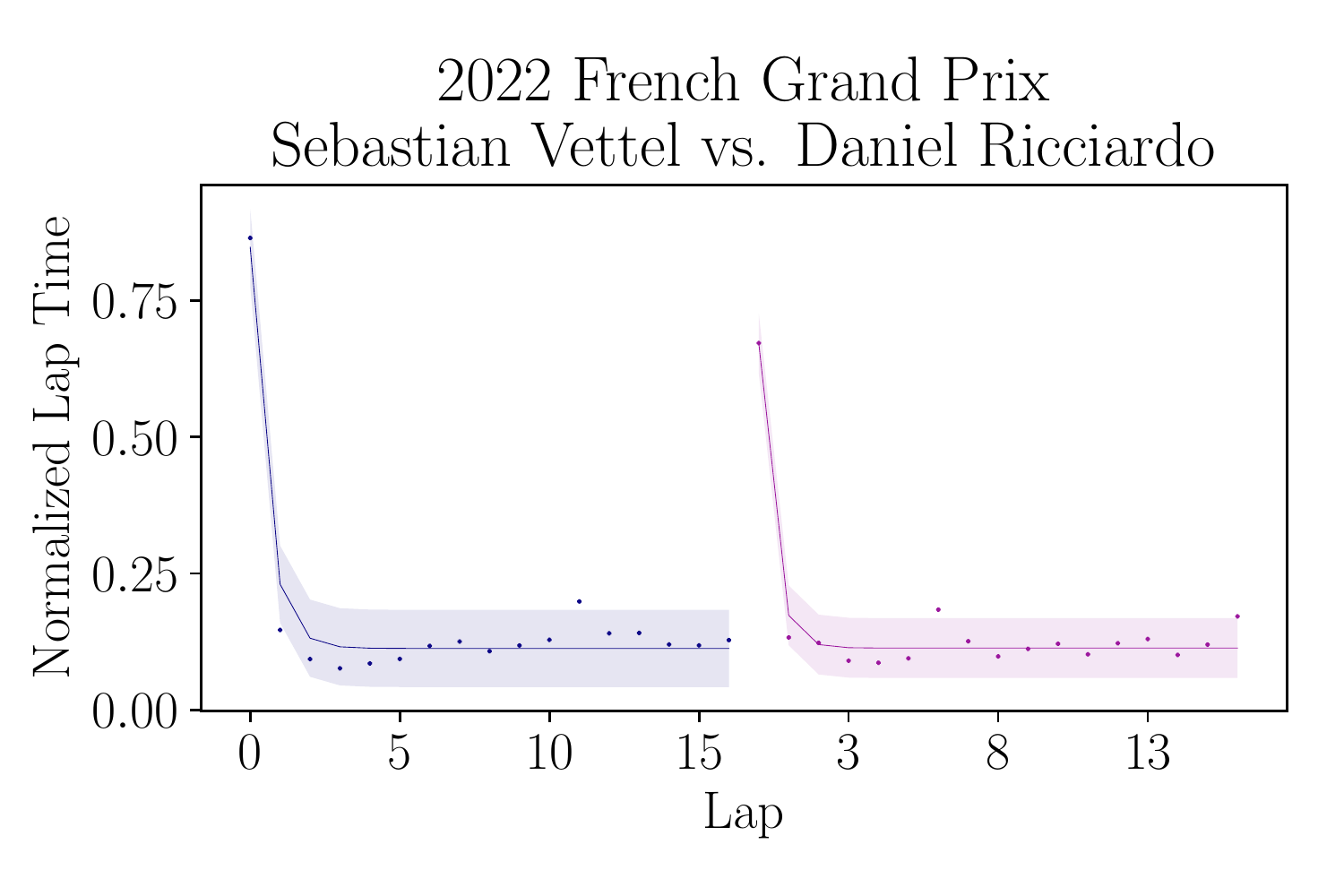} &  
\includegraphics[width=.33\linewidth]{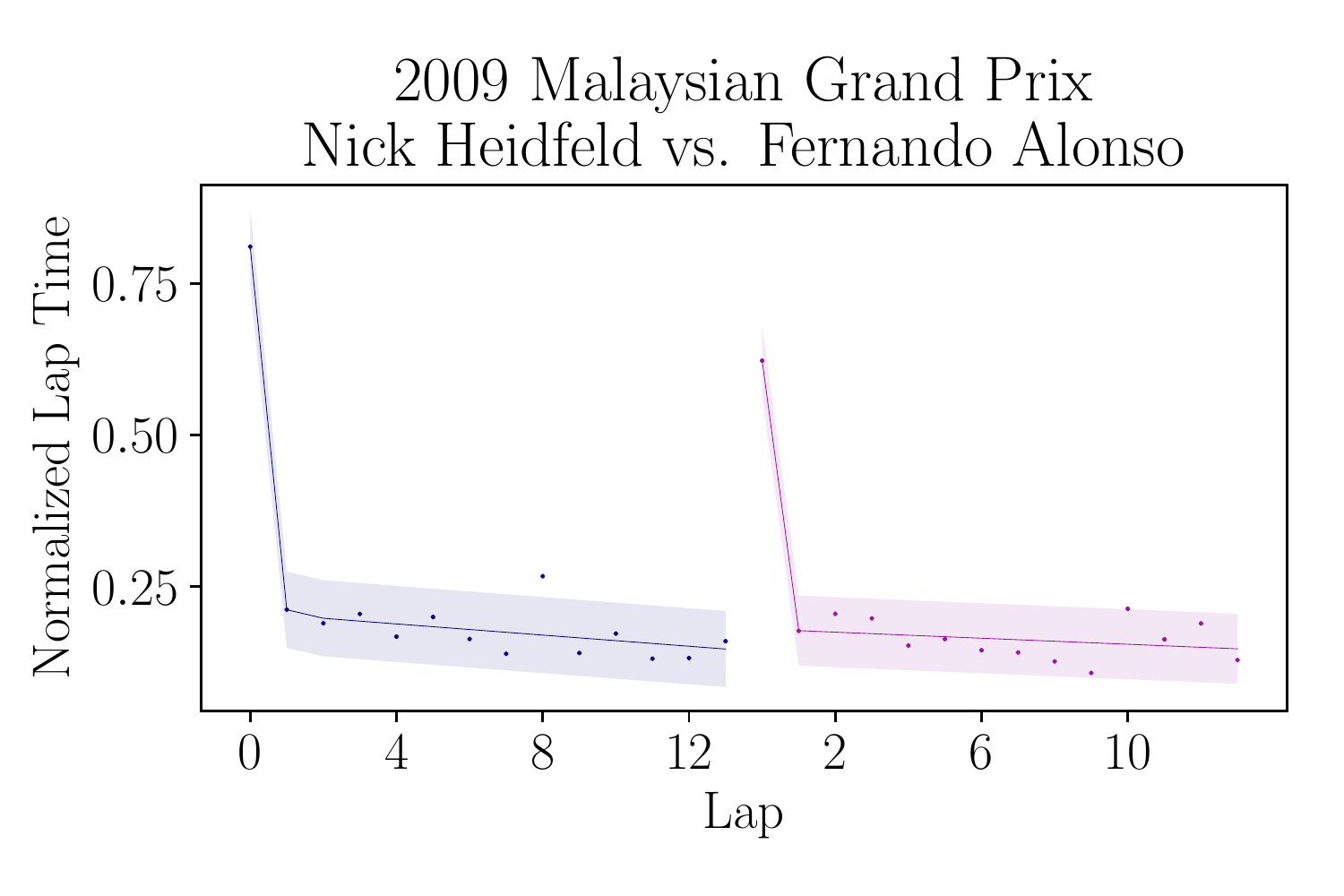}   &   \includegraphics[width=.33\linewidth]{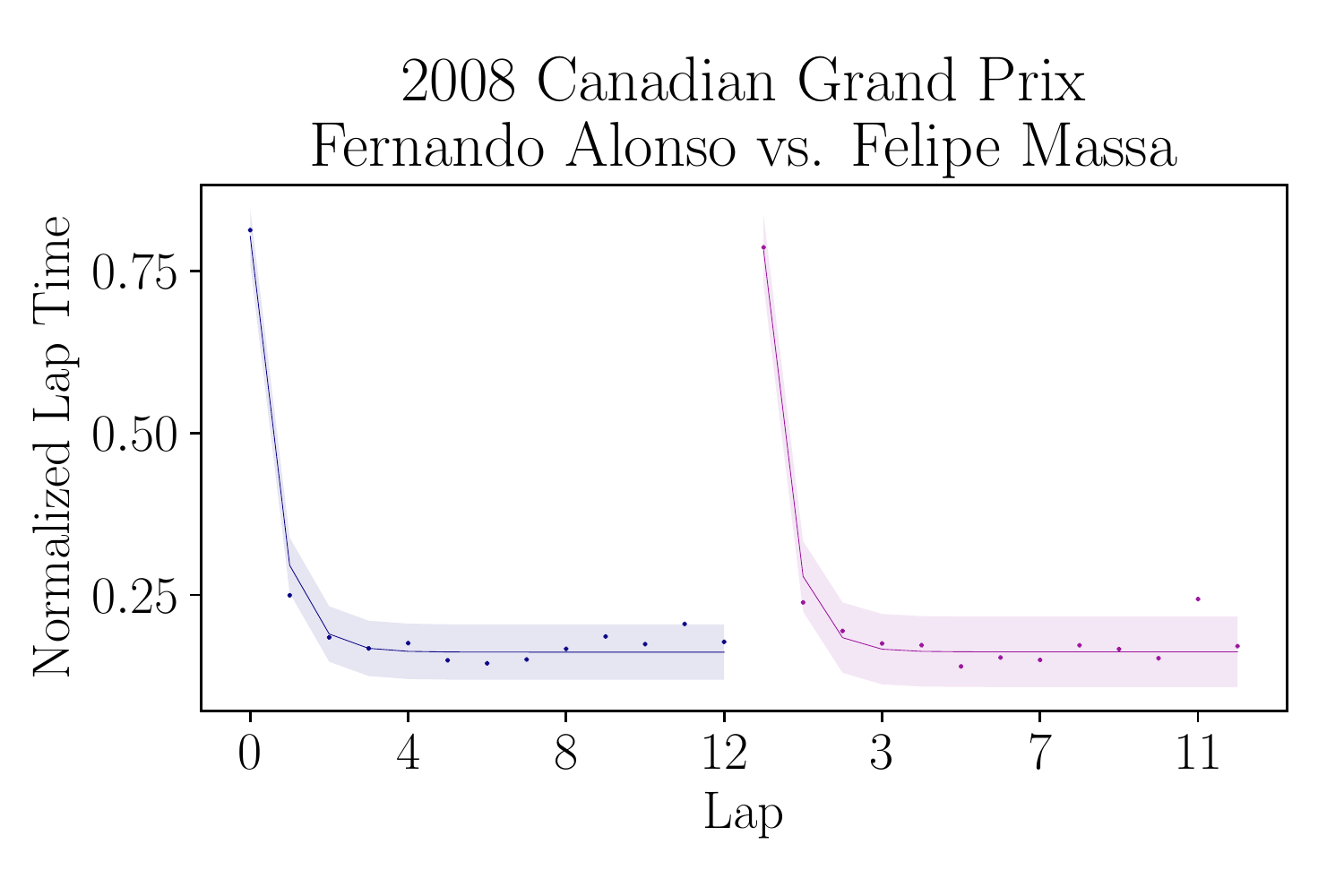}  \\
\includegraphics[width=.33\linewidth]{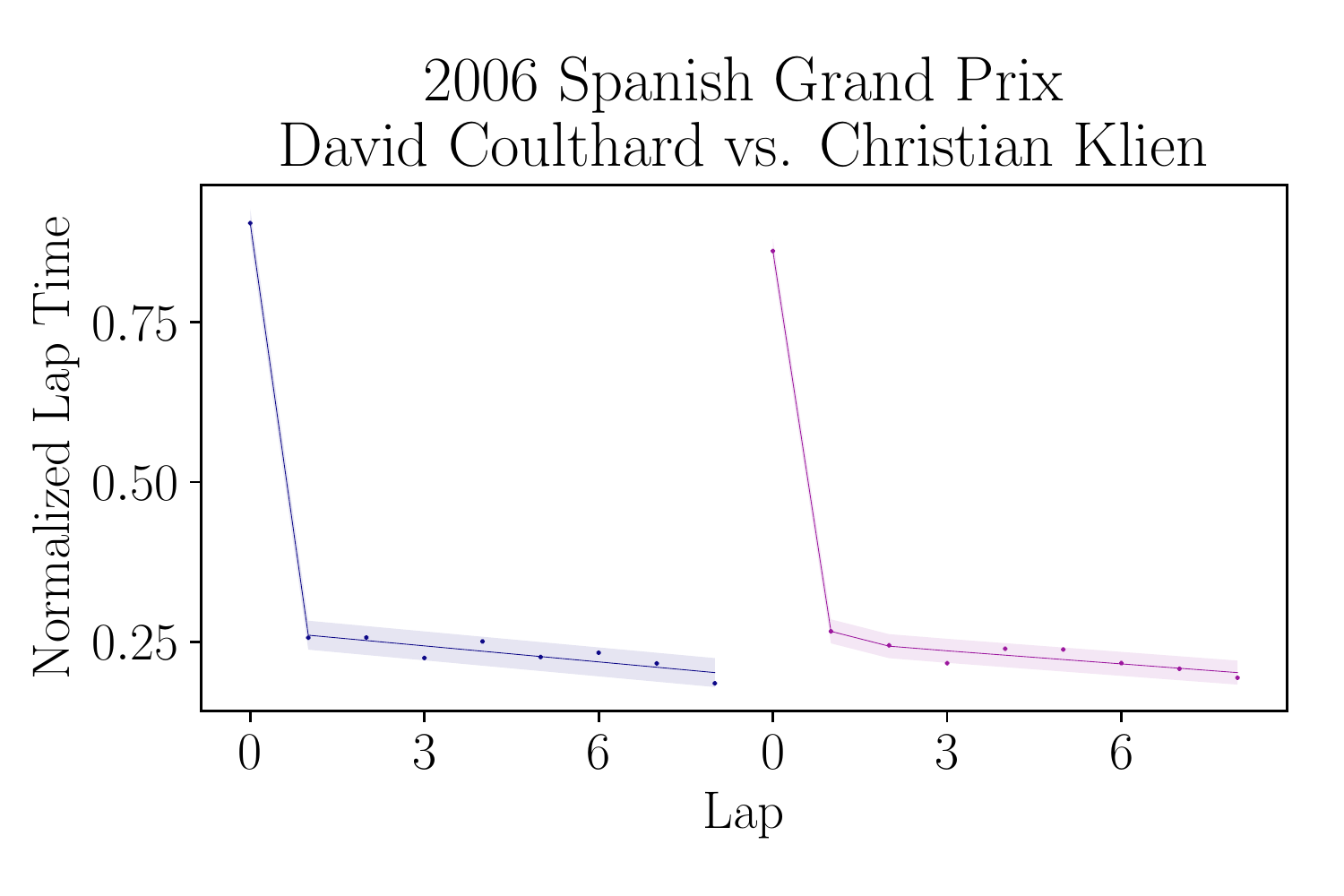}  &
\includegraphics[width=.33\linewidth]{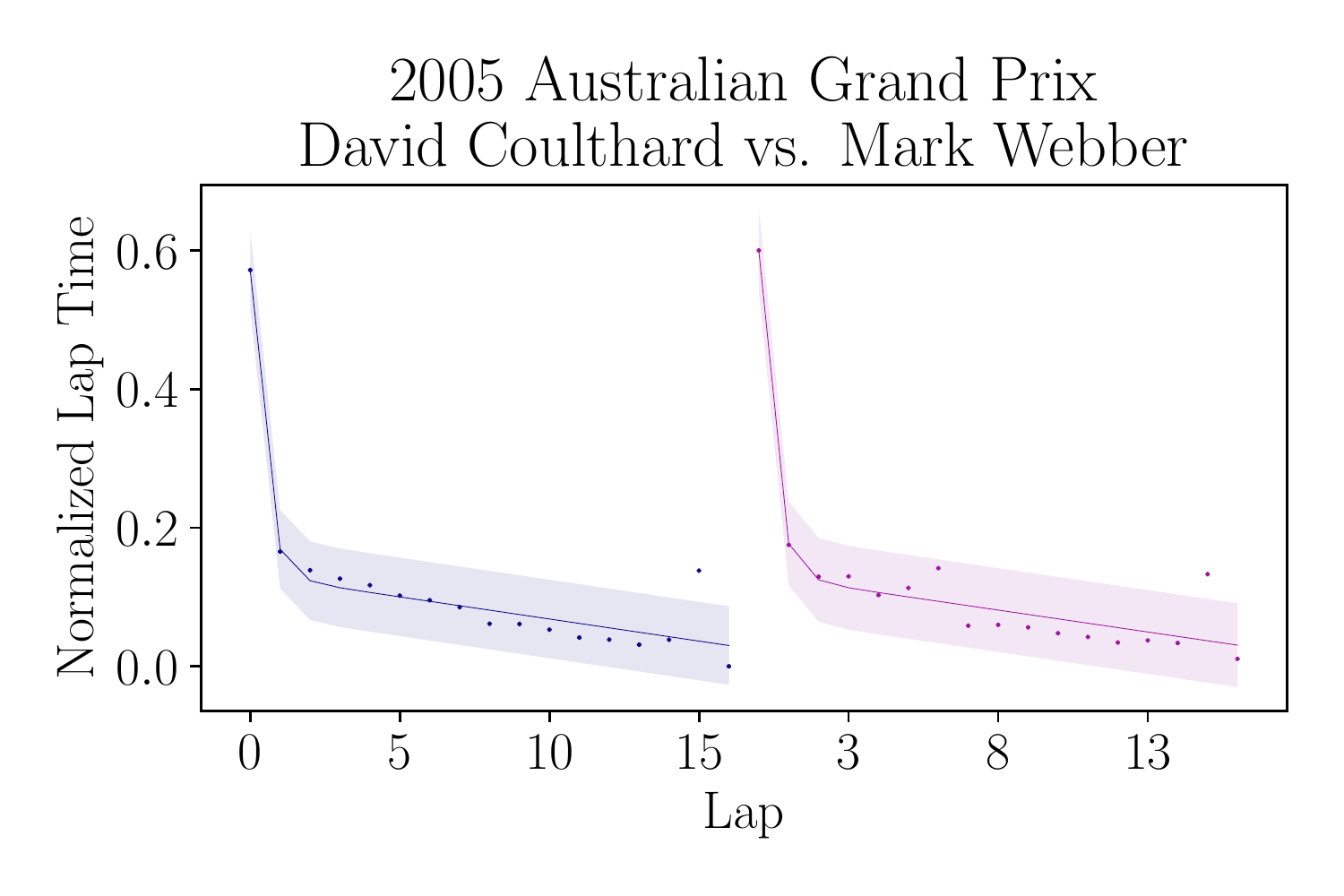} &  \includegraphics[width=.33\linewidth]{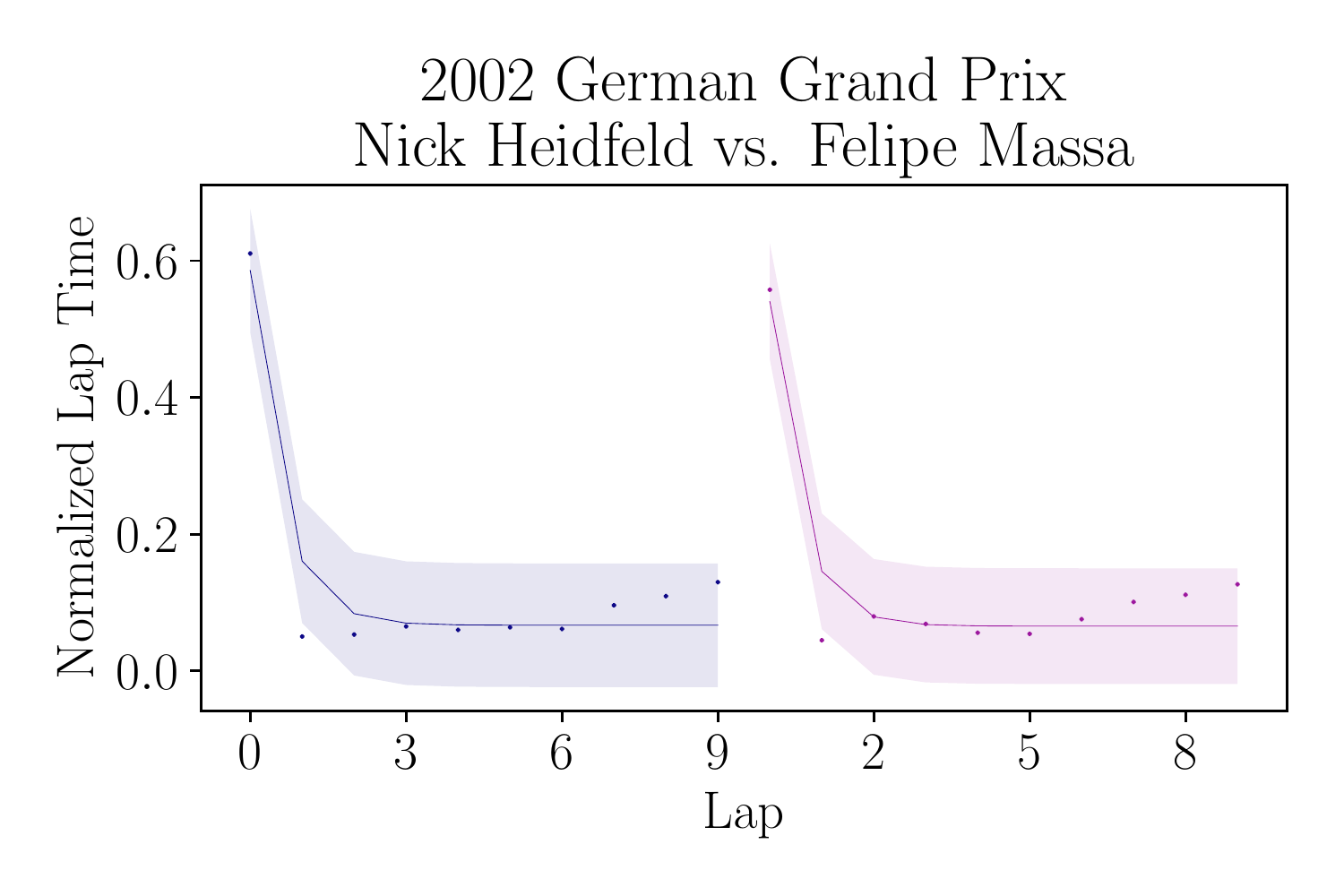}  \\
\includegraphics[width=.33\linewidth]{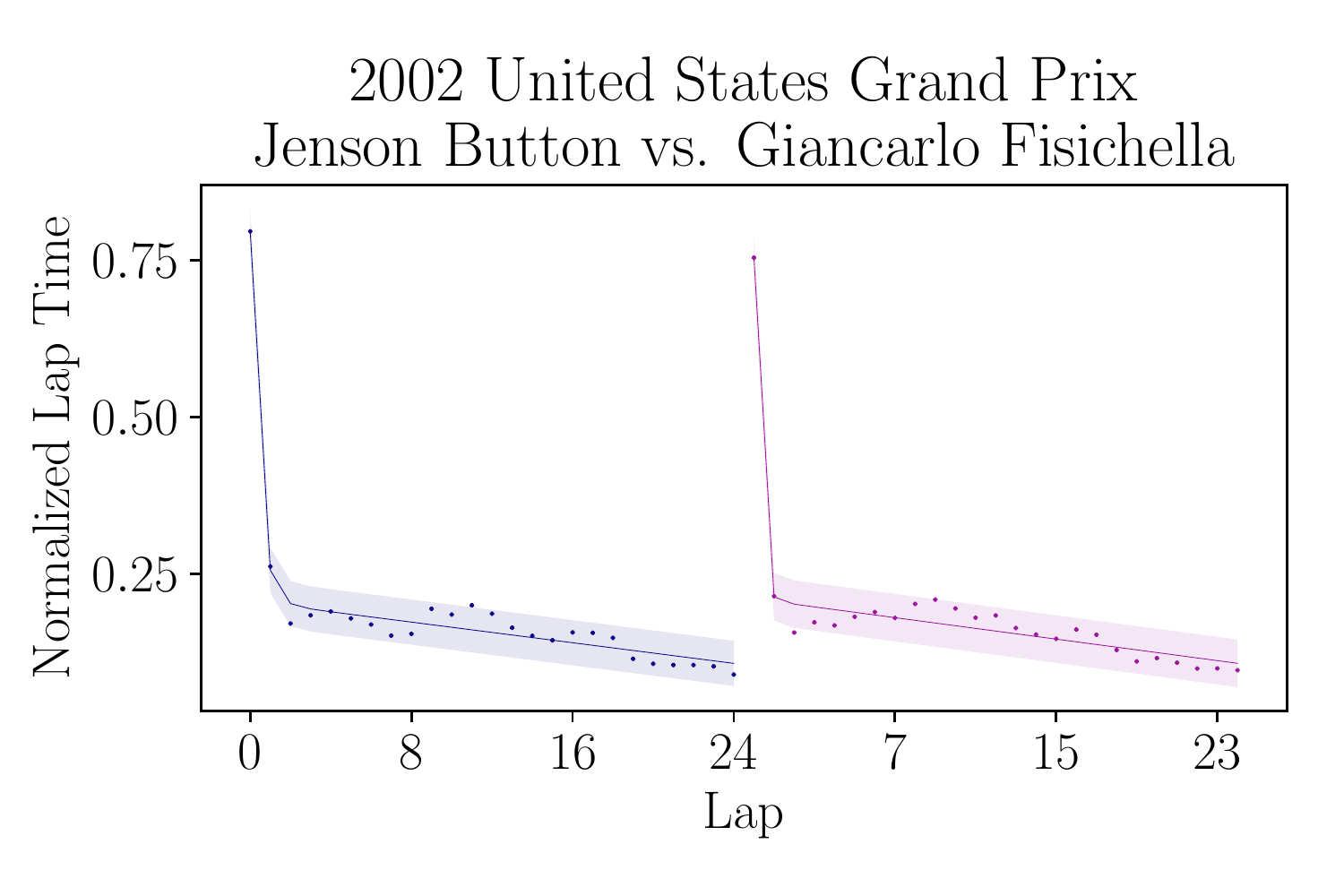} &
 \includegraphics[width=.33\linewidth]{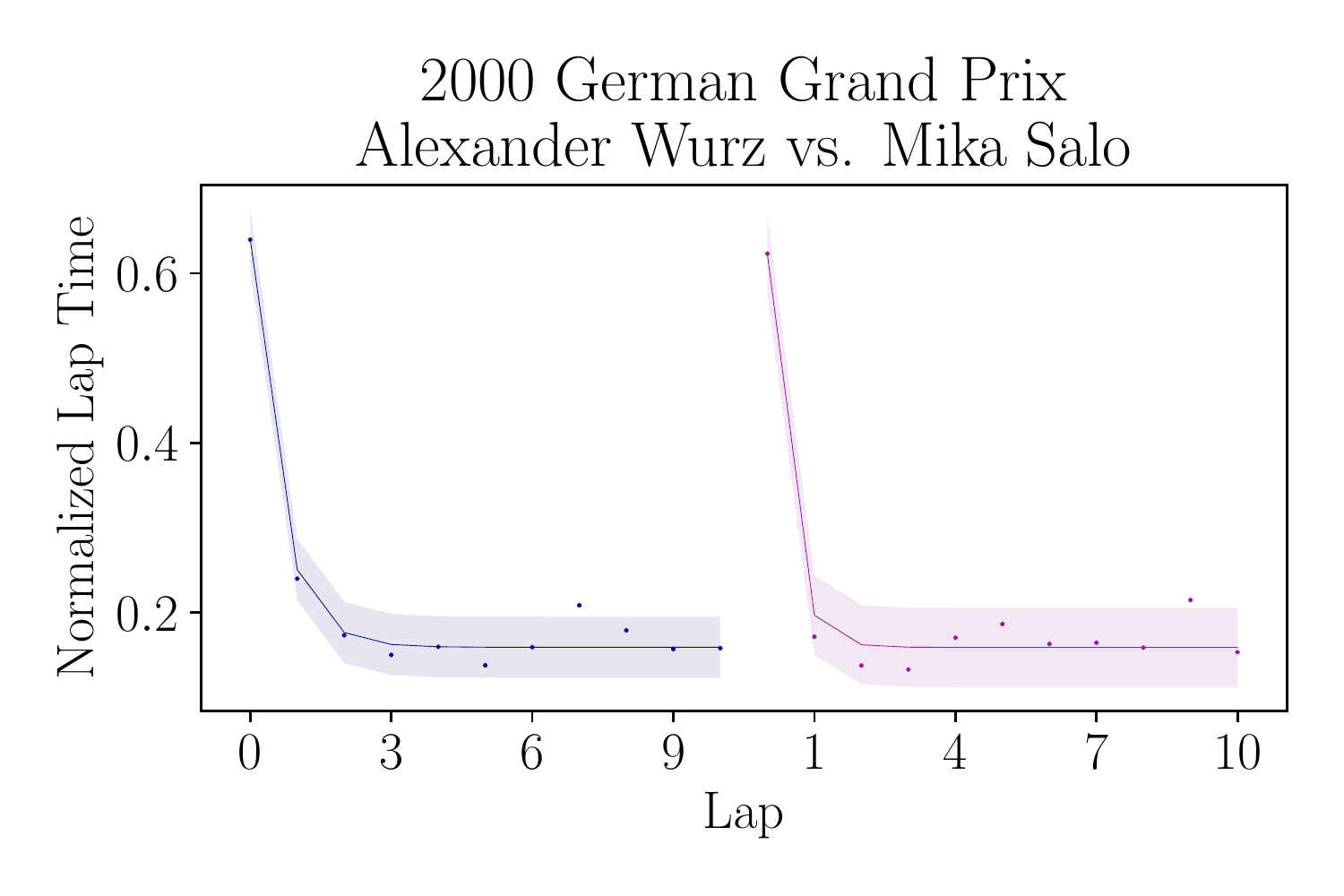} &
\includegraphics[width=.33\linewidth]{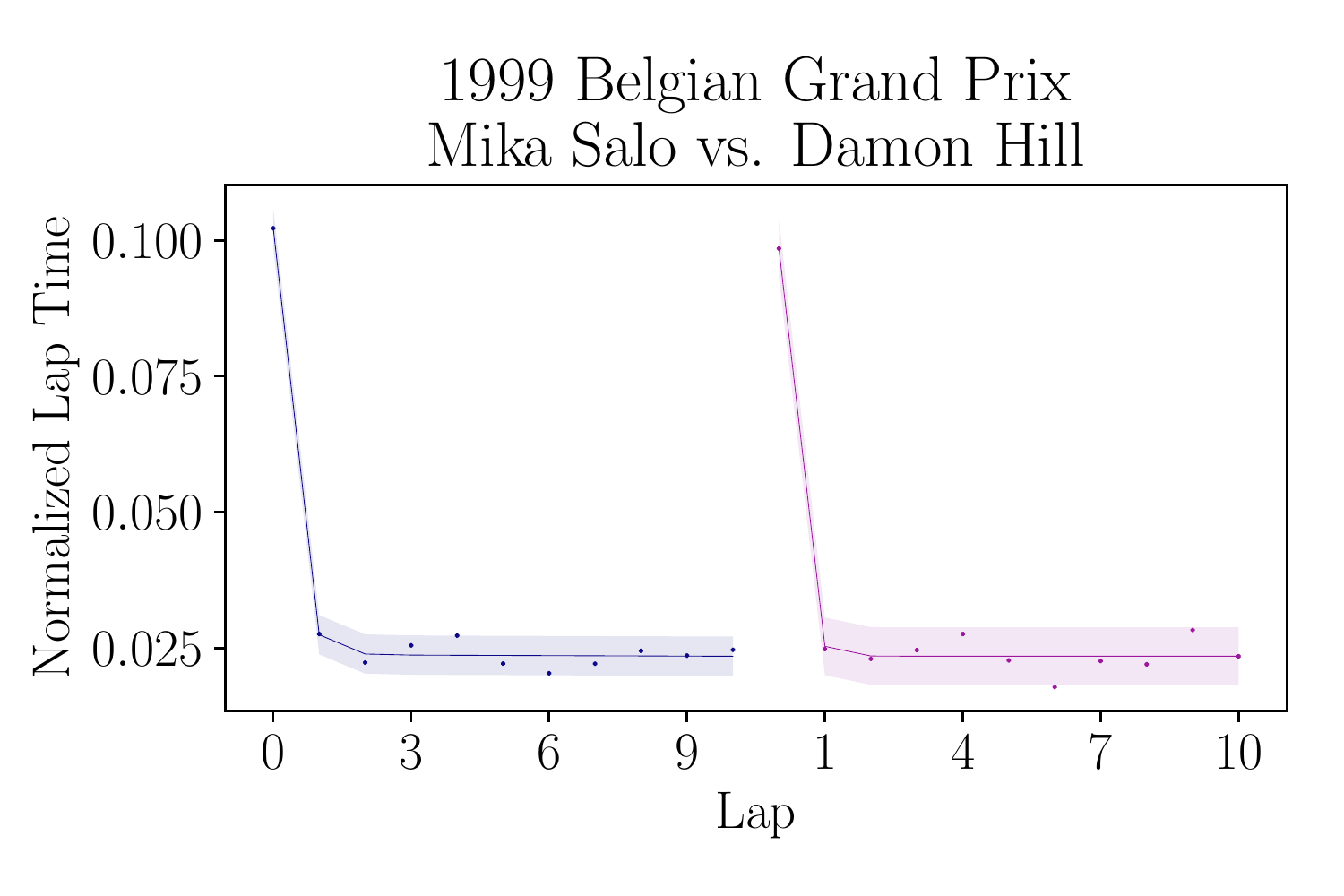}  \end{tabular}
\caption{In each plot, we select a driver pair and a particular race, and we depict our fitted probabilistic lap time model. In each plot, our probabilistic model parameterizes a distribution over lap times for each lap index (note that different plots have different lap indices). The solid line depicts the mean of this distribution, for each lap index. The shaded region contains $\pm 2$ standard deviations of this distribution, centered around the distribution's mean. The dotted points are the actual lap times (we note that these almost always lie within 2 standard deviations of the mean). Note that all the lap times are normalized, so that each lap time lies in the interval $[0,1]$.}
\label{fig:appendix_racing_reo_holds}
\end{figure}

At a high level, our probabilistic model stipulates that for each driver $i$ and race $j$, the lap times decrease exponentially at first and then linearly. More concretely, as the lap index $k$ increases, the expected lap time of the driver $i$ in race $j$ decreases at a rate determined by $\alpha_{i, j}$ (with initial condition $\beta_{i, j}$), but it never drops below $- |\gamma_{i,j}|k$ (our $k$ values are small, and so the lap times plateau to a positive number). The variance of the lap times for driver $i$ in race $j$ is $\sigma_{i, j}$ (note that this quantity is independent of $k$). Figure~\ref{fig:appendix_racing_reo_holds} shows normalized lap time data and the learned probabilistic model for 9 races (the probabilistic model for the 10th race is given in Figure~\ref{fig:racing_reo_holds}). By inspection, the probabilistic model is a reasonable approximation for the distribution of (sequential) lap times.

\subsubsection{Constructing Meaningful Tallying Bandit Instances}
In order to test SE, we further filter the drivers such that the driver lap time models define a meaningfully challenging bandit problem. Specifically, we define an eligible TB driver pair to be two drivers competing in the same race such that both of their terminal mean lap times are difficult to discriminate. In particular, denote the terminal expected lap time for driver $i$ (resp. $i'$) with $\mu_i$ (resp. $\mu_{i'}$). Then, drivers $i$ and $i'$ are considered to define an eligible TB driver pair if the following conditions hold:
\begin{itemize}
\item driver $i$ and $i'$ both compete in the same race $j$,
\item $\mu_i \in [\mu_{i'} - \sigma^2_{i',j}, \mu_{i'} + \sigma^2_{i',j}]$,
\item $\mu_{i'} \in [\mu_i - \sigma^2_{i,j}, \mu_i + \sigma^2_{i,j}]$.
\end{itemize}
Note these conditions essentially imply that the problem is at least as hard as distinguishing the means of two Gaussians that are ``close'' to each other. Ten eligible TB driver pairs meet these criteria, which we use to define 10 TB instances. Specifically, the lap time model of each driver in the pair for race $j$ defines the loss in a TB instance, with each driver corresponding to an arm. The (stochastic) instantaneous loss at timestep $t$ for `playing' driver $i$ in race $j$ depends on the number of times driver $i$ has completed a lap in race $j$ in the previous $m$ timesteps, or equivalently (the tally) denoted $k$. This instantaneous loss is thus $\tau_{i,j,k} \sim  \mathcal{N}(\beta_{i,j} \exp( - k \alpha_{i,j}) - |\gamma_{i,j}|k, \sigma^2_{i,j})$.  

\subsubsection{Extended Results}

We now show results for all 10 instances of our simulated F1 tournament, where each instance is constructed in similar fashion to the instance constructed in Section~\ref{sec:numerical_results_f1}. We select 10 different driver pairs and races in which they competed.  For each driver pair and race, we utilize F1 lap time data~\citep{f1dataset} to fit a probabilistic lap time model as discussed above. In Figure~\ref{fig:appendix_racing_reo_holds}, we illustrate our probabilistic models of lap times for each of the selected driver pairs, and show that their lap times tend to decrease as the lap index increases.

\begin{figure}[t]
\begin{tabular}{ccc}
\includegraphics[width=.33\linewidth]{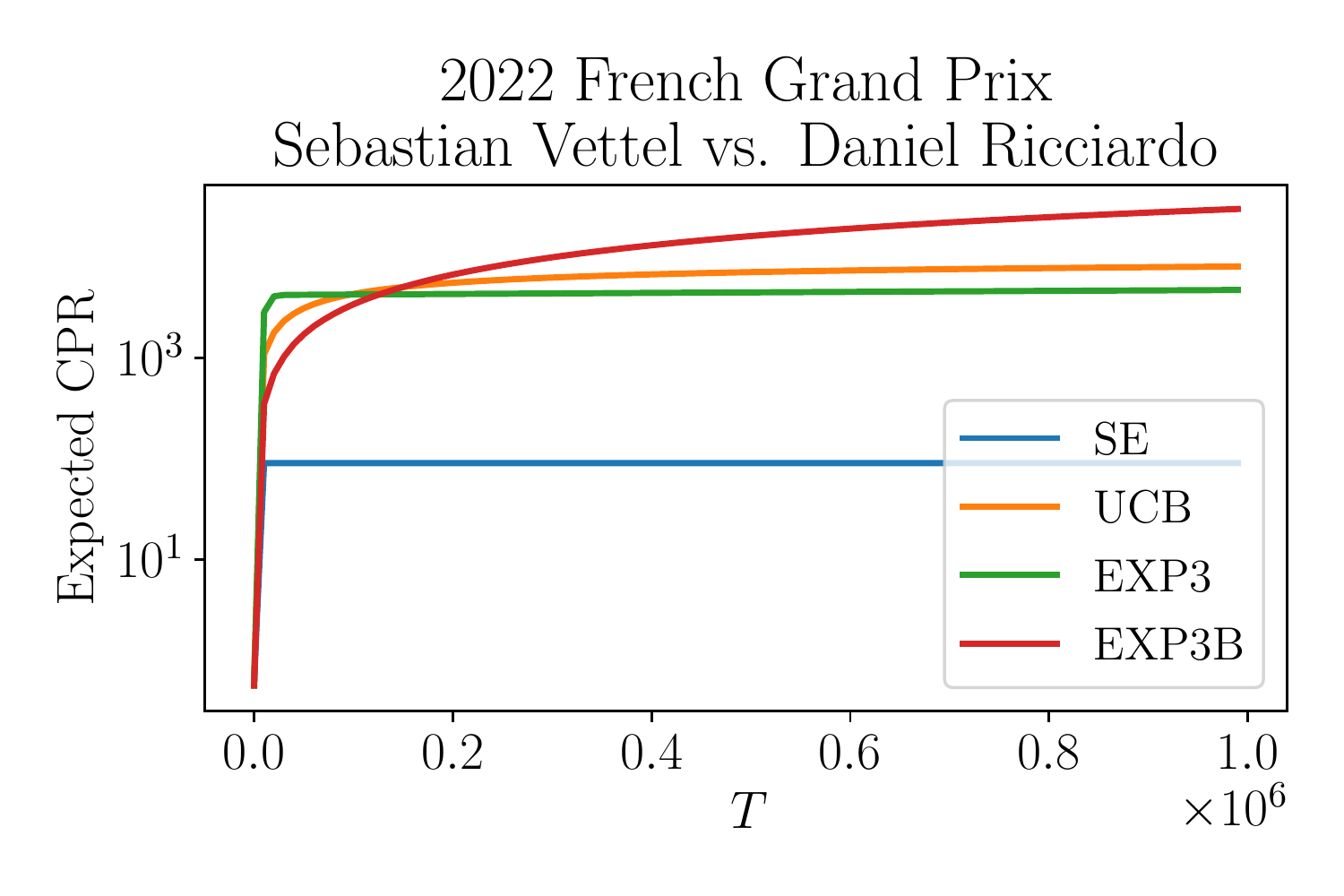} &  
\includegraphics[width=.33\linewidth]{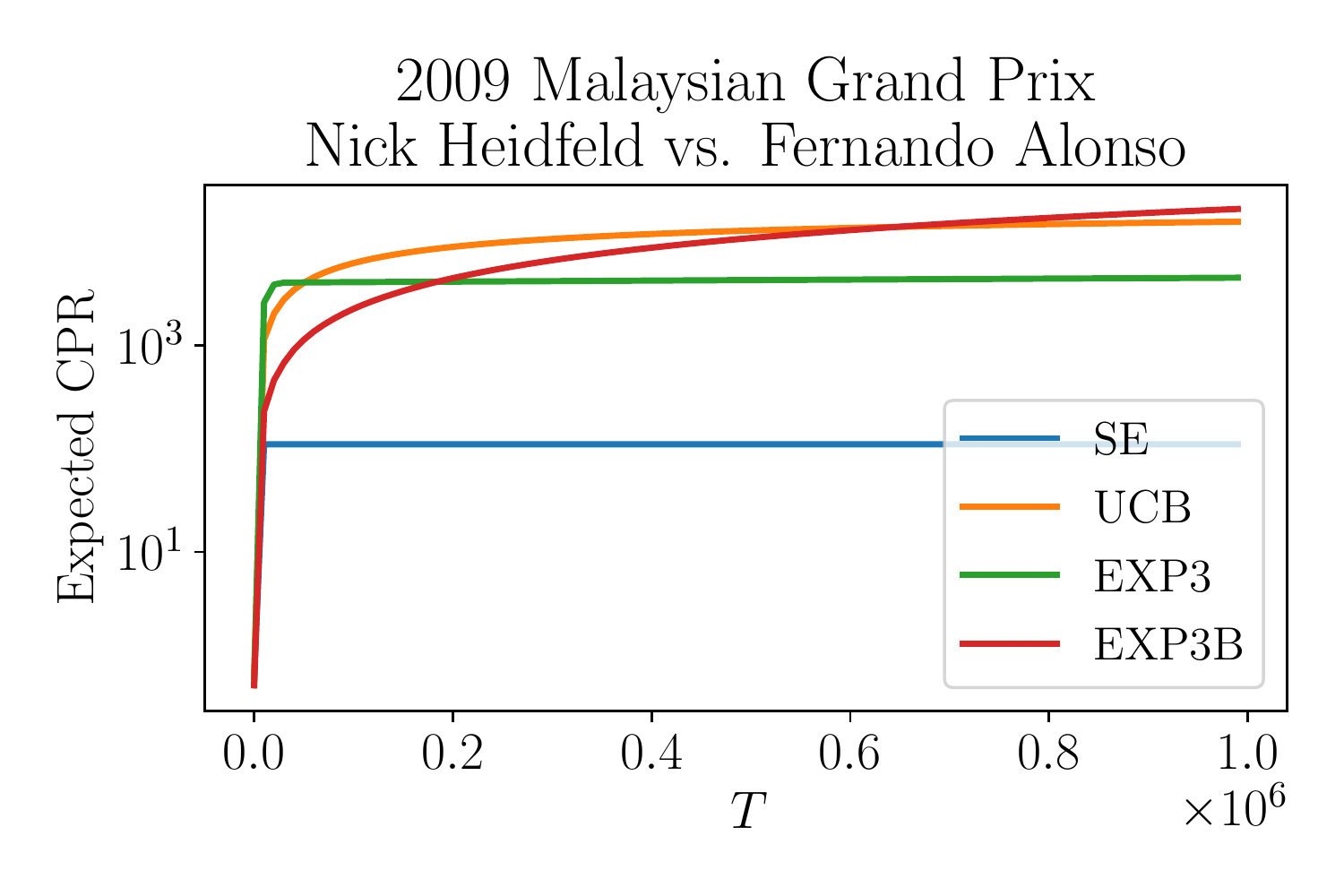}   &   \includegraphics[width=.33\linewidth]{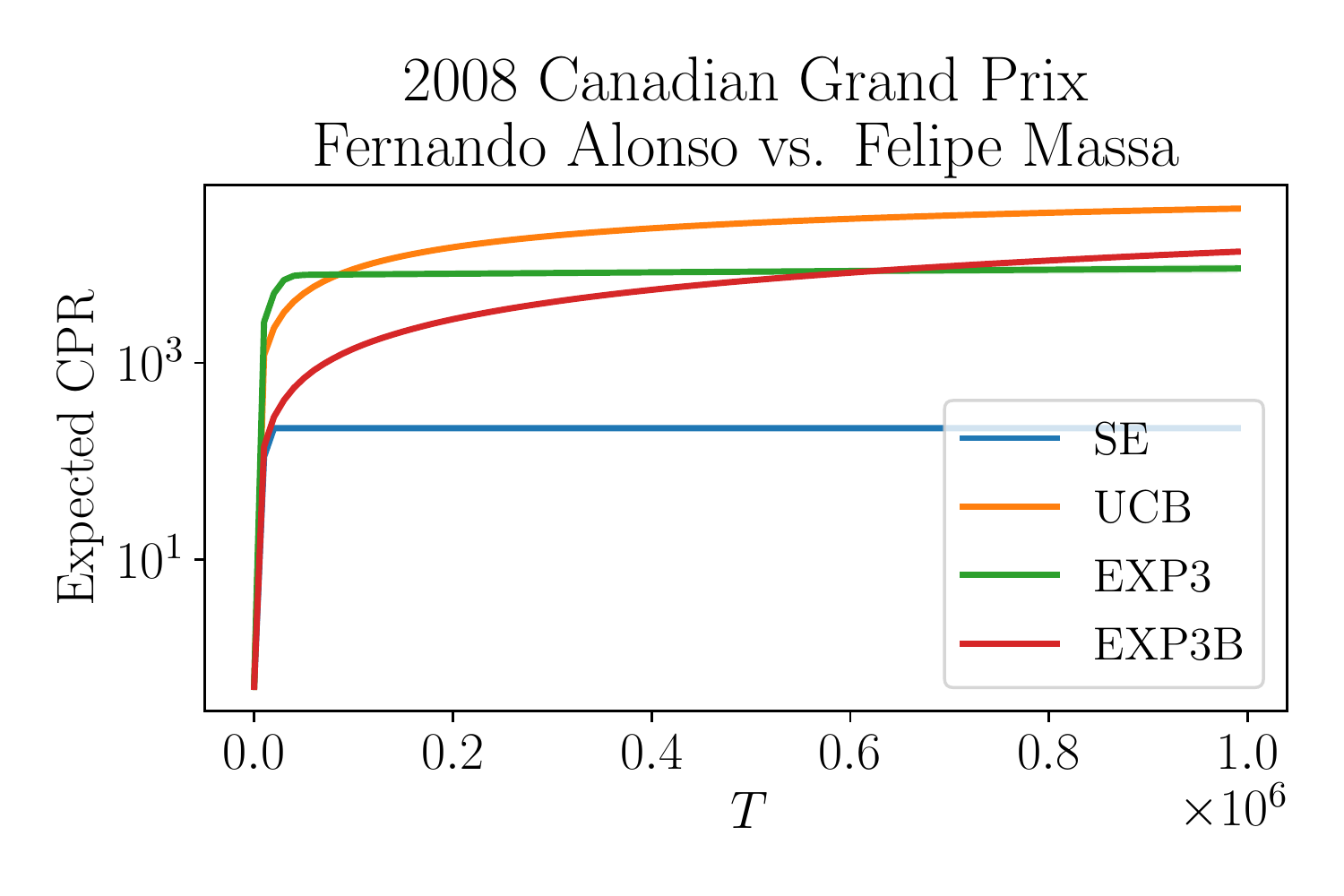}  \\
\includegraphics[width=.33\linewidth]{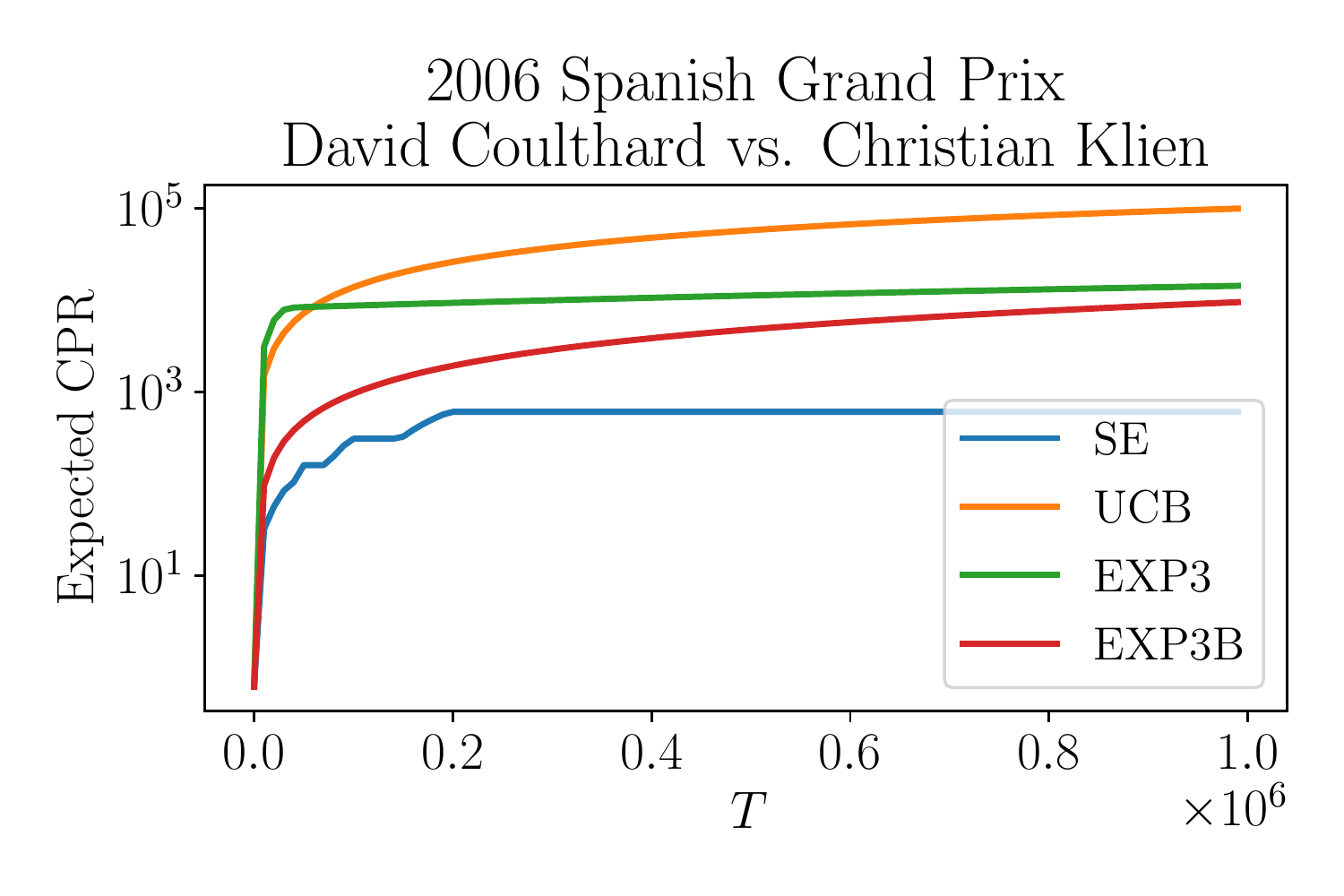}  &
\includegraphics[width=.33\linewidth]{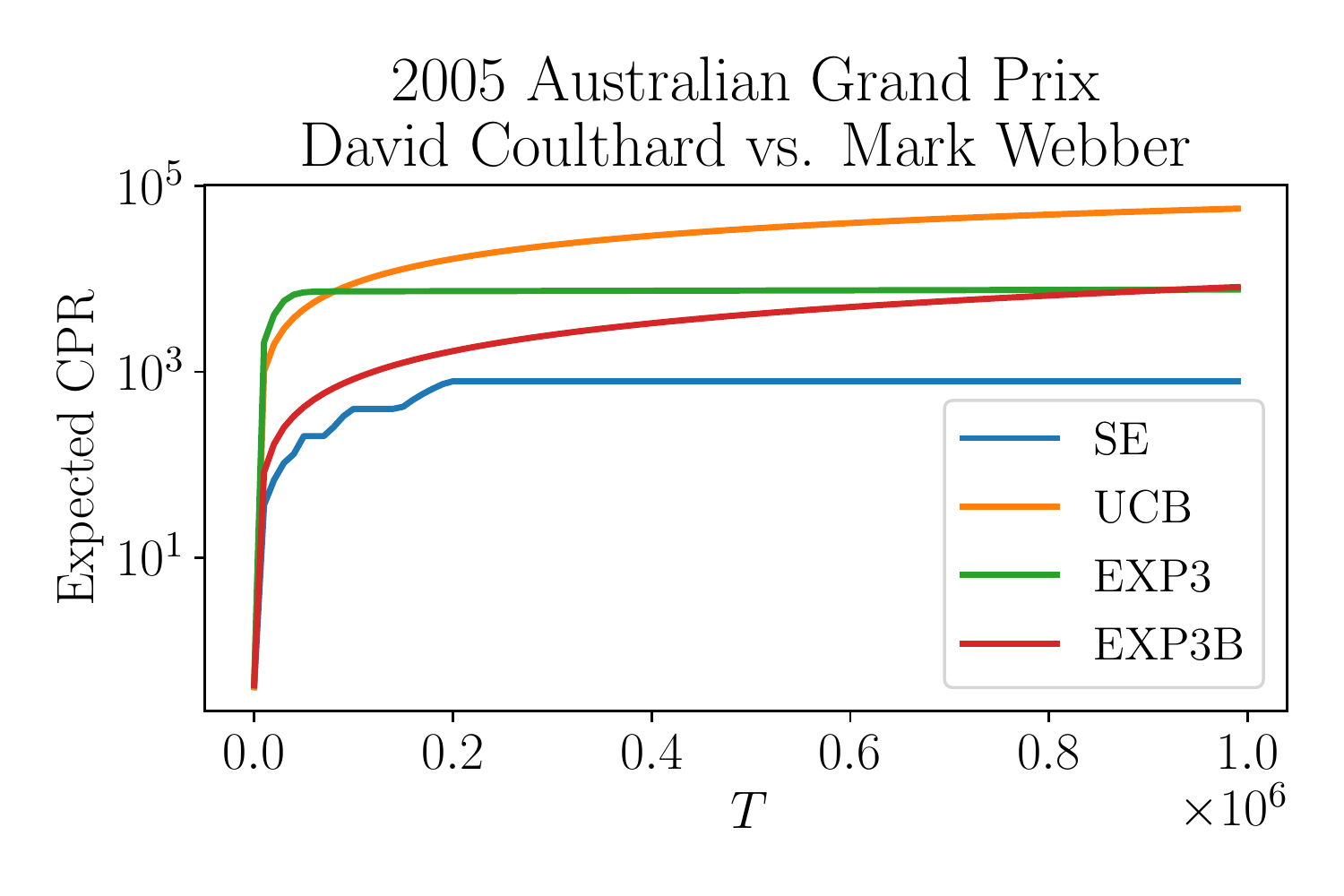} &  \includegraphics[width=.33\linewidth]{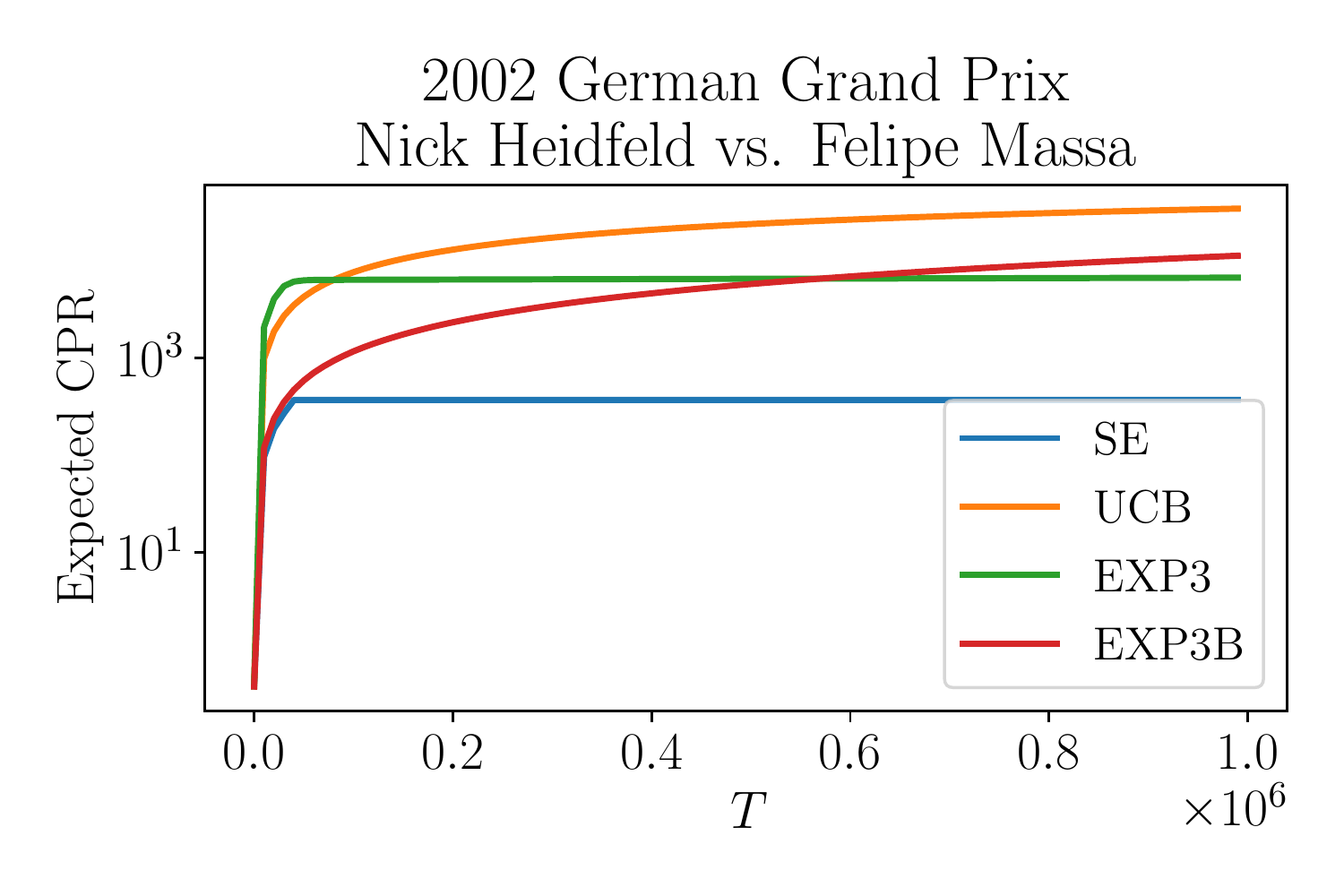}  \\
\includegraphics[width=.33\linewidth]{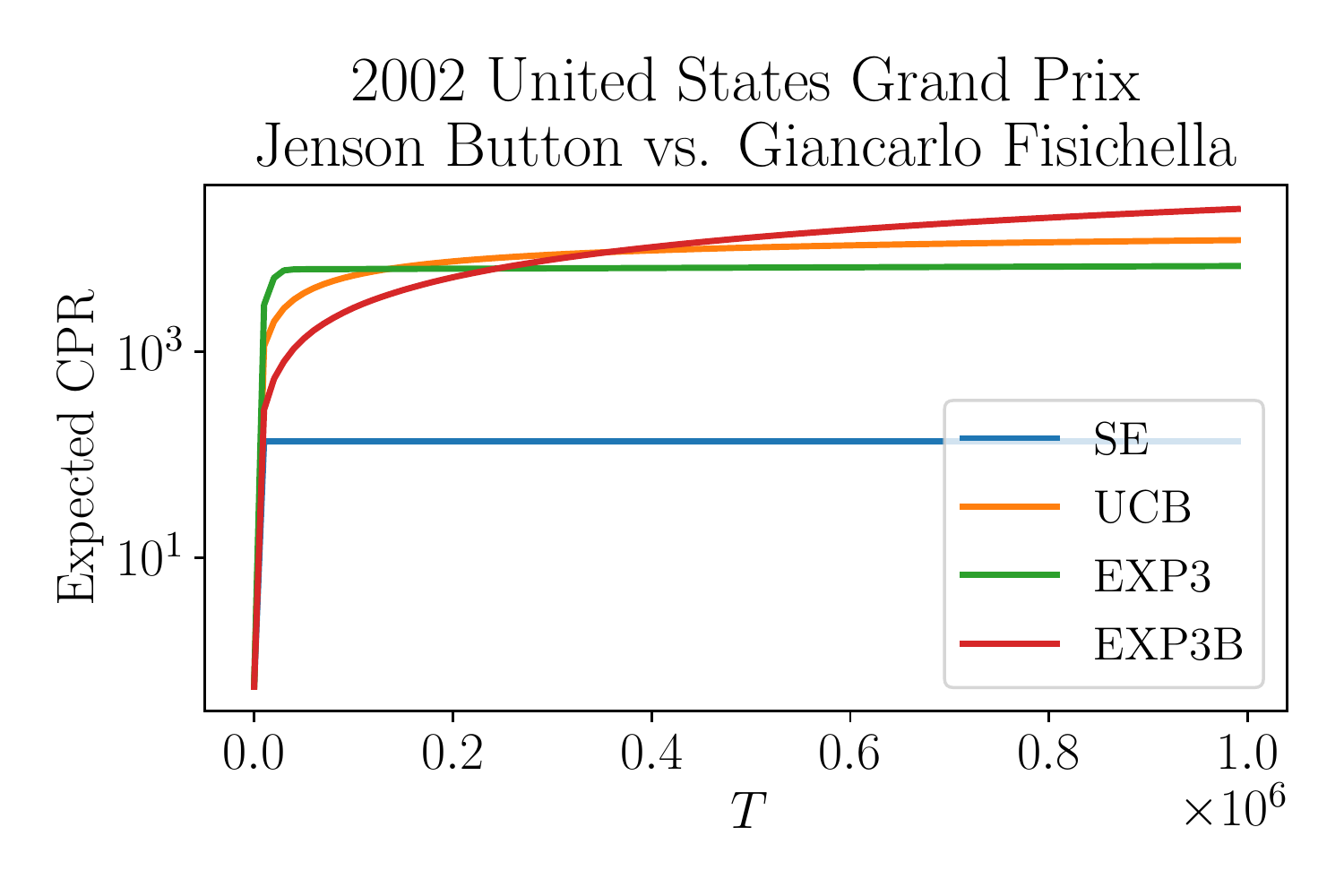} &
\includegraphics[width=.33\linewidth]{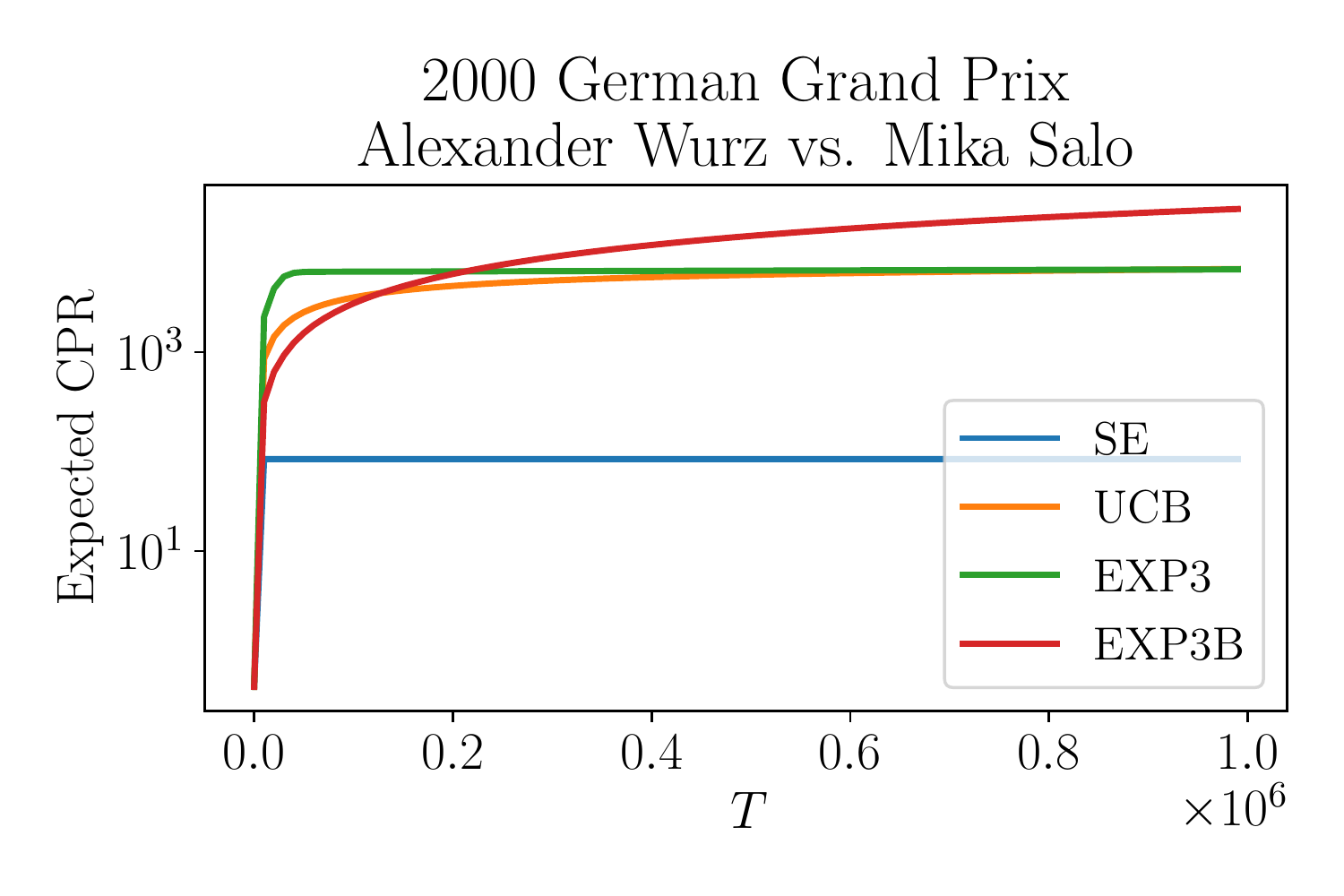} &
\includegraphics[width=.33\linewidth]{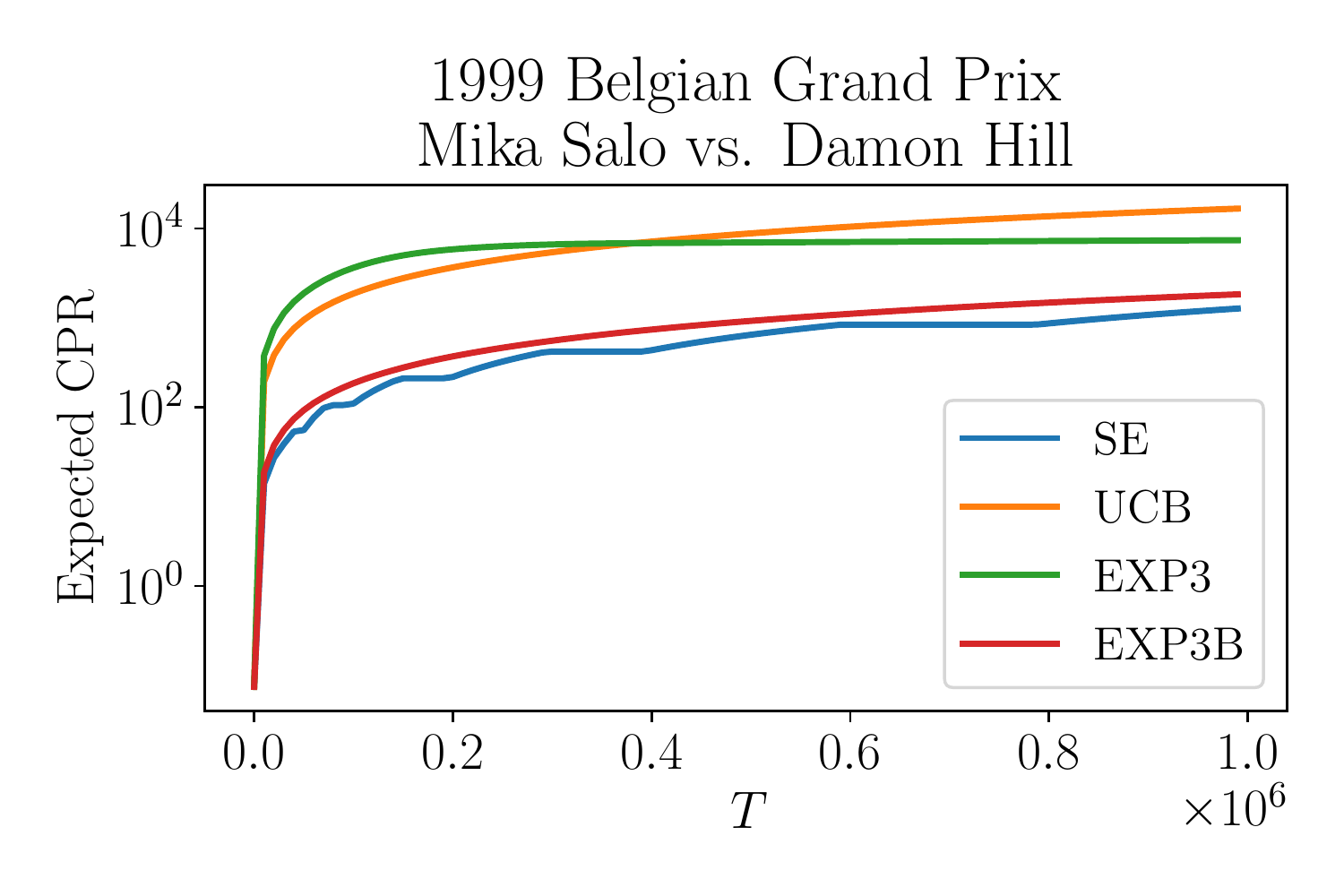}  \end{tabular}
\caption{For each plot in Figure~\ref{fig:appendix_racing_reo_holds}, we simulate a WTB instance and compare the performance of the various algorithms. We depict as a function of time the expected CPR of each algorithm. Data is obtained by averaging over 20 problem instances, each with $K = 2$ and $T = 10^6$. The shaded region depicts $\pm 1$ standard error around the mean. The value $m$ used in a plot is the length of the lap index in the corresponding plot in Figure~\ref{fig:appendix_racing_reo_holds}. We set $M$ to equal $m$.}
\label{fig:appendix_racing_results}
\end{figure}

We use the probabilistic models depicted in Figure~\ref{fig:appendix_racing_reo_holds} to simulate 9 instances of this tournament with $K = 2$ and $m$ equaling the number of lap indices in the appropriate plot. The results for the 10th race were given in Section~\ref{sec:numerical_results_f1}. For each instance, we maintain a tally of the number of times each driver in that instance was chosen in the prior $m$ timesteps. The loss associated with picking a driver is governed by the distribution parameterized by our fitted probabilistic model. In particular, if we pick driver $x$ and we have picked them $y$ times in the last $m$ timesteps, then the instantaneous loss is sampled from the distribution parameterized by our fitted probabilistic model for driver $x$ at lap index $y$. The two drivers for each instance are chosen such that their calibrated performance is difficult to distinguish, as discussed above. Note that in this setting, one has $o(T)$ CPR if and only if one plays the worse driver $o(T)$ many times. In Figure~\ref{fig:appendix_racing_results}, we plot each method's CPR over time for each of the 10 instances of this tournament, showing in each case that SE outperforms the baselines.

\newpage
\bibliographystyle{alpha}
\bibliography{reo_arxiv_submission}

\end{document}